\documentclass[runningheads]{llncs}

\usepackage{eccv}

\usepackage{eccvabbrv}

\usepackage{graphicx}
\usepackage{booktabs}

\usepackage[accsupp]{axessibility}  

\usepackage{hyperref}
\usepackage{orcidlink}
\usepackage{multirow}
\usepackage{adjustbox}
\usepackage{wrapfig}
\usepackage{caption}
\usepackage{array}
\usepackage{rotating}   
\usepackage{needspace}
\usepackage{enumitem}
\setlist[itemize]{label=\textbullet}

\usepackage{algorithm}
\usepackage{algpseudocode}
\usepackage{float}
\usepackage{makecell}   
\definecolor{skipgreen}{RGB}{51,116,35}
\usepackage[table]{xcolor}
\definecolor{casbg}{RGB}{242,252,254}
\usepackage{overpic}
\begin{document}

\title{To Adapt or Not to Adapt? Selective Adaptation for Vision-Language Models}
\titlerunning{Selective Adaptation for Vision-Language Models}

\author{Siru Jiang\inst{1,2}\textsuperscript{$\dagger$}\orcidlink{0009-0009-2210-2116} 
\and Yuwei Liang\inst{2,3}\textsuperscript{$\dagger$}
\and Jian Liang\inst{2,3}\thanks{Corresponding author. $^\dagger$ These authors are co-first authors.}\orcidlink{0000-0003-3890-1894}
\and \\ Ran He\inst{2,3}\orcidlink{0000-0002-3807-991X}
\and Tieniu Tan\inst{1,2,4}\orcidlink{0000-0003-4054-5649}
}

\authorrunning{Jiang et al.}

\institute{
School of Advanced Interdisciplinary Sciences, University of Chinese Academy of Sciences, China
\and NLPR \& MAIS, Institute of Automation, Chinese Academy of Sciences, China
\and School of Artificial Intelligence, University of Chinese Academy of Sciences, China
\and Nanjing University, China \\
\email{\{sirujiang324, liangjian92\}@gmail.com}
}

\maketitle

\begin{abstract}
  Test-time adaptation (TTA) has emerged as a prominent strategy for adapting vision-language models to distribution shifts during inference.
We conduct a per-sample analysis of model predictions before and after adaptation, and observe two failure modes in existing TTA methods that echo previous work. 
Adaptations are frequently negligible, yielding no change in the model’s predictions, and more severely, they can be detrimental by flipping previously correct predictions to incorrect ones.
This naturally raises a question: \emph{Can we identify and skip such negligible or harmful adaptations?}
In this work, we introduce a new problem of \textbf{selective adaptation}, which aims to determine whether a given test sample should undergo adaptation or be skipped.
To this end, we propose Cross-Augmentation Similarity (CAS), a simple baseline that performs adaptation only when predictions across augmented views exhibit low similarity. 
Notably, CAS not only preserves but in some cases improves overall accuracy, even when skipping nearly 85\% of the adaptation process. 
We hope other researchers will explore this new direction and surpass the performance of our baseline. 
Our code is available at \url{https://github.com/sirujiang/selective-adaptation}.

\end{abstract}

\section{Introduction}
\begin{figure}[t]
\centering
\begin{subfigure}[b]{0.35\textwidth}
    \centering
    \includegraphics[width=\textwidth]{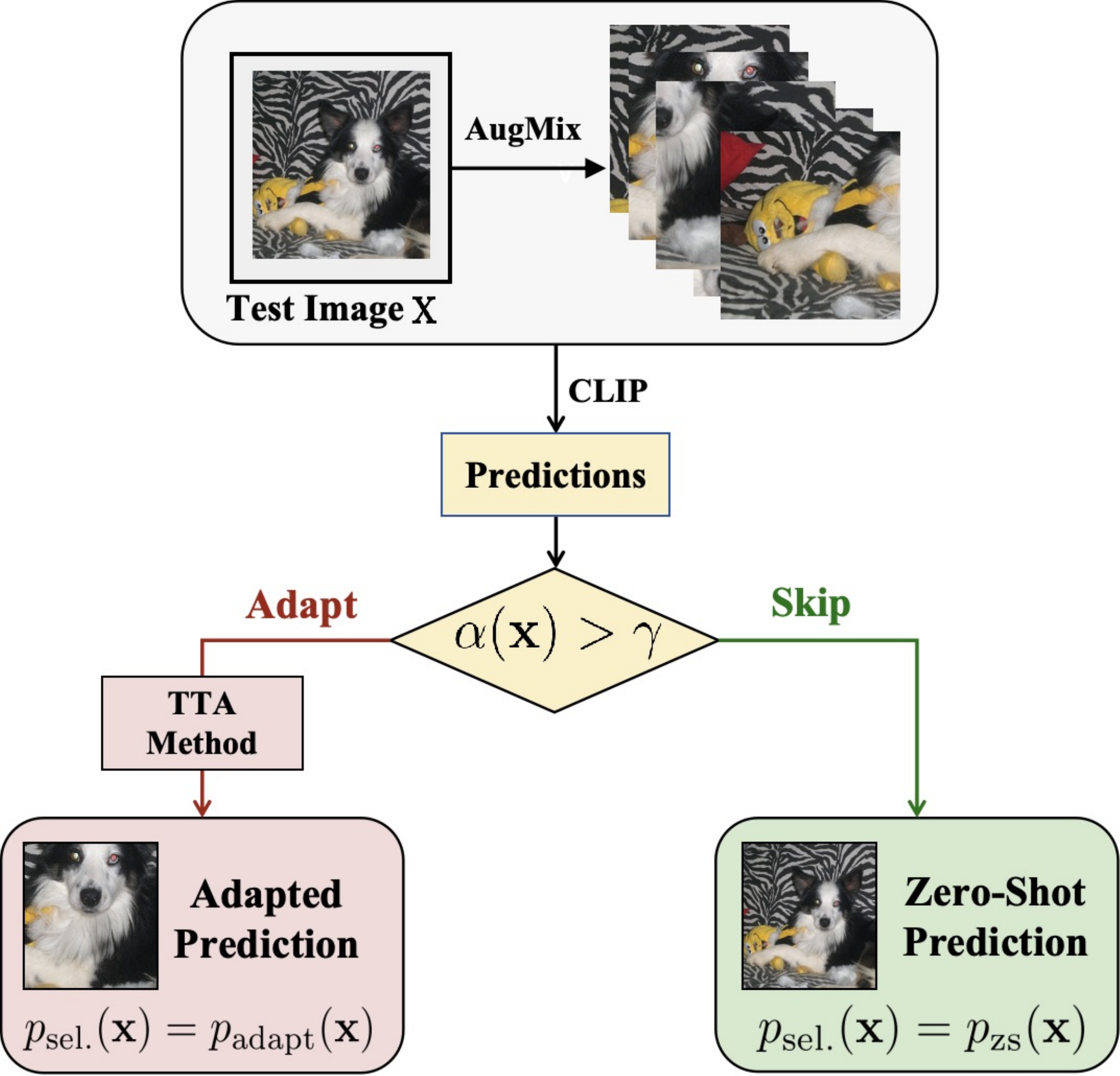}
    \caption{Selective Adaptation}
    \label{fig:tpt_ab}
\end{subfigure}
\hspace{20pt} 
\begin{subfigure}[b]{0.42\textwidth}
    \centering
    \includegraphics[width=\textwidth]{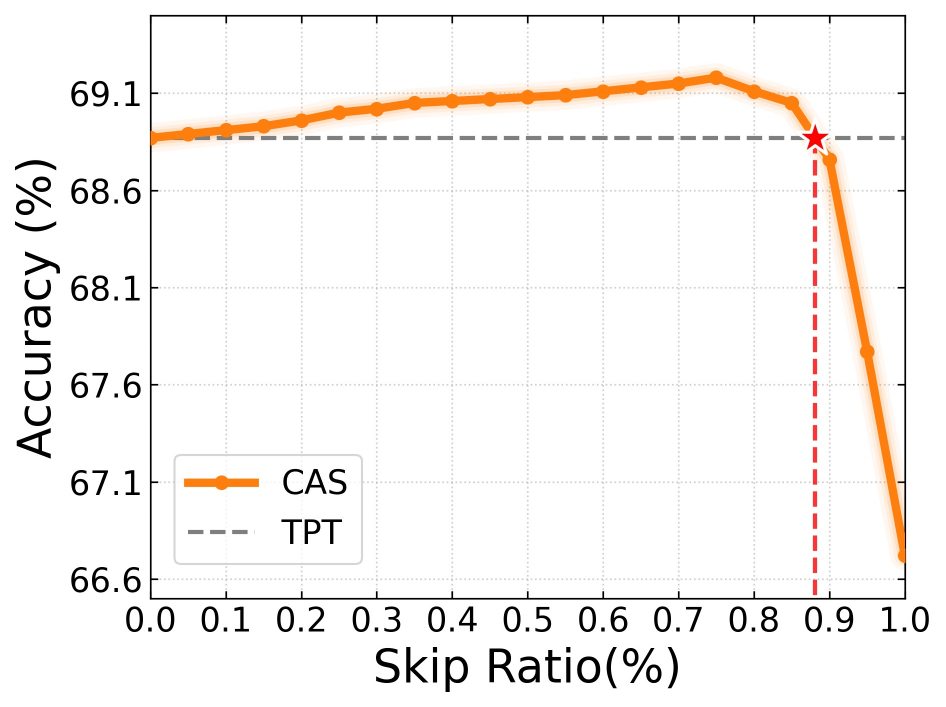}
    \caption{Accuracy versus Skip Ratio}
    \label{fig:flow_ab}
\end{subfigure}

\caption{
(a) Given a test image $x$, a score $\alpha(x)$ is computed. 
Samples with low scores undergo adaptation, while those with high scores are skipped, with the zero-shot prediction used instead. 
(b) The proposed CAS maintains or slightly improves accuracy across a wide range of skip ratios (the proportion of \textcolor{skipgreen}{skipped adaptations}) on ImageNet.
}

\label{fig:abs_double}
\vspace{-15pt}
\end{figure}

Vision-language models (VLMs), such as CLIP~\cite{radford2021learning}, ALIGN~\cite{jia2021scaling}, Flamingo~\cite{alayrac2022flamingo}, and LLaVA~\cite{liu2023visual}, are pretrained on large-scale image–text pairs and have demonstrated strong generalization across diverse vision tasks~\cite{zhang2024vision,wu2023cap4video,zhou2022learning,zhou2022conditional}.
Among them, CLIP~\cite{radford2021learning} aligns visual and textual representations in a shared embedding space, enabling impressive zero-shot performance.
However, VLMs remain sensitive to distribution shifts and often suffer from performance degradation when the test distribution differs from that of pretraining~\cite{saenko2010adapting,liang2025comprehensive,kim2021domain,yu2023benchmarking}.

In recent years, test-time adaptation (TTA) has emerged as an effective paradigm for mitigating distribution shifts by adapting VLMs with unlabeled test data.
In the context of image classification, existing TTA methods mainly focus on improving prediction accuracy~\cite{shu2022testtime,dafnis2025testtime,farina2024frustratingly} or enhancing model calibration~\cite{yoon2024ctpt,sharifdeen2025otpt,ahamed2026atpt}.
Despite their effectiveness, they implicitly assume that adaptation is beneficial for all test samples.
To better understand this issue, we conduct a preliminary per-sample analysis of the adaptation process under the classic TPT framework~\cite{shu2022testtime}.

Consistent with observations in prior work~\cite{farina2024frustratingly}, we find that a large proportion of predictions remain unchanged before and after adaptation, leading to unnecessary computational overhead.
More critically, some originally correct predictions flip to incorrect after adaptation, leading to performance degradation.
Notably, these negligible or even harmful adaptations account for more than 90\% of all adaptation processes.
Motivated by these observations, we propose to identify and skip ineffective adaptations, thereby improving efficiency while preserving accuracy.
Unlike previous TTA approaches that enhance efficiency through parameter-free retrieval~\cite{karmanov2024efficient,dastmalchi2025etta} or lightweight parameter optimization~\cite{imam2025test,dafnis2025testtime}, our method improves efficiency at the per-sample level by adapting only when necessary.

In this work, we introduce a new problem, termed \textbf{selective adaptation}, which formulates a binary detection task to identify whether the adaptation for a given test sample should be performed or skipped.
An illustration of this problem is shown in Fig.~\ref{fig:abs_double} (a).
Similar to out-of-distribution (OOD) detection~\cite{hendrycks2016baseline,hendrycks2019scaling,liu2020energy,ming2022delving,granese2021doctor,lu2025out,yang2023auto}, it relies on a scoring function to identify ineffective adaptations.
Specifically, samples with higher scores tend to correspond to negligible or harmful cases and use zero-shot predictions directly, while those with lower scores continue to undergo adaptation.
Generally, the goal of selective adaptation is to skip as many ineffective adaptations as possible while maintaining or even improving overall accuracy.
To evaluate this problem, we adopt the standard Area Under the ROC Curve (AUC)~\cite{davis2006relationship,fawcett2006introduction} for detection quality, and introduce Accuracy Expectation with a Triangular Prior (AEP) to measure expected accuracy under different skip ratios.

Test-time augmentation~\cite{shanmugam2021better,li2025pataug,shu2022testtime,farina2024frustratingly} has been used in many TTA methods~\cite{zhang2022memo,shu2022testtime} to generate multiple views for an input sample.
We argue that the prediction similarity between the test sample and its augmented views may be closely correlated with adaptation effectiveness.
This motivates our simple baseline Cross-Augmentation Similarity (CAS), which computes a score based on the prediction similarity across multiple augmented views.
We compare CAS against random skipping and established OOD detection methods~\cite{liu2020energy, ming2022delving} under representative TTA frameworks~\cite{shu2022testtime,sheng2025r,dafnis2025testtime,farina2024frustratingly}.
Overall, CAS achieves an AUC of around 90\%.
More importantly, it preserves and even improves the accuracy of full adaptation while skipping 85\% of adaptations on ImageNet, its variants, and multiple fine-grained benchmarks.
The result of ImageNet is shown in Fig.~\ref{fig:abs_double} (b).
Beyond classification accuracy, CAS also maintains calibration performance in calibration-oriented TTA methods~\cite{yoon2024ctpt,sharifdeen2025otpt}.
Our contributions are summarized as follows:
\begin{itemize}
\item We introduce selective adaptation,
an underexplored direction to improve TTA efficiency by detecting whether a test sample can benefit from adaptation.
\item We provide Cross-Augmentation Similarity (CAS), a simple baseline based on prediction similarity across test-time augmented views.
\item Extensive experiments validate that CAS maintains and even improves TTA performance, providing a baseline for future research to advance.
\end{itemize}

\section{Related Work}

\textbf{Test-time adaptation (TTA).}
TTA aims to mitigate performance degradation caused by distribution shifts by adapting models to unlabeled test data~\cite{liang2025comprehensive,sheng2025illusion}.
TTA approaches can be broadly categorized into two paradigms based on how they process test data.
Online TTA~\cite{wang2021tent,wang2022continual,yuan2023robust,xiao2025dynaprompt,zhou2025bayesian,yu2024stamp} processes streaming data and updates model parameters by leveraging historical knowledge from previous test samples.
In contrast, episodic TTA, such as MEMO~\cite{zhang2022memo} and TTT~\cite{sun2020test}, treats each test sample independently, making adaptation more challenging.
Throughout this work, we focus exclusively on the episodic paradigm.

With the rise of VLMs~\cite{radford2021learning,jia2021scaling,alayrac2022flamingo}, increasing attention has been devoted to applying TTA~\cite{sheng2025r,sharifdeen2025otpt,shu2022testtime} to CLIP~\cite{radford2021learning}.
A line of work explores training-based TTA methods~\cite{shu2022testtime,feng2023diverse,li2025pataug}, which adapt models at test time by optimizing a subset of parameters using unlabeled test samples.
The pioneering work TPT~\cite{shu2022testtime} adapts the model by optimizing learnable prompts through entropy minimization with confidence selection.
Alternatively, another line of work explores training-free paradigms, such as ZERO~\cite{farina2024frustratingly}, MTA~\cite{zanella2024test}, and TPS~\cite{sui2025just}, which enable fast test-time adaptation without requiring gradient updates.
In addition to accuracy improvement, recent studies have also explored other aspects of performance, including calibration~\cite{yoon2024ctpt,ahamed2026atpt,sharifdeen2025otpt} and adversarial robustness~\cite{sheng2025r,xing2025clip,wang2025tapt}.

\textbf{Efficiency in TTA.}
In addition to improving adaptation performance, several studies~\cite{dafnis2025testtime,karmanov2024efficient,zhou2025trainingfree,niu2022efficient} focus on enhancing efficiency.
In episodic TTA, most existing work~\cite{imam2025test,dafnis2025testtime} focuses on making the optimization process more efficient. 
TTL~\cite{imam2025test} improves efficiency by optimizing low-rank adapters
and STS~\cite{dafnis2025testtime} adapts only a small number of parameters.
Instead of refining the optimization algorithm, we propose selective skipping, which identifies and skips ineffective adaptations to improve efficiency without sacrificing performance.
Notably, a line of work in online TTA~\cite{karmanov2024efficient,zhang2024boostadapter,zhou2025trainingfree,niu2022efficient} has also explored efficiency improvements. For example, EATA~\cite{niu2022efficient} improves efficiency by selecting informative samples for adaptation based on entropy, while other methods reduce computational overhead via key–value cache retrieval~\cite{karmanov2024efficient,zhang2024boostadapter,zhou2025trainingfree,dastmalchi2025etta}. These approaches differ fundamentally from our selective adaptation approach.

\textbf{Out-of-distribution (OOD) detection and selective classification.}
OOD detection\cite{hendrycks2019scaling,yang2024generalized,ming2022delving,granese2021doctor} aims to identify test samples that differ from the training distribution to ensure model reliability.
Early work initially introduced MSP~\cite{hendrycks2016baseline} for detecting misclassified or OOD samples, followed by Energy~\cite{liu2020energy}, MCM~\cite{ming2022delving} and Doctor~\cite{granese2021doctor}.
We leverage the scoring functions provided by these methods as a comparison strategy and employ AUC to evaluate the effectiveness of our selective adaptation baseline. Selective classification~\cite{geifman2017selective,cortes2016learning,fisch2022calibrated,liang2024selective}, also known as classification with a rejection option, is a machine learning framework that allows a model to abstain from making a prediction when it is uncertain.
While selective classification typically abstains from making predictions after inference~\cite{geifman2019selectivenet}, selective adaptation introduced in this paper instead focuses on rejecting samples before performing adaptation.

\section{Method}

\subsection{Preliminaries}
CLIP~\cite{radford2021learning} is a widely used VLM due to its strong zero-shot generalization capability.
It consists of two parts, a visual encoder $f_v(\cdot)$ and a text encoder $f_t(\cdot)$.
For a $K$-class classification task with a label space $\mathcal{Y} = \{y_1, y_2, \dots, y_K\}$, let $x$ denote an input image and $y \in \mathcal{Y}$ denote its ground-truth label.
The visual encoder extracts visual features from the input image $v = f_v(x)$.
For the text encoder, each class label $y_k \in \mathcal{Y}$ is converted into a textual prompt $c_k$ (e.g., using a template such as “a photo of a [class]”) and encoded into a textual feature $t_k = f_t(c_k)$.
The prediction probability is then computed as
\begin{equation}
    p^k(x)= p(y=k \mid x) = \frac{\exp\left(\mathrm{cos}(v, t_k)/\tau  \right)}{\sum_{j=1}^{K} \exp\left(\mathrm{cos}(v, t_j)/\tau  \right)},
\end{equation}
where $\tau$ is a temperature parameter and 
$\cos(\cdot,\cdot)$ denotes cosine similarity.
Given a test image $x$, $y_{\text{zs}}(x) = \arg\max_{k}\; p_{\text{zs}}^k(x)$ and $y_{\text{adapt}}(x) = \arg\max_{k}\; p_\text{adapt}^k(x)$, where $p_\text{zs}(x)$ and $p_\text{adapt}(x)$ denote the zero-shot and adapted probability vectors, and $y_{\text{zs}}(x)$ and $y_{\text{adapt}}(x)$ are their corresponding predicted labels.
\begin{figure}[t]
\centering
\begin{subfigure}[b]{0.6\textwidth}
\centering
\includegraphics[width=\textwidth]{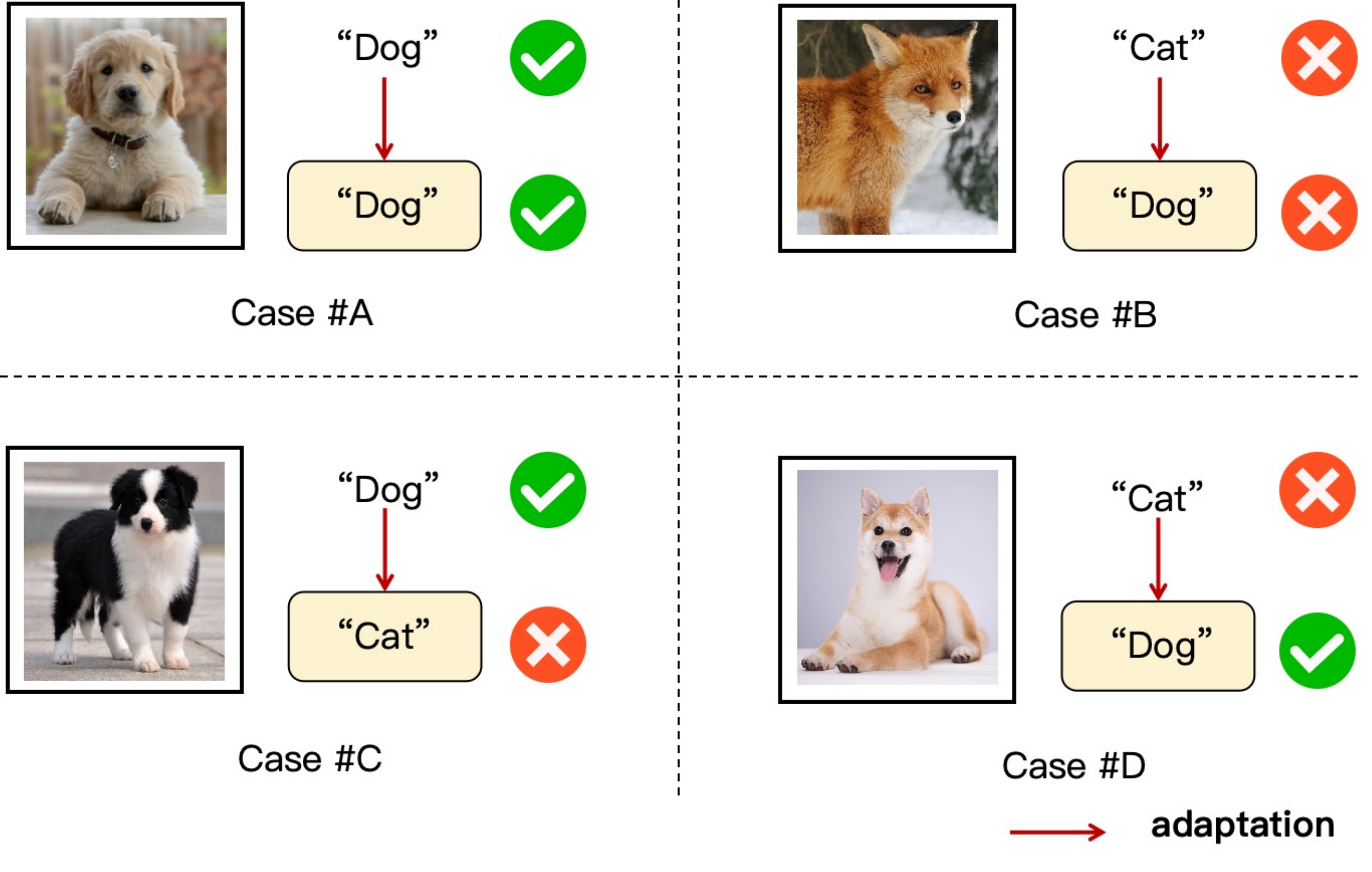}
\caption{Four Adaptation Cases}
\label{fig:large_left}
\end{subfigure}
\hspace{0.6cm}
\begin{subfigure}[b]{0.27\textwidth}
\centering
\includegraphics[width=\textwidth]{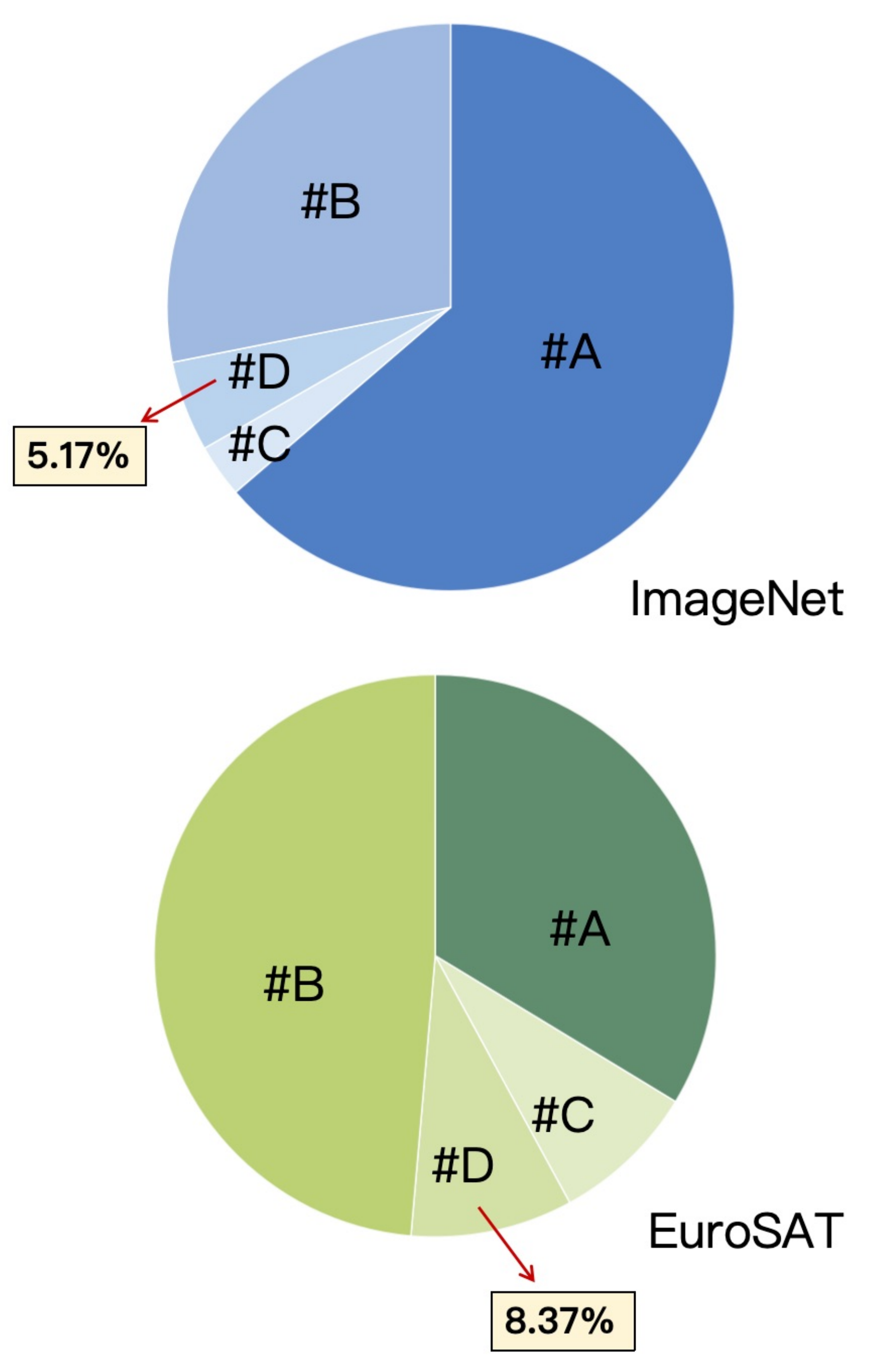}
\caption{Distribution of Cases}
\label{fig:small_top}
\end{subfigure}
\caption{ 
(a) We categorize the adaptation process into four cases (A–D). 
Checkmarks and crosses indicate prediction correctness, with zero-shot prediction shown above and TTA prediction shown below.
The arrow denotes the adaptation process. 
(b) Proportion of each case on ImageNet and EuroSAT under TPT~\cite{shu2022testtime} framework. The percentage of the beneficial-only case is highlighted.}
\label{fig:intro_fourcase}
\end{figure}

\subsection{Problem Formulation: Selective Adaptation}
\textbf{Motivation.}
Consistent with ZERO~\cite{farina2024frustratingly}, we find that a large proportion of adaptations are negligible or even harmful.
As illustrated in Fig.~\ref{fig:intro_fourcase} (a), the outcomes of adaptation can be categorized into four cases based on prediction changes between the zero-shot and adapted models.
Based on their impact on performance, we further group these cases into three types: negligible adaptation, harmful adaptation, and beneficial adaptation.
\begin{itemize}
    \item \textbf{Harmful adaptation.} The zero-shot prediction is correct, but becomes incorrect after adaptation$(y_{\text{zs}}(x) = y,
    y_{\text{adapt}}(x) \neq y)$. These adaptations will reduce the overall performance.
    \item \textbf{Negligible adaptation.} The zero-shot prediction remains unchanged after adaptation $(y_{\text{zs}}(x) = y, \; y_{\text{adapt}}(x) = y)\text \ {or} \ 
    (y_{\text{zs}}(x) \neq y, \; y_{\text{adapt}}(x) \neq y)$. 
    Adapting these adaptations incurs unnecessary computational overhead.
    \item \textbf{Beneficial adaptation.} The zero-shot prediction is incorrect but is adapted to correct successfully $(
    y_{\text{zs}}(x) \neq y, \  
    y_{\text{adapt}}(x) = y)$. 
    These adaptations are the only ones to improve the performance.
\end{itemize}
Notably, negligible and harmful adaptations dominate the test set in many datasets. 
Fig.~\ref{fig:intro_fourcase} (b) further shows that these cases account for over 90\% of test samples on datasets such as ImageNet and EuroSAT, highlighting the importance of identifying and skipping ineffective adaptations.

\textbf{Formulation.}
We formulate the selective adaptation problem as a binary detection task. 
Given a test sample $x$, we decide whether to perform adaptation or skip it.
Following the paradigm used in OOD detection, we define a decision function $G_\gamma(x)$ based on a scoring function $\alpha(x)$ and a threshold $\gamma$:
\begin{equation}
G_\gamma(x) =
\begin{cases}
\text{Not Adapt} & \text{if } \alpha(x) \geq \gamma \\
\text{Adapt} & \text{otherwise}
\end{cases}
\end{equation}
where $\gamma$ is designed to control the trade-off between performance and efficiency, and  $\alpha(x)$ is defined based on prediction similarity across augmented views.
A larger $\alpha(x)$ indicates a higher likelihood of skipping adaptation for the input image $x$.
Given a fixed threshold $\gamma$, a unique skip ratio $s$ is determined,
where $s$ denotes the proportion of test samples for which adaptation is skipped.

\textbf{Evaluation metrics.}
To evaluate the selective skipping strategy comprehensively, we introduce the following metrics:
\begin{itemize}
    \item \textbf{Area under the ROC curve (AUC).} We formulate the identification of ineffective adaptations as a binary detection task. AUC~\cite{davis2006relationship,fawcett2006introduction} measures how well a scoring function distinguishes ineffective adaptations from beneficial ones. An AUC of 0.5 corresponds to random guessing, while higher values indicate stronger discriminative capability.
    \item \textbf{Accuracy expectation with triangular prior (AEP).} Since the optimal skip ratio $s^*$ may vary across deployment scenarios, we propose a metric to evaluate overall performance by computing the expected accuracy under a prior distribution of $s$. Specifically, we adopt a triangular prior with probability density function $f(s) = 2(1-s)$ for $s \in [0,1]$. As $s$ increases, efficiency gains become more significant, and slight accuracy degradation becomes more acceptable. Therefore, performance is assigned a lower weight at larger skip ratios. The metric is defined as 
    \begin{equation}
        \text{AEP} = \int_{0}^{1} \text{acc}(s)\cdot 2(1-s)\, ds,
        \label{eq:aep}
    \end{equation}
    where $\text{acc}(s)$ denotes the model accuracy under a skip ratio $s$. By integrating the skip-accuracy curve, AEP provides a comprehensive evaluation of performance under different computational budgets. Notably, $\text{acc}(s)$ can be replaced with other performance metrics (e.g., ECE~\cite{guo2017calibration}) to evaluate different aspects of model behavior.
\end{itemize}

\subsection{Cross-Augmentation Similarity as a Simple Baseline}
Test-time augmentation~\cite{shanmugam2021better,kim2020learning}, derived from data augmentation techniques~\cite{yin2019fourier}, applies random transformations to test samples during inference.
It has been widely adopted in TTA methods such as MEMO~\cite{zhang2022memo} and TPT~\cite{shu2022testtime}.
Specifically, TPT~\cite{shu2022testtime} generates $(N-1)$ augmented views for a test image $x$ using AugMix~\cite{hendrycks2019augmix}. 
Let $\{\mathcal{A}_i(x)\}_{i=0}^{N-1}$ denote the augmented views, where $\mathcal{A}_0(x)$ is the original image. 
A cutoff percentile $\rho \in [0,1]$ is applied over the $N$ augmented views to select high-confidence samples.
Views whose prediction entropy is lower than the threshold $\beta$ are retained to form the high-confidence set $S$:
\begin{equation}
    S = \{ i \mid {H}(p_{\text{zs}}(\mathcal{A}_i(x))) \le \beta, \ i \in [0, N-1]\},
\end{equation}
where $\beta$ represents the $\rho$-percentile entropy threshold, and $H(\cdot)$ denotes Shannon entropy.
To promote cross-view consistency, TPT optimizes textual prompts by minimizing the entropy of the averaged predictions over the selected set $S$:
\begin{equation}
    {L}_{\text{TPT}}(x) = {H} \left( \frac{1}{|S|} \sum_{i \in S} p_{\text{zs}}(\mathcal{A}_i(x)) \right),
\end{equation}
The adapted prediction $p_{\text{adapt}}(x)$ is then obtained from the original view $\mathcal{A}_0(x)$ using the updated model.
However, when predictions in $S$ are consistent with $p_{\text{zs}}(\mathcal{A}_0(x))$, they provide little informative supervision for model updates. 
We therefore argue that the similarity among selected predictions is closely related to the effectiveness of subsequent adaptation.
To measure this, we define a prediction consistency score $\alpha_\text{Con}(x)=\sum_{i \in S} \mathbb{I}( p_{\text{zs}}(\mathcal{A}_i(x)) = p_{\text{zs}}(\mathcal{A}_0(x)))$, which measures how many augmented views produce the same prediction as the original view.
Higher consistency suggests that the prediction is already stable across different views, implying limited benefit from further adaptation.
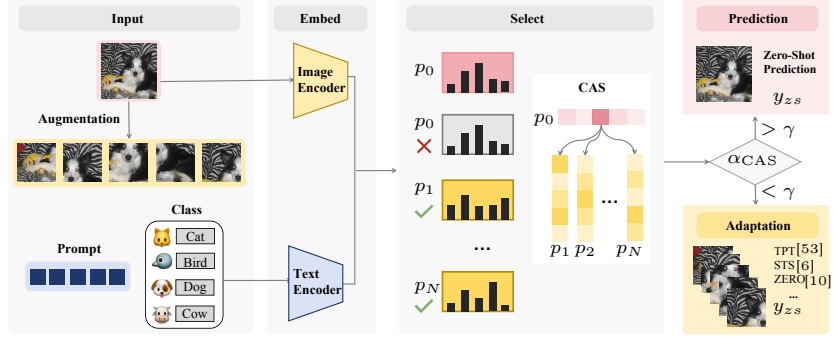
\begin{figure}[t]
\centering
\begin{overpic}[scale=0.14]{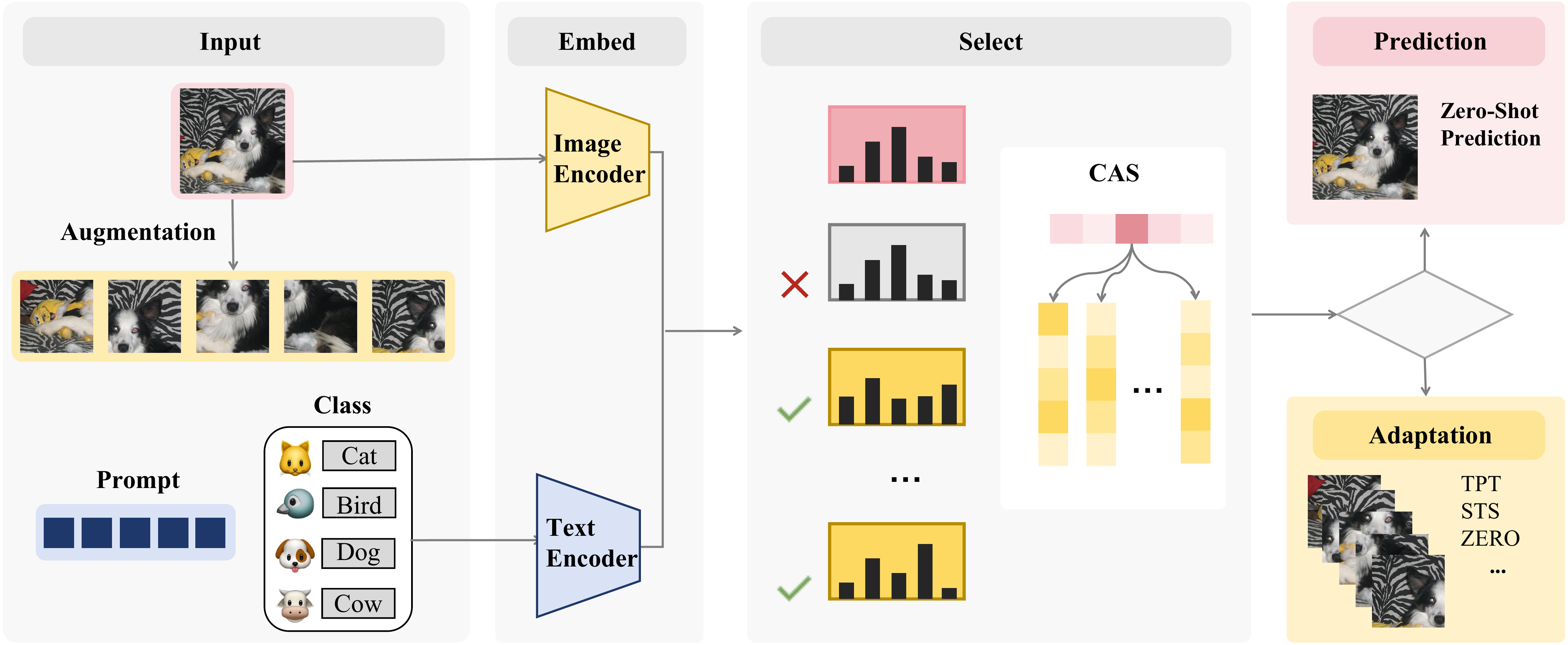}
    \put(49.5,32){\scriptsize{$p_0$}}
    \put(49.5,25.5){\scriptsize{$p_0$}}
    \put(49.5,17.5){\scriptsize{$p_1$}}
    \put(49.5,5.5){\scriptsize{$p_N$}}

    \put(64,26){\scriptsize{$p_0$}}
    \put(66,10){\scriptsize{$p_1$}}
    \put(69,10){\scriptsize{$p_2$}}
    \put(74,10){\scriptsize{$p_N$}}

    \put(87.5,21){\scriptsize{$\alpha_\text{CAS}$}}

    \put(93,28.5){\scriptsize{$y_{zs}$}}
    \put(93,3){\scriptsize{$y_{zs}$}}

    \put(95,10){\tiny{~\cite{shu2022testtime}}}
    \put(95,8){\tiny{~\cite{dafnis2025testtime}}}
    \put(96,6.2){\tiny{~\cite{farina2024frustratingly}}}

    \put(91.5,17){\scriptsize{$<\gamma$}}
    \put(91.5,24.5){\scriptsize{$>\gamma$}}

\end{overpic}
\caption{Pipeline of the proposed selective adaptation framework.
Augmented views and class prompts are encoded to obtain zero-shot predictions. 
CAS measures cross-view prediction agreement.
Samples with low CAS scores undergo adaptation, while those with high scores directly use zero-shot predictions, enabling efficient adaptation.}
\label{fig:zhutu}
\end{figure}

Hard prediction consistency can be improved by cross-augmentation similarity.
Samples may share the same predicted label while exhibiting different probability distributions. 
In such cases, adaptation may still be beneficial and should therefore not be skipped.
To address this issue, we introduce CAS, a scoring function for selective adaptation that jointly considers prediction consistency and distribution similarity:
\begin{equation}
    \alpha_\text{CAS}(x) = \sum_{i \in S} \widetilde{\text{cos}}(p_\text{zs}(\mathcal{A}_i(x)),p_\text{zs}(\mathcal{A}_0(x))) \cdot \mathbb{I}(y_\text{zs}(\mathcal{A}_i(x)) = y_\text{zs}(\mathcal{A}_0(x))).
\end{equation}
For each augmented view $\mathcal{A}_i(x)$, a reweighting factor is computed via cosine similarity and normalized across selected views.
Specifically, letting $p_i$ and $p_0$ denote $p_\text{zs}(\mathcal{A}_i(x))$ and $p_\text{zs}(\mathcal{A}_0(x))$ respectively, the normalized similarity is given by $\widetilde{\cos}(p_i, p_0) = \frac{\cos(p_i, p_0)}{\sum_{j \in S} \cos(p_j, p_0)}$.
As illustrated in Fig.~\ref{fig:zhutu}, augmentations with higher consistency with the original prediction are assigned larger weights, thereby contributing more to the final score.
The pseudo-code is provided in the Appendix.

\begin{table*}[t]
\centering
\caption{Performance comparison of different selection strategies under various TTA methods with ViT-B/16. We report AUC and AEP on ImageNet and its variants. The best results are highlighted in \textbf{bold}.}
\label{tab:main_i}
\small
\setlength{\tabcolsep}{1.8pt}
\resizebox{\linewidth}{!}{
\begin{tabular}{@{}llcccccccccccc@{}}
\toprule
\multirow{2}{*}{Method} & \multirow{2}{*}{Strategy} & \multicolumn{2}{c}{ImageNet} & \multicolumn{2}{c}{ImageNet-A} & \multicolumn{2}{c}{ImageNet-V} & \multicolumn{2}{c}{ImageNet-R} & \multicolumn{2}{c}{ImageNet-K} & \multicolumn{2}{c}{Avg.} \\ 
\cmidrule(lr){3-4} \cmidrule(lr){5-6} \cmidrule(lr){7-8} \cmidrule(lr){9-10} \cmidrule(lr){11-12} \cmidrule(l){13-14} 
& & AUC{$\uparrow$}&AEP{$\uparrow$}&AUC{$\uparrow$}&AEP{$\uparrow$}&AUC{$\uparrow$}&AEP{$\uparrow$}&AUC{$\uparrow$}&AEP{$\uparrow$}&AUC{$\uparrow$}&AEP{$\uparrow$}&AUC{$\uparrow$}&AEP{$\uparrow$}\\ 
\midrule
\multirow{4}{*}{TPT~\cite{shu2022testtime}} & Random & 50.77 & 68.19 & 51.08 & 52.54 & 51.38 & 62.62 & 48.54 & 75.97 & 49.46 & 47.23 & 50.25 & 61.31 \\
& Energy~\cite{liu2020energy} & 57.27 & 68.41 & 52.50 & 52.70 & 57.70 & 62.95 & 64.31 & 76.64 & 54.04 & 47.41 & 57.16 & 61.62 \\
& MCM~\cite{ming2022delving} & 63.82 & 68.50 & 53.73 & 52.73 & 61.44 &62.95 & 71.09 & 76.77 &54.33 & 47.36 & 60.88& 61.66 \\
& \cellcolor{casbg}CAS & \cellcolor{casbg}\textbf{91.10} & \cellcolor{casbg}\textbf{69.00} & \cellcolor{casbg}\textbf{88.62} & \cellcolor{casbg}\textbf{54.70} & \cellcolor{casbg}\textbf{89.69} & \cellcolor{casbg}\textbf{63.45} & \cellcolor{casbg}\textbf{93.45} & \cellcolor{casbg}\textbf{77.14} & \cellcolor{casbg}\textbf{84.24} & \cellcolor{casbg}\textbf{47.93} & \cellcolor{casbg}\textbf{89.42} & \cellcolor{casbg}\textbf{62.44} \\
\midrule
\multirow{4}{*}{R-TPT~\cite{sheng2025illusion}} & Random & 50.63 & 68.50 & 50.67 & 54.55 & 51.71 & 63.09 & 48.27 & 75.87 & 49.85 & 47.14 & 50.23 & 61.83 \\
& Energy~\cite{liu2020energy} & 58.17 & 68.78 & 52.22 & 54.82 & 58.69 & 63.45 & 64.03 & 76.56 & 54.11 & 47.35 & 57.44 & 62.19 \\
& MCM~\cite{ming2022delving} & 65.35 & 68.90 & 53.33& 54.75 & 63.44 &63.46& 71.37 & 76.68 & 55.17 & 47.27 & 61.73 & 62.21 \\
& \cellcolor{casbg}CAS & \cellcolor{casbg}\textbf{91.54} & \cellcolor{casbg}\textbf{69.45} & \cellcolor{casbg}\textbf{89.77} & \cellcolor{casbg}\textbf{57.76} & \cellcolor{casbg}\textbf{89.55} & \cellcolor{casbg}\textbf{64.03} & \cellcolor{casbg}\textbf{93.52} & \cellcolor{casbg}\textbf{77.13} & \cellcolor{casbg}\textbf{84.86} & \cellcolor{casbg}\textbf{48.02} & \cellcolor{casbg}\textbf{89.85} & \cellcolor{casbg}\textbf{63.28} \\
\midrule
\multirow{4}{*}{STS~\cite{dafnis2025testtime}} & Random & 50.23 & 68.17 & 50.84 & 57.21 & 51.41 & 63.17 & 48.30 & 75.94 & 50.11 & 47.45 & 50.18 & 62.39 \\
& Energy~\cite{liu2020energy} & 58.21 & 68.41 & 53.69 & 57.78 & 59.03 & 63.58 & 64.71 & 76.78 & 53.78 &47.55 & 57.88 & 62.82 \\
& MCM~\cite{ming2022delving}  & 65.62 & 68.45 & 55.68 & 57.75 & 64.29 & 63.55 & 71.67 & 76.78 & 54.17 & 47.38 & 62.29 & 62.78 \\
& \cellcolor{casbg}CAS & \cellcolor{casbg}\textbf{94.51} & \cellcolor{casbg}\textbf{68.89} & \cellcolor{casbg}\textbf{91.02} & \cellcolor{casbg}\textbf{61.23} & \cellcolor{casbg}\textbf{93.03} & \cellcolor{casbg}\textbf{64.14} & \cellcolor{casbg}\textbf{94.96} & \cellcolor{casbg}\textbf{77.11} & \cellcolor{casbg}\textbf{88.10} & \cellcolor{casbg}\textbf{48.14} & \cellcolor{casbg}\textbf{92.32} & \cellcolor{casbg}\textbf{63.90} \\
\midrule
\multirow{4}{*}{ZERO~\cite{farina2024frustratingly}} & Random & 51.08 & 68.57 & 50.71 & 55.92 & 51.42 & 63.22 & 48.59 & 76.10 & 49.80 & 47.69 & 50.32 & 62.30 \\
& Energy~\cite{liu2020energy} & 57.04 & 68.72 & 51.89 & 56.29 & 57.66 & 63.53 & 63.79 & 76.78 & 54.09 & 47.83 & 56.89 & 62.63 \\
& MCM ~\cite{ming2022delving} & 64.22 & 68.83 & 52.92 & 56.16 & 62.71 & 63.57 & 70.64 & 76.86 & 54.82 & 47.74 & 61.06 & 62.63 \\
& \cellcolor{casbg}CAS & \cellcolor{casbg}\textbf{94.58} & \cellcolor{casbg}\textbf{69.36} & \cellcolor{casbg}\textbf{91.71} & \cellcolor{casbg}\textbf{59.58} & \cellcolor{casbg}\textbf{93.35} & \cellcolor{casbg}\textbf{64.22} & \cellcolor{casbg}\textbf{95.05} & \cellcolor{casbg}\textbf{77.27} & \cellcolor{casbg}\textbf{88.41} & \cellcolor{casbg}\textbf{48.50} & \cellcolor{casbg}\textbf{92.62} & \cellcolor{casbg}\textbf{63.79} \\
\bottomrule
\end{tabular}
}
\end{table*}

\section{Experiment}

\begin{table*}[!ht]
\centering
\caption{Performance comparison of different selection strategies under various TTA methods with ViT-B/16. We report AUC and AEP on fine-grained datasets. The best results are highlighted in \textbf{bold}.}
\label{tab:main_fg}
\small
\setlength{\tabcolsep}{1.8pt} 
\resizebox{1\linewidth}{!}{
\begin{tabular}{@{}llccccccccccccc@{}}
\toprule
Method & Strategy &Metric& Flow. & DTD & Pets & UCF & Cal. & Air. & Euro. & Cars & Food & SUN & Avg.\\ \midrule
\multirow{8}{*}{TPT~\cite{shu2022testtime}} 
& \multirow{2}{*}{Random} & AUC & 49.04 & 49.06 & 55.59 & 49.33 & 46.03 & 53.72 & 49.55 & 49.94 & 50.63 & 50.12 & 50.30 \\
& &AEP & 68.23 & 45.98 &\textbf{ 87.60 }& 67.08 & 94.12 & \textbf{23.89} & 42.55 & 66.28 & 84.35 & 64.48 & 64.46 \\
\cmidrule(lr){2-14}
& \multirow{2}{*}{Energy~\cite{liu2020energy}} & AUC &63.48 & 53.52 & 69.46 & 60.01 & 63.46 & 46.97 & 75.99 & 54.58 & 69.82 & 59.56 & 61.69 \\
&  & AEP &
68.42 & 46.26 & 87.36 & 67.62 & \textbf{94.39 }& 23.36 & \textbf{43.97} & 66.28 & 84.54 & 64.95 & 64.72\\
\cmidrule(lr){2-14}
& \multirow{2}{*}{MCM~\cite{ming2022delving}} & AUC & 68.47	& 69.48	& 75.21	& 65.47	& 73.69	& 47.83	& 70.17	& 63.01	& 79.74	& 65.92& 	67.90\\
&  & AEP & 68.53& 46.78	& 87.30	& 67.64	& 94.33	& 23.09	& 43.38	& 66.47	& 84.60	& 65.08	& 64.72 \\

\cmidrule(lr){2-14}
& \cellcolor{casbg} & \cellcolor{casbg}AUC & \cellcolor{casbg}\textbf{89.64} & \cellcolor{casbg}\textbf{86.47} & \cellcolor{casbg}\textbf{95.61} & \cellcolor{casbg}\textbf{89.67} & \cellcolor{casbg}\textbf{94.42} & \cellcolor{casbg}\textbf{65.06} & \cellcolor{casbg}\textbf{82.57} & \cellcolor{casbg}\textbf{89.45} & \cellcolor{casbg}\textbf{96.50} & \cellcolor{casbg}\textbf{90.34} & \cellcolor{casbg}\textbf{87.97} \\
& \cellcolor{casbg} \multirow{-2}{*}{{CAS}} & \cellcolor{casbg}AEP & \cellcolor{casbg}\textbf{68.65 }& \cellcolor{casbg}\textbf{47.25} & \cellcolor{casbg}87.30 & \cellcolor{casbg}\textbf{68.09} & \cellcolor{casbg}94.22 & \cellcolor{casbg}23.86 & \cellcolor{casbg}43.19 & \cellcolor{casbg}\textbf{66.68} & \cellcolor{casbg}\textbf{84.68} & \cellcolor{casbg}\textbf{65.58} & \cellcolor{casbg}\textbf{64.95} \\
\midrule
\multirow{8}{*}{R-TPT~\cite{sheng2025illusion}} 
& \multirow{2}{*}{Random} & AUC & 47.42 & 50.76 & 53.97 & 49.31 & 49.28 & 55.72 & 49.08 & 50.93 & 50.50 & 50.10 & 50.71 \\
&  & AEP & 67.91 & 45.53 &\textbf{ 87.38} & 66.66 & 94.01 & \textbf{24.42} & 36.84 & 66.47 & 84.06 & 64.58 & 63.79 \\
\cmidrule(lr){2-14}
& \multirow{2}{*}{Energy~\cite{liu2020energy}} & AUC &62.55 & 56.79 & 68.78 & 59.34 & 62.93 & 46.64 & 69.58 & 53.17 & 70.36 & 59.60 & 60.97 \\
&  & AEP &
68.14 & 45.72 & 87.02 & 67.15 &\textbf{ 94.26} & 23.67 & 38.26 & 66.29 & 84.20 & 65.05 & 63.98\\
\cmidrule(lr){2-14}
& \multirow{2}{*}{MCM~\cite{ming2022delving}} & AUC &
66.63	&69.75	&73.97	&65.86	&73.75	&48.17	&65.63	&63.70	&80.36	&66.48	&67.43\\
&  & AEP &68.25	&46.10&	86.94&	67.12	&94.14	&23.55	&\textbf{37.49}	&66.54	&84.19	&65.17	&63.95\\
\cmidrule(lr){2-14}
& \cellcolor{casbg} & \cellcolor{casbg}AUC & \cellcolor{casbg}\textbf{89.14} & \cellcolor{casbg}\textbf{87.20} & \cellcolor{casbg}\textbf{97.44} & \cellcolor{casbg}\textbf{89.31} & \cellcolor{casbg}\textbf{95.35} & \cellcolor{casbg}\textbf{64.73} & \cellcolor{casbg}\textbf{78.31} & \cellcolor{casbg}\textbf{90.12} & \cellcolor{casbg}\textbf{96.47} & \cellcolor{casbg}\textbf{90.59} & \cellcolor{casbg}\textbf{87.87} \\
& \cellcolor{casbg}\multirow{-2}{*}{{CAS}} & \cellcolor{casbg}AEP & \cellcolor{casbg}\textbf{68.48} & \cellcolor{casbg}\textbf{46.66} & \cellcolor{casbg}86.92 & \cellcolor{casbg}\textbf{67.52} & \cellcolor{casbg}94.01 & \cellcolor{casbg}24.20 & \cellcolor{casbg}37.26 & \cellcolor{casbg}\textbf{66.88} & \cellcolor{casbg}\textbf{84.26} & \cellcolor{casbg}\textbf{65.71} & \cellcolor{casbg}\textbf{64.19} \\
\midrule
\multirow{8}{*}{STS~\cite{dafnis2025testtime}} 
& \multirow{2}{*}{Random} & AUC &49.91 & 49.81 & 52.74 & 48.41 & 51.65 & 54.23 & 50.33 & 51.31 & 50.16 & 50.18 & 50.87 \\
&  & AEP & \textbf{66.37} & 45.40 &\textbf{ 87.14 }& 65.93 & 93.87 & \textbf{24.59} & 39.41 & 66.82 & \textbf{83.34} & 64.12 & 63.70 \\ \cmidrule(lr){2-14}
& \multirow{2}{*}{Energy~\cite{liu2020energy}} & AUC &60.70 & 58.16 & 68.22 & 57.45 & 63.41 & 47.38 & 69.11 & 53.17 & 70.39 & 59.50 & 60.75 \\
&  & AEP &
66.10 & 45.79 & 86.71 & 66.41 &\textbf{ 94.05} & 24.32 & \textbf{39.86} & 66.63 & 83.31 & 64.47 & 63.77\\
\cmidrule(lr){2-14}
& \multirow{2}{*}{MCM~\cite{ming2022delving}} & AUC &64.91&	69.78	&74.22&	65.83&	76.92	&48.47&	66.03	&63.23	&79.69	&66.54	&67.56\\
&  & AEP & 66.05&	45.96&	86.60&	66.32	&93.93	&24.18	&39.57	&66.91	&83.17	&64.54	&63.72 \\
\cmidrule(lr){2-14}
& \cellcolor{casbg} & \cellcolor{casbg} AUC & \cellcolor{casbg}\textbf{93.07} & \cellcolor{casbg}\textbf{89.97} & \cellcolor{casbg}\textbf{97.68} & \cellcolor{casbg}\textbf{93.24} & \cellcolor{casbg}\textbf{98.94} & \cellcolor{casbg}\textbf{76.44} & \cellcolor{casbg}\textbf{81.97} & \cellcolor{casbg}\textbf{92.29} & \cellcolor{casbg}\textbf{96.85} & \cellcolor{casbg}\textbf{93.70} & \cellcolor{casbg}\textbf{91.42} \\
& \cellcolor{casbg}\multirow{-2}{*}{{CAS}} & \cellcolor{casbg}AEP & \cellcolor{casbg}66.03 & \cellcolor{casbg}\textbf{46.18} & \cellcolor{casbg}86.49 & \cellcolor{casbg}\textbf{66.59} & \cellcolor{casbg}93.63 & \cellcolor{casbg}24.56 & \cellcolor{casbg}39.26 & \cellcolor{casbg}\textbf{67.20} & \cellcolor{casbg}83.17 & \cellcolor{casbg}\textbf{64.88} & \cellcolor{casbg}\textbf{63.80} \\
\midrule
\multirow{8}{*}{ZERO~\cite{farina2024frustratingly}} 
& \multirow{2}{*}{Random} & AUC &46.49 & 50.30 & 54.13 & 47.44 & 49.82 & 56.01 & 49.12 & 50.85 & 50.87 & 49.98 & 50.50 \\
&  &AEP & 66.80 & 44.96 & \textbf{87.68} & 65.76 & 93.93 & \textbf{24.90} & 38.73 & 66.89 & 83.71 & 64.59 & 63.80 \\ \cmidrule(lr){2-14}
& \multirow{2}{*}{Energy~\cite{liu2020energy}} & AUC &56.72 & 56.22 & 67.86 & 57.99 & 66.72 & 47.31 & 69.19 & 53.62 & 69.98 & 59.01 & 60.46 \\
&  &AEP & 
66.71 & 45.16 & 87.44 & 66.26 & \textbf{94.23 }& 24.36 & \textbf{39.17 }& 66.71 & \textbf{83.76} & 65.02 & 63.88\\ \cmidrule(lr){2-14}
& \multirow{2}{*}{MCM~\cite{ming2022delving} }& AUC &62.91	&67.47	&73.80	&65.89	&79.23	&48.10	&64.73	&62.39&	79.73&	65.93	&67.02 \\
&  & AEP &66.84	&45.26&	87.35	&66.19&	94.11	&24.41	&38.59	&66.96	&83.70&	65.15&	63.86\\
\cmidrule(lr){2-14}
& \cellcolor{casbg} & \cellcolor{casbg}AUC & \cellcolor{casbg}\textbf{91.95} & \cellcolor{casbg}\textbf{90.93} & \cellcolor{casbg}\textbf{97.58} & \cellcolor{casbg}\textbf{93.37} & \cellcolor{casbg}\textbf{98.29} & \cellcolor{casbg}\textbf{76.71} & \cellcolor{casbg}\textbf{81.48} & \cellcolor{casbg}\textbf{91.60} & \cellcolor{casbg}\textbf{96.87} & \cellcolor{casbg}\textbf{93.75} & \cellcolor{casbg}\textbf{91.25}\\
& \cellcolor{casbg}\multirow{-2}{*}{{CAS}} & \cellcolor{casbg}AEP & \cellcolor{casbg}\textbf{66.99} & \cellcolor{casbg}\textbf{45.46} & \cellcolor{casbg}87.30 & \cellcolor{casbg}\textbf{66.44} & \cellcolor{casbg}93.90 & \cellcolor{casbg}24.81 & \cellcolor{casbg}38.69 & \cellcolor{casbg}\textbf{67.32} & \cellcolor{casbg}83.73 & \cellcolor{casbg}\textbf{65.58} & \cellcolor{casbg}\textbf{64.02} \\
\bottomrule
\end{tabular}
}
\end{table*}

\subsection{Experimental Setup}
\label{sec:4.1}
\textbf{Datasets.}
To comprehensively evaluate the efficiency and performance of CAS, we conduct experiments across a range of benchmarks, including ImageNet and its variants, as well as fine-grained datasets.
We first use ImageNet~\cite{deng2009imagenet} and its four variants: ImageNet-A (natural adversarial examples)~\cite{hendrycks2021natural}, ImageNet-V (re-collected images)~\cite{recht2019imagenet}, ImageNet-R (artistic renditions)~\cite{hendrycks2021many}, and ImageNet-K (sketch-style images with domain shifts)~\cite{wang2019learning}.
We further evaluate our method on fine-grained datasets to assess cross-domain generalization, including Flowers102~\cite{nilsback2008automated}, DTD~\cite{cimpoi2014describing}, Pets~\cite{parkhi2012cats}, UCF101~\cite{soomro2012ucf101}, Caltech101~\cite{fei2004learning}, Aircraft~\cite{maji2013fine}, EuroSAT~\cite{helber2018introducing}, Cars~\cite{krause20133d}, Food101~\cite{bossard2014food}, and SUN397~\cite{xiao2010sun}.
In the episodic TTA setting, no training data is available at test time, and all experiments are conducted strictly in a zero-shot manner.

\textbf{Baselines.}
To validate the generalizability of our method, we integrate CAS into four representative TTA methods: TPT~\cite{shu2022testtime}, ZERO~\cite{farina2024frustratingly}, R-TPT~\cite{sheng2025r}, and STS~\cite{dafnis2025testtime}.
We compare CAS with three sample selection strategies (Random Skipping, Energy~\cite{liu2020energy}, and MCM~\cite{ming2022delving}) to demonstrate its effectiveness.
Random skipping serves as a lower-bound baseline for comparison.
As representative OOD detection approaches, Energy~\cite{liu2020energy} utilizes a logit-based energy score, while MCM~\cite{ming2022delving} measures confidence by evaluating the alignment between visual features and textual concepts.

\textbf{Metrics.}
To comprehensively evaluate the effectiveness of selective adaptation, we introduce two complementary metrics that assess both detection quality and overall performance.
AUC measures the selection strategy’s ability to distinguish between beneficial and ineffective adaptations, independent of the decision threshold. 
Meanwhile, AEP evaluates the strategy's performance across varying skip ratios.
It computes the expected accuracy under a predefined prior distribution of skip ratios (e.g., a triangular prior), thereby reflecting the overall efficiency–accuracy trade-off of the selection strategy.

\textbf{Implementation details.}
We use CLIP-ViT-B/16~\cite{radford2021learning} as the backbone model and follow the standard TPT setting.
The text prompt is initialized with the template “a photo of a”.
For each image, we generate $N=64$ augmented views via AugMix~\cite{hendrycks2019augmix}.
The confidence threshold $\rho$ is set to 0.1 and kept fixed across all datasets.
Experiments are conducted with multiple random seeds.
All baseline results are reproduced following the previous benchmark~\cite{sheng2025illusion}.

\begin{table*}[t]
\centering
\caption{Performance comparison of different selection strategies across various TTA methods with ViT-B/16. 
We report AEP, ECE expectation with triangular prior (EEP), and AUC on ImageNet and its variants.}
\label{tab:ece_i}
\small
\setlength{\tabcolsep}{1.2pt} 
\resizebox{\linewidth}{!}{
\begin{tabular}{ll ccc ccc ccc ccc ccc ccc}
\toprule
& & \multicolumn{3}{c}{{ImageNet}} & \multicolumn{3}{c}{{ImageNet-A}} & \multicolumn{3}{c}{{ImageNet-V}} & \multicolumn{3}{c}{{ImageNet-R}} & \multicolumn{3}{c}{{ImageNet-K}} & \multicolumn{3}{c}{{Avg.}} \\
\cmidrule(lr){3-5} \cmidrule(lr){6-8} \cmidrule(lr){9-11} \cmidrule(lr){12-14} \cmidrule(lr){15-17} \cmidrule(lr){18-20}

\multirow{-2}{*}{Method} & \multirow{-2}{*}{Strategy} & AEP{$\uparrow$} & EEP{$\downarrow$}& AUC{$\uparrow$} & AEP{$\uparrow$} & EEP{$\downarrow$} & AUC{$\uparrow$} & AEP{$\uparrow$} & EEP{$\downarrow$}& AUC{$\uparrow$} &  AEP{$\uparrow$} & EEP{$\downarrow$}& AUC{$\uparrow$} & AEP{$\uparrow$} & EEP{$\downarrow$}& AUC{$\uparrow$} & AEP{$\uparrow$} & EEP{$\downarrow$} & AUC{$\uparrow$}  \\
\midrule
\multirow{4}{*}{TPT~\cite{shu2022testtime}} 
& Random & 68.19 & 7.40 & 50.77 & 52.54 & 12.62 & 51.08 & 62.62 & 8.52 & 51.38 & 75.97 & \textbf{3.25} & 48.54 & 47.23 & 12.07 & 49.46 & 61.31 & 8.77 & 50.25 \\
& Energy~\cite{liu2020energy} & 68.41 & 7.68 & 57.27 & 52.70 & 13.23 & 52.50 & 62.95 & 9.01 & 57.70 & 76.64 & 3.27& 64.31 & 47.41 & 12.12 & 54.04 & 61.62 & 9.06& 57.16 \\
& MCM~\cite{ming2022delving}& 68.50& 7.91 & 63.82 & 52.73 & 13.66 & 53.73 &62.95 & 9.15 & 61.44 & 76.77 & 3.54 & 71.09 & 47.36 & 12.17 & 54.33 & 61.66& 9.29 & 60.88 \\
& \cellcolor{casbg}CAS & \cellcolor{casbg}\textbf{69.00} & \cellcolor{casbg}\textbf{7.33} & \cellcolor{casbg}\textbf{91.10} & \cellcolor{casbg}\textbf{54.70} & \cellcolor{casbg}\textbf{12.02} & \cellcolor{casbg}\textbf{88.62} & \cellcolor{casbg}\textbf{63.45} & \cellcolor{casbg}\textbf{8.54} & \cellcolor{casbg}\textbf{89.69} & \cellcolor{casbg}\textbf{77.14} & \cellcolor{casbg}3.89 & \cellcolor{casbg}\textbf{93.45} & \cellcolor{casbg}\textbf{47.93} & \cellcolor{casbg}\textbf{11.24} & \cellcolor{casbg}\textbf{84.24} & \cellcolor{casbg}\textbf{62.44} & \cellcolor{casbg}\textbf{8.60} & \cellcolor{casbg}\textbf{89.42} \\
\midrule
\multirow{4}{*}{C-TPT~\cite{yoon2024ctpt}} 
& Random & 67.88 & \textbf{3.88} & 50.83 & 50.14 & 7.95 & 50.04 & 62.03 & \textbf{5.02 }& 51.21 & 75.20 & 2.05 & 49.29 & 47.03 & 8.34 & 50.57 & 60.46 & 5.45 & 50.39 \\
& Energy~\cite{liu2020energy} & 68.14 & 4.26 & 58.26 & 50.66 & 8.20 & 55.21 & 62.31 & 5.42 & 60.21 & 75.63 & \textbf{1.79} & 64.36 & 47.15 & 8.61  & 54.40 & 60.78 & 5.66 & 58.49 \\
& MCM~\cite{ming2022delving}& 68.21 & 4.31 & 64.98 & 50.63 & 8.56 & 56.72 & 62.28 & 5.48 &63.78 & 75.73 & 1.80 & 71.42 & 47.16 &8.62 & 55.57 & 60.80 & 5.75& 62.49 \\
& \cellcolor{casbg}CAS & \cellcolor{casbg}\textbf{68.59} & \cellcolor{casbg}4.01 & \cellcolor{casbg}\textbf{86.50} & \cellcolor{casbg}\textbf{51.74} & \cellcolor{casbg}\textbf{7.53} & \cellcolor{casbg}\textbf{84.29} & \cellcolor{casbg}\textbf{62.68} & \cellcolor{casbg}5.05 & \cellcolor{casbg}\textbf{86.40} & \cellcolor{casbg}\textbf{76.02} & \cellcolor{casbg}2.91 & \cellcolor{casbg}\textbf{90.22} & \cellcolor{casbg}\textbf{47.60} & \cellcolor{casbg}\textbf{7.67} & \cellcolor{casbg}\textbf{80.95} & \cellcolor{casbg}\textbf{61.33} & \cellcolor{casbg}\textbf{5.43} & \cellcolor{casbg}\textbf{85.67} \\
\midrule
\multirow{4}{*}{O-TPT~\cite{sharifdeen2025otpt}} 
& Random & 67.18 &\textbf{ 1.95} & 51.35 & 48.11 & 7.22 & 51.65 & 61.32 & 3.02 & 52.08 & 74.02 & 3.87 & 50.15 & 46.51 & 5.36 & 50.96 & 59.43 & \textbf{4.28} & 51.24 \\
& Energy~\cite{liu2020energy}& 67.28 & 2.05 & 57.35 & 48.26 & 7.05 & 55.41 & 61.35 & \textbf{2.97} & 57.80 & 74.02 & \textbf{3.81} & 62.41 & 46.49 & 5.59 & 52.00 & 59.48 & 4.29 & 56.99 \\
& MCM~\cite{ming2022delving} & 67.35& 2.03&63.50 & 48.22 & 7.18 & 57.28 & 61.32 & \textbf{2.97} & 61.37 & 74.10 & 3.83 & 69.81 & 46.50 & 5.64 & 53.12 & 59.50& 4.33 & 61.02 \\
& \cellcolor{casbg}CAS & \cellcolor{casbg}\textbf{67.70} & \cellcolor{casbg}2.25 & \cellcolor{casbg}\textbf{76.05} & \cellcolor{casbg}\textbf{49.29} & \cellcolor{casbg}\textbf{6.57} & \cellcolor{casbg}\textbf{78.18} & \cellcolor{casbg}\textbf{61.71} & \cellcolor{casbg}3.16 & \cellcolor{casbg}\textbf{75.91} & \cellcolor{casbg}\textbf{74.54} & \cellcolor{casbg}4.24 & \cellcolor{casbg}\textbf{83.31} & \cellcolor{casbg}\textbf{46.93} & \cellcolor{casbg}\textbf{5.27} & \cellcolor{casbg}\textbf{69.70} & \cellcolor{casbg}\textbf{60.03} & \cellcolor{casbg}4.30 & \cellcolor{casbg}\textbf{76.63} \\
\bottomrule
\end{tabular}
}
\end{table*}

\subsection{Results}
\label{sec:4.2}
\textbf{Performance on ImageNet and its variants.}
We evaluate the performance of CAS on ImageNet and its variants. 
As demonstrated in Table~\ref{tab:main_i}, CAS consistently outperforms all competing selection strategies across diverse TTA methods~\cite{shu2022testtime,farina2024frustratingly,dafnis2025testtime,sheng2025r} with respect to both AEP and AUC. 
Specifically, when integrated into the ZERO~\cite{farina2024frustratingly}, CAS achieves state-of-the-art performance on ImageNet, attaining an AEP of 69.36\% and an AUC of 94.58\%. 
The latter represents an outperformance of 30.36\% over the second-best baseline, MCM~\cite{ming2022delving}.
On ImageNet-A, CAS improves the AEP of R-TPT~\cite{sheng2025r} to 57.76\%, validating its superior ability in filtering out ineffective adaptations. 
Across the evaluated datasets, CAS maintains an AUC around 90\%, peaking at 92.62\% when combined with ZERO~\cite{farina2024frustratingly}. 
These results show that CAS effectively distinguishes beneficial from harmful adaptations, achieving a good balance between efficiency and accuracy under distribution shifts.

\textbf{Performance on fine-grained datasets.}
We evaluate CAS on fine-grained benchmarks, with results summarized in Table~\ref{tab:main_fg}. 
Across all evaluated strategies, CAS consistently delivers the highest average performance. 
Under the TPT framework~\cite{shu2022testtime}, CAS achieves a peak average AUC of 87.97\% and an AEP of 64.95\%.
It outperforms MCM~\cite{ming2022delving} and Energy~\cite{liu2020energy} by margins of 20.07\% and 26.28\% in AUC, respectively, demonstrating a superior ability to identify ineffective adaptations. 
Specifically, when integrated into ZERO~\cite{farina2024frustratingly}, CAS demonstrates highly competitive results, achieving an AUC of 98.29\% on Caltech101 and 96.87\% on Food101.

\newcommand{\cellimg}[2][]{\includegraphics[width=0.98\linewidth,#1]{#2}}

\begin{figure*}[t]
\centering
\setlength{\tabcolsep}{2pt}
\renewcommand{\arraystretch}{0.95}
\begin{tabular}{>{\centering\arraybackslash}
p{0.025\textwidth}*{4}{>{\centering\arraybackslash}p{0.235\textwidth}}}

\raisebox{0.5cm}{\rotatebox{90}{\scriptsize ImageNet}}

  & \cellimg{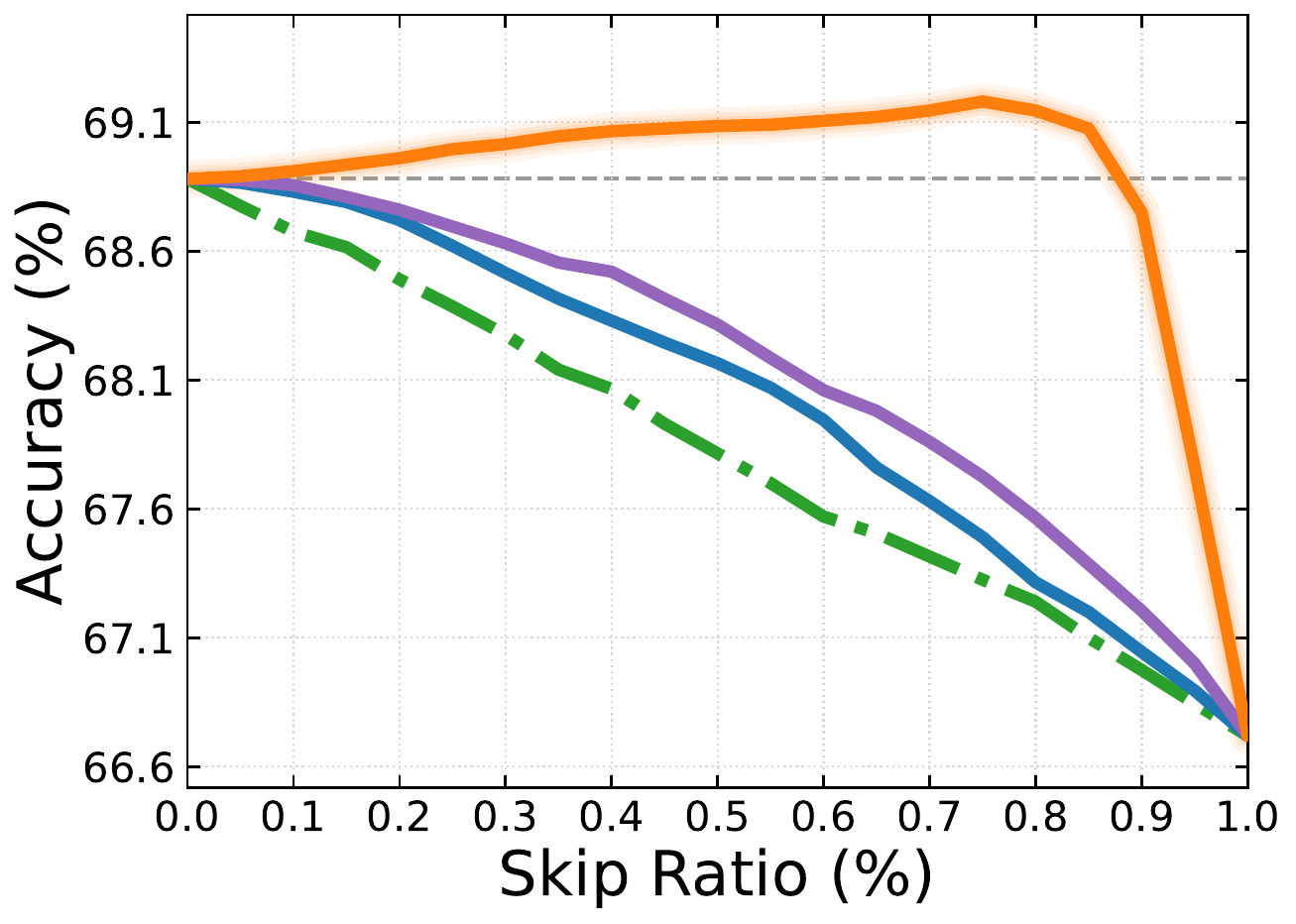}
  & \cellimg{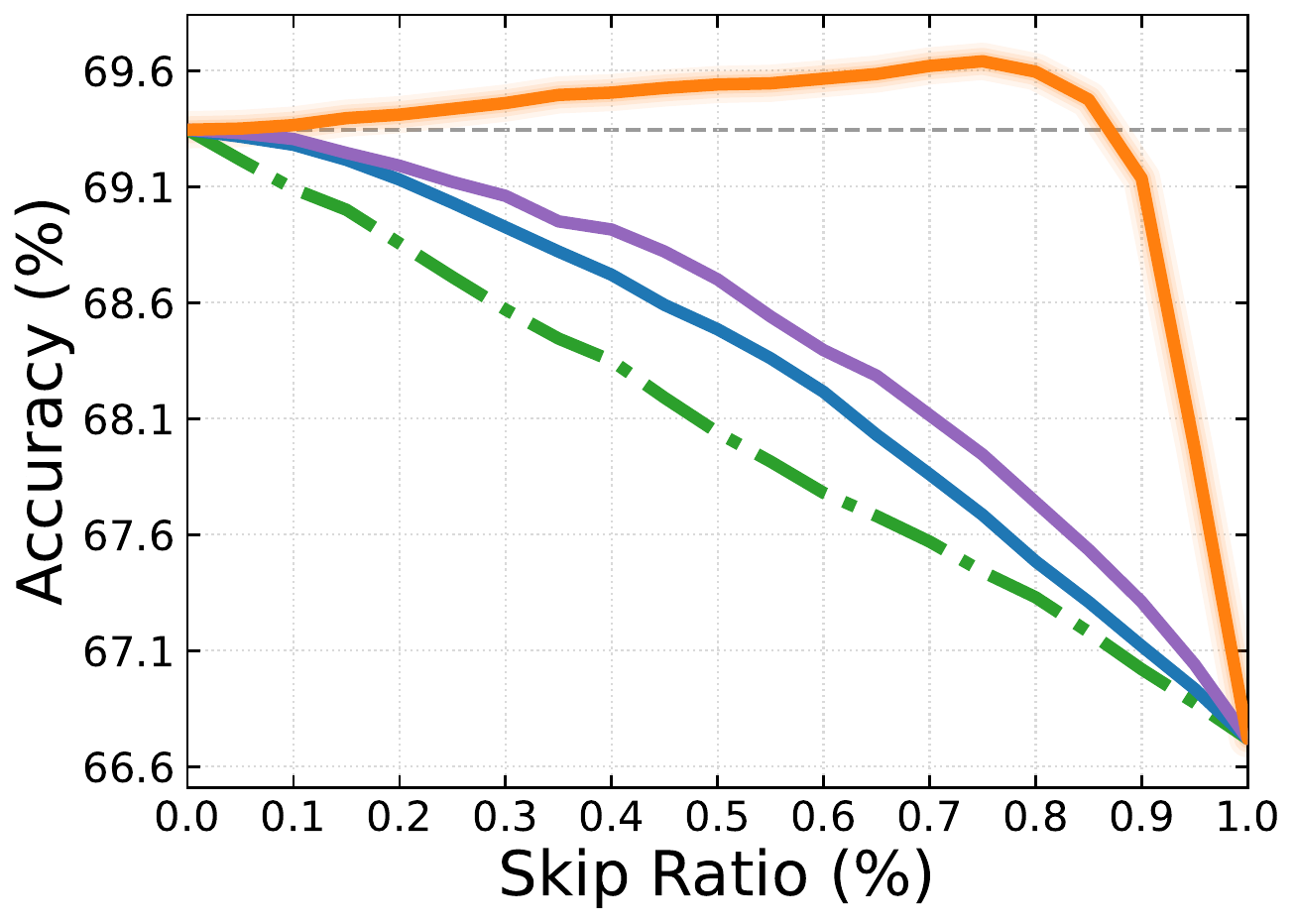}
  & \cellimg{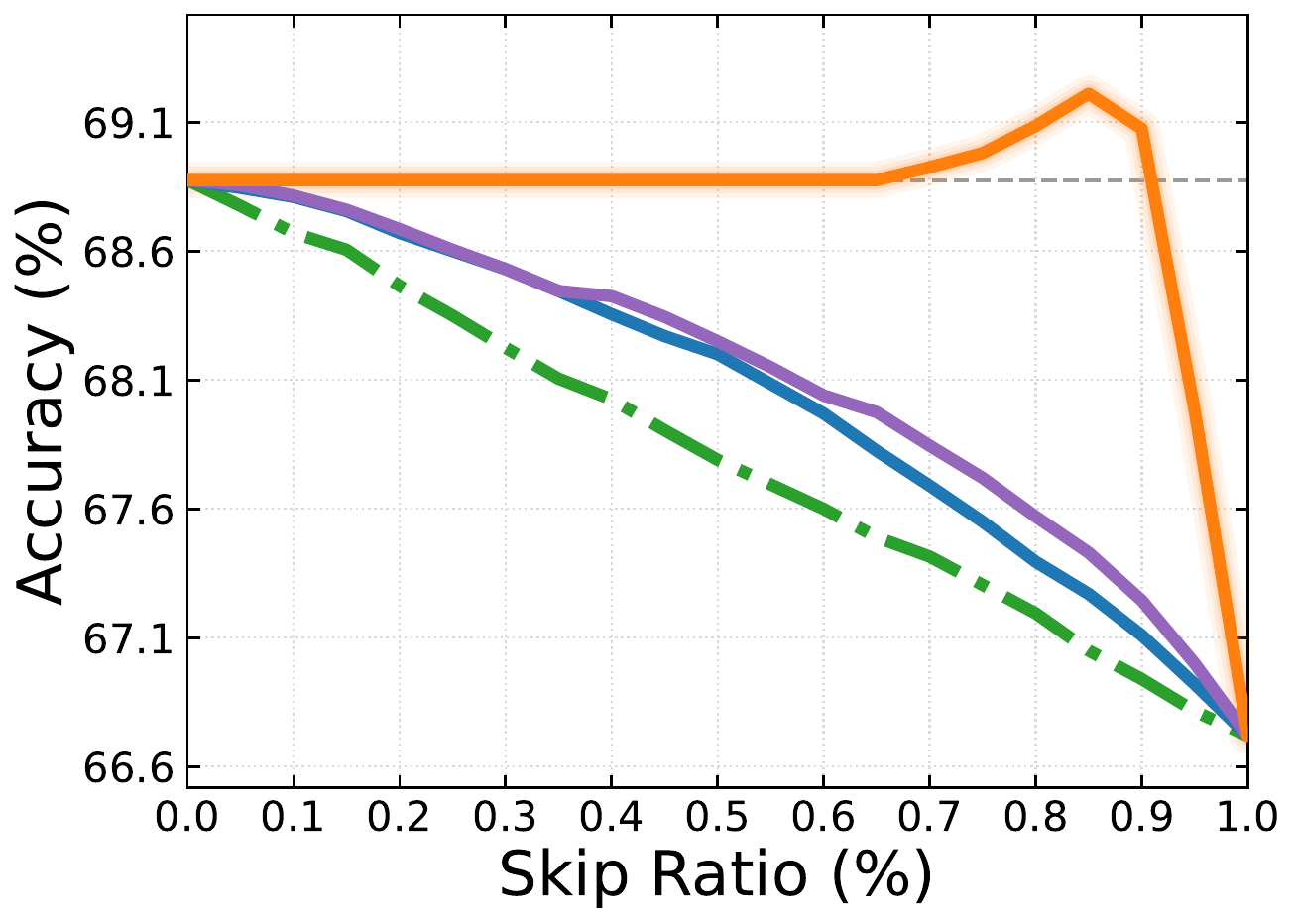}
  & \cellimg{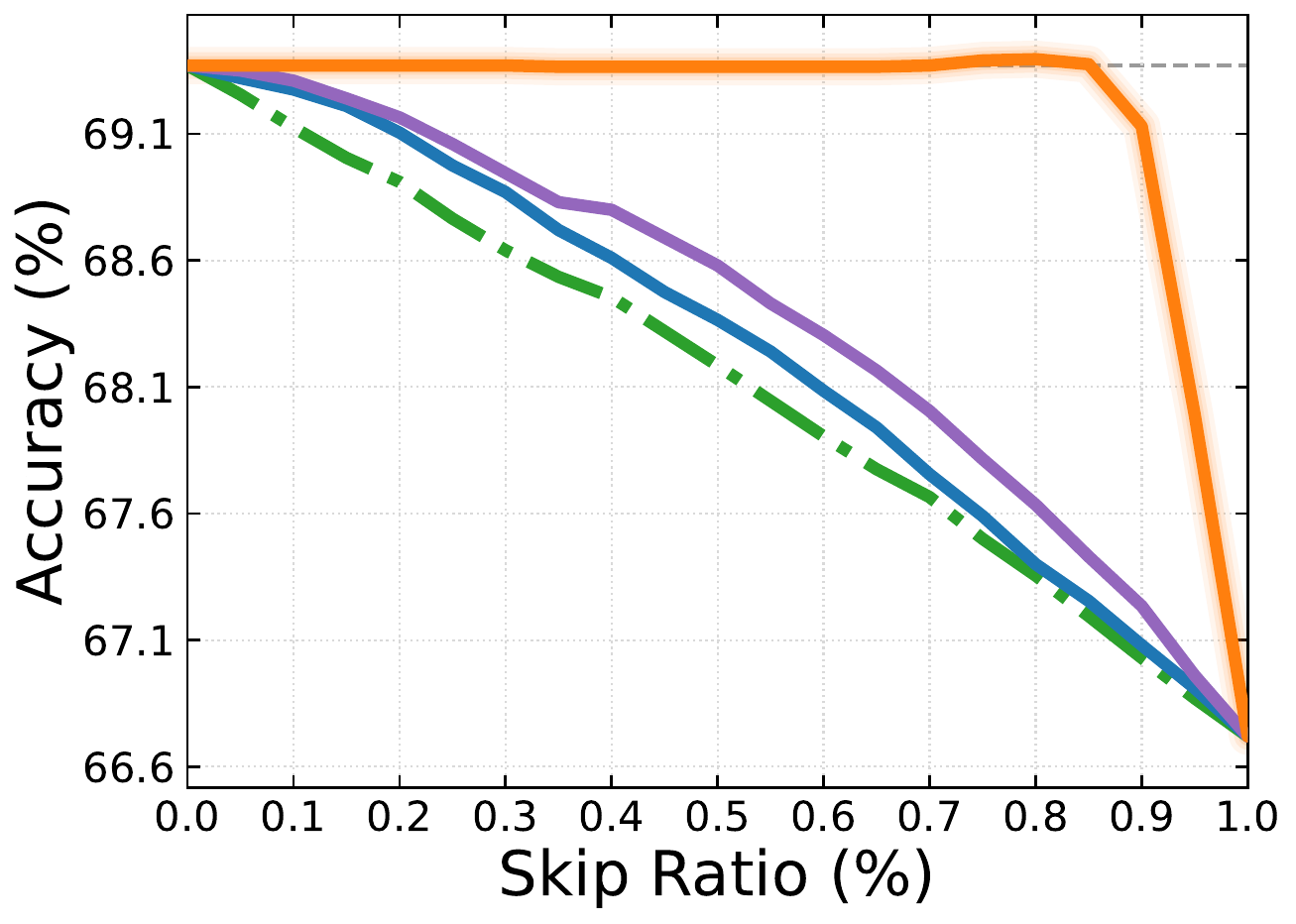} \\[-2pt]

\raisebox{0.3cm}{\rotatebox{90}{\scriptsize ImageNet-A}}
  & \cellimg{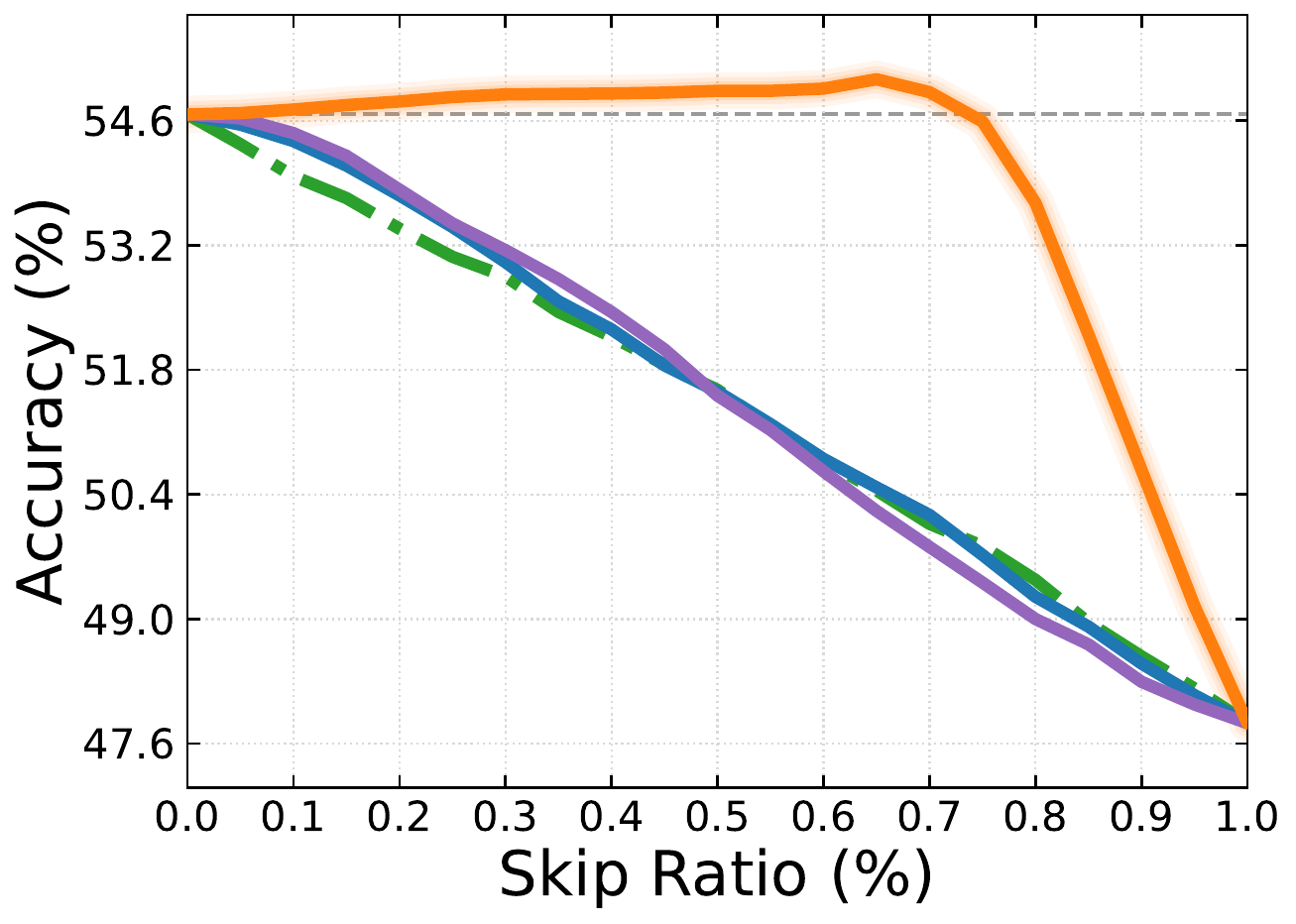}
  & \cellimg{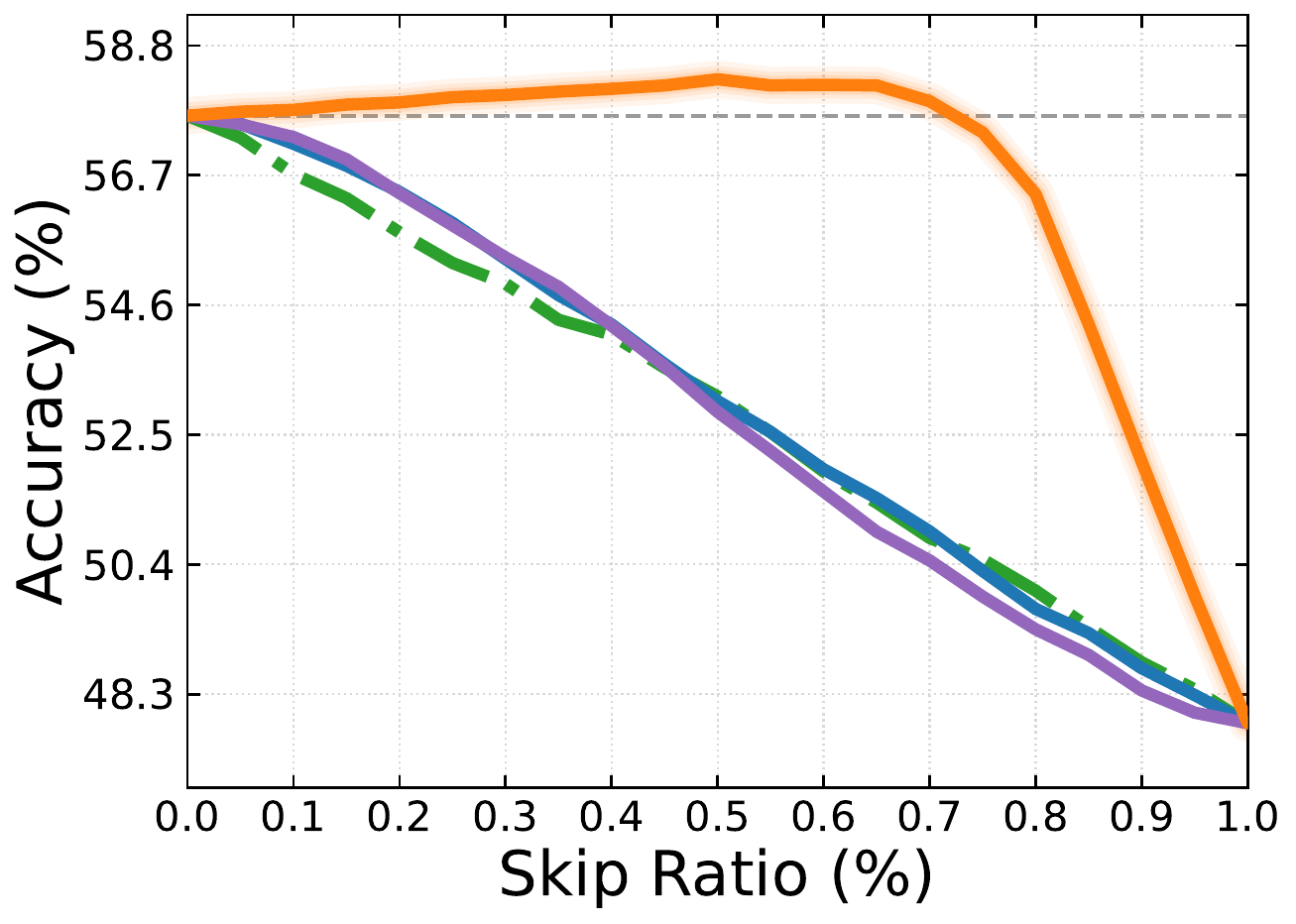}
  & \cellimg{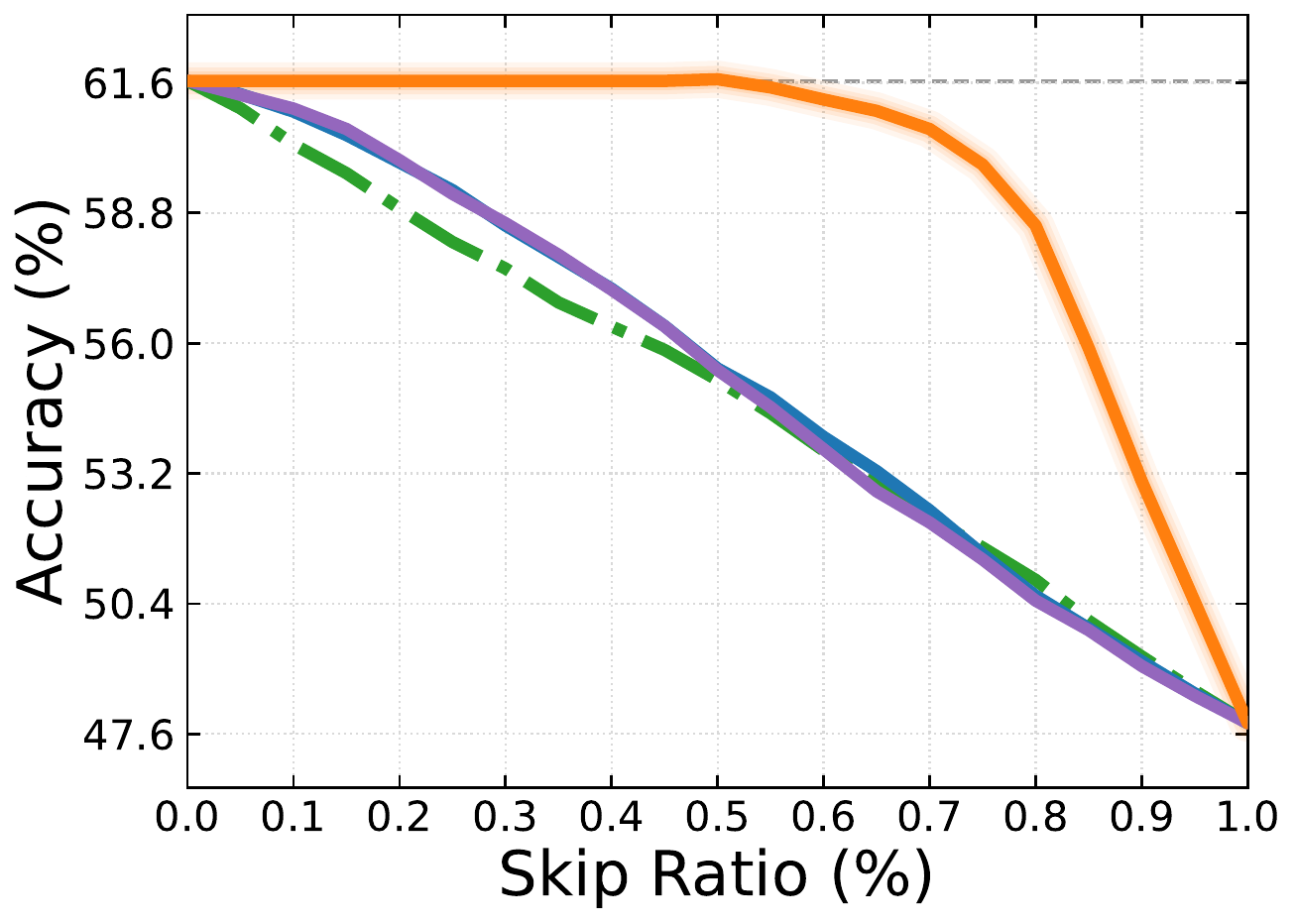}
  & \cellimg{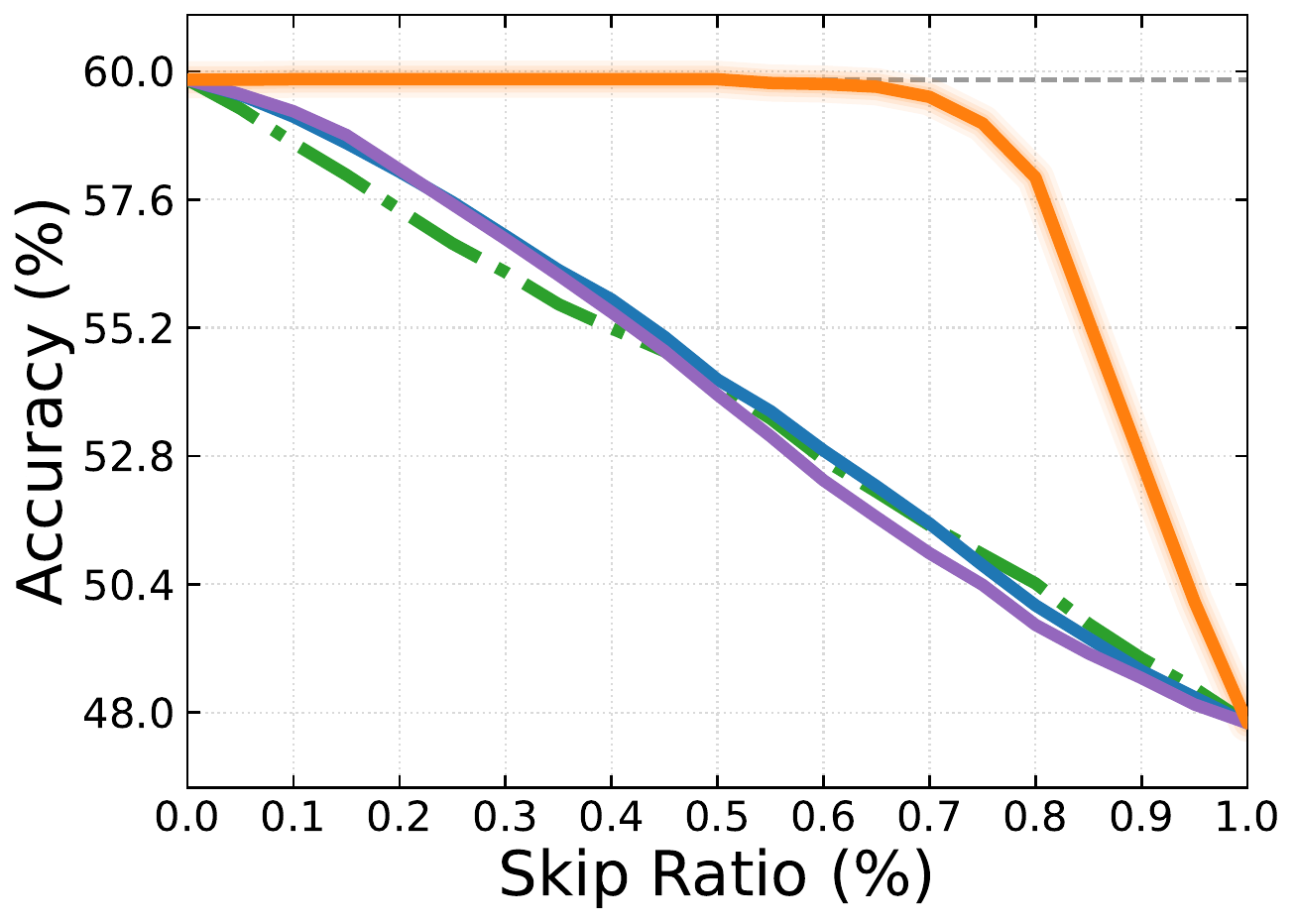} \\[-2pt]

\raisebox{0.8cm}{\rotatebox{90}{\scriptsize DTD}}
  & \cellimg{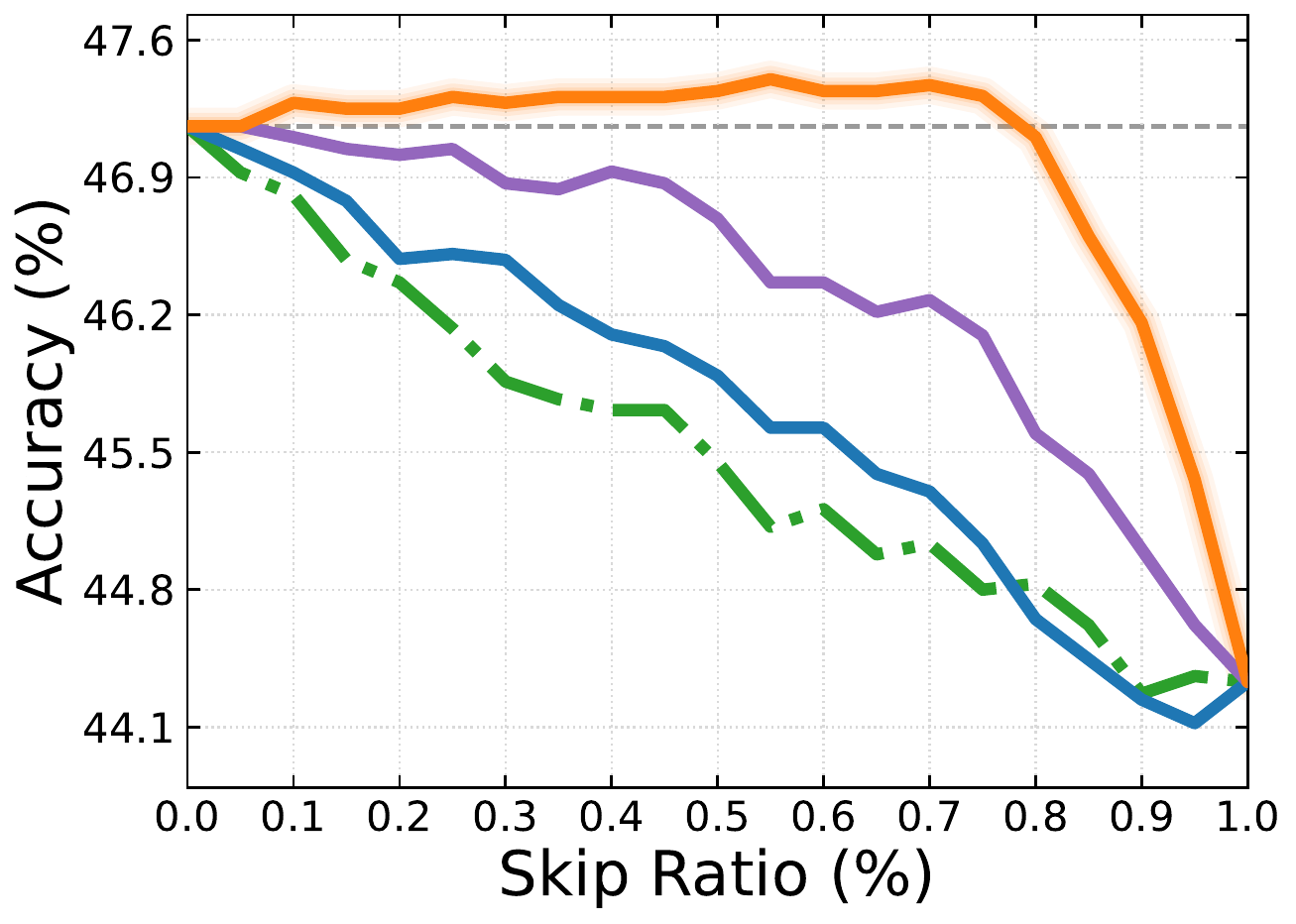}
  & \cellimg{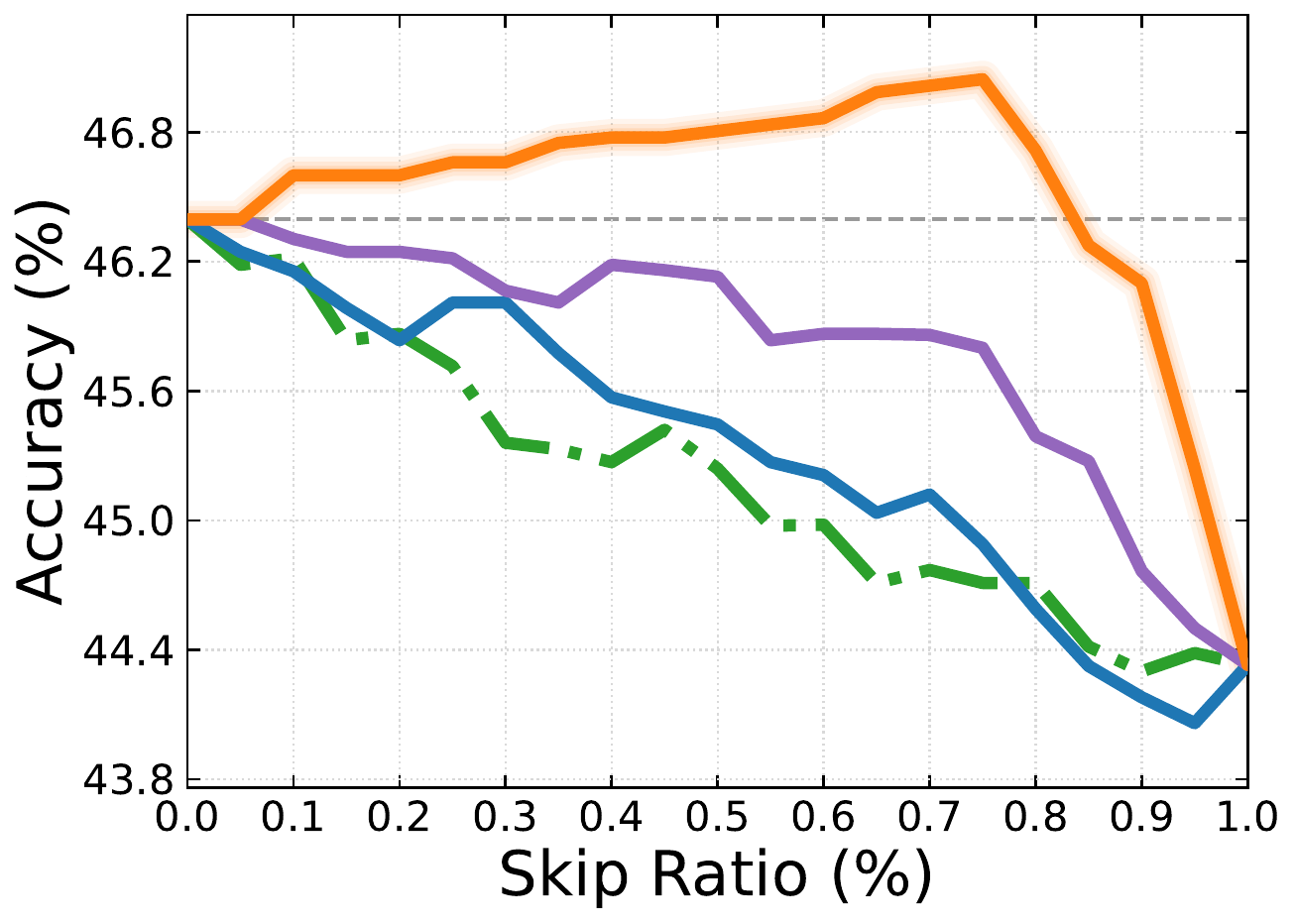}
  & \cellimg{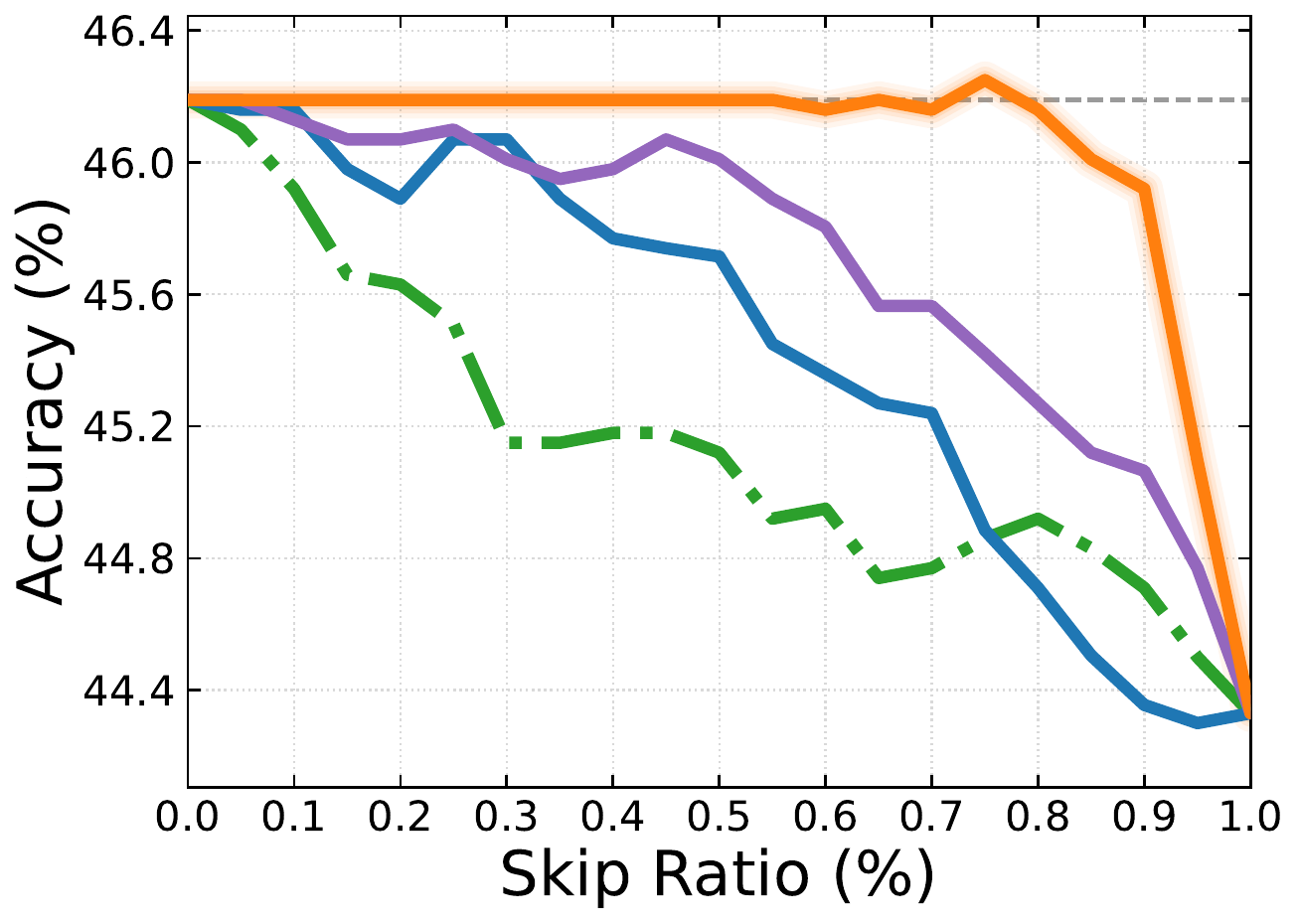}
  & \cellimg{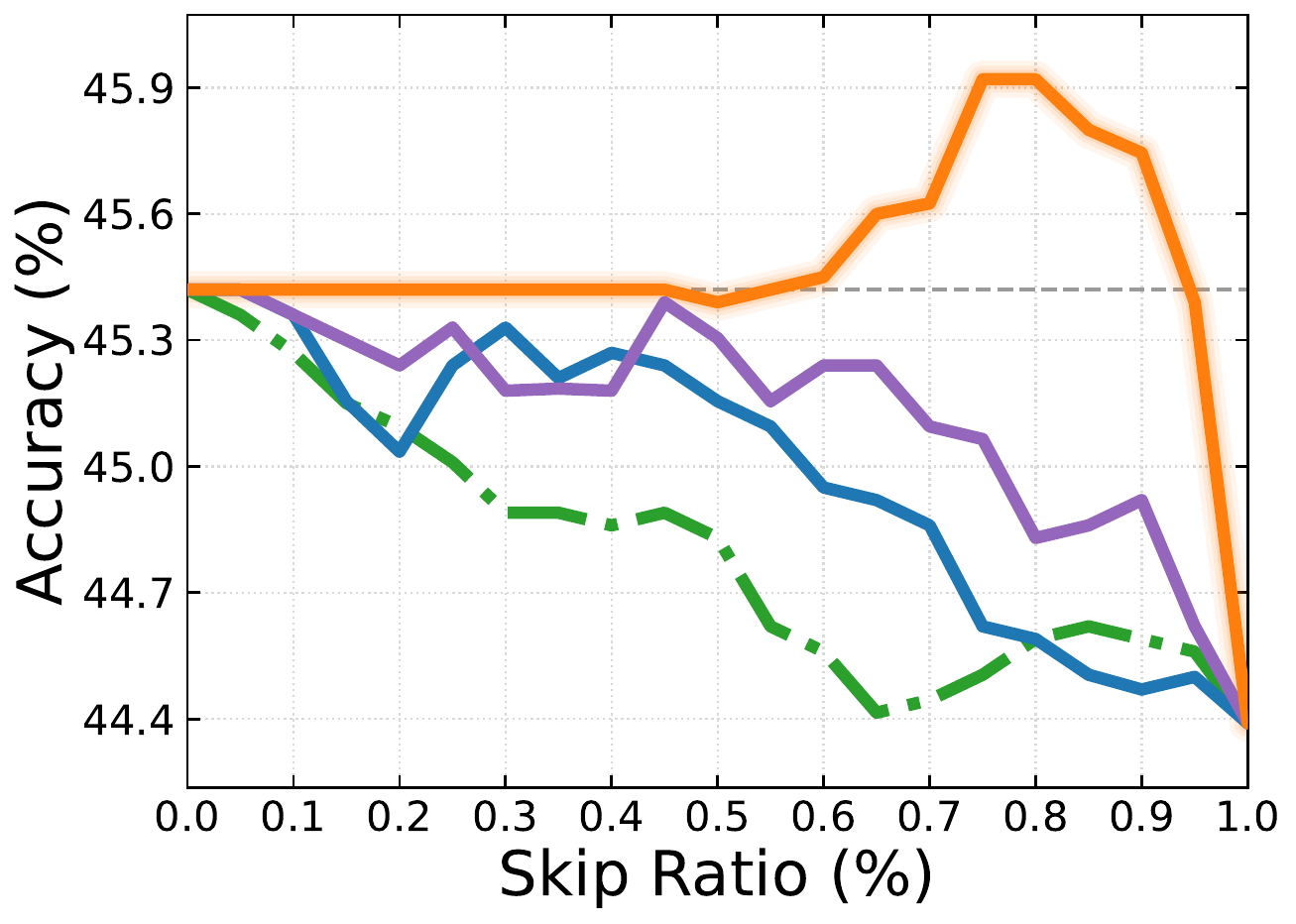} \\[2pt]

  & \cellimg[height=0.45cm,keepaspectratio]{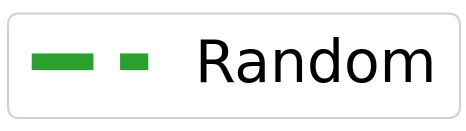}
  & \cellimg[height=0.45cm,keepaspectratio]{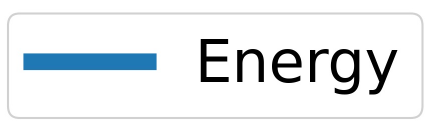}
  & \cellimg[height=0.45cm,keepaspectratio]{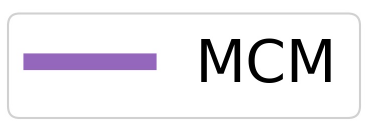}
  & \cellimg[height=0.45cm,keepaspectratio]{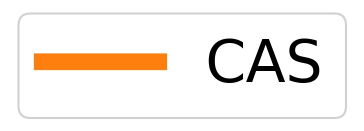} \\[2pt]
  
  & \text{(a) TPT~\cite{shu2022testtime}}
  & \text{(b) R-TPT~\cite{sheng2025r}}
  & \text{(c) STS~\cite{dafnis2025testtime}}
  & \text{(d) ZERO~\cite{farina2024frustratingly}} \\

\end{tabular}

\caption{Accuracy (\%) versus skip ratio (\%) under different skipping strategies across four TTA methods on ImageNet, ImageNet-A, and DTD. CAS consistently maintains higher accuracy even when skipping a large proportion of samples. Compared with other strategies, CAS demonstrates a superior efficiency–accuracy trade-off.}
\label{fig:trade_off}
\end{figure*}

\subsection{Impact on Calibration}
\label{sec:4.3}
While our main results focus on TTA methods~\cite{shu2022testtime,farina2024frustratingly,dafnis2025testtime,sheng2025r} that aim to improve classification accuracy, several prior works emphasize calibration, including C-TPT~\cite{yoon2024ctpt} and O-TPT~\cite{sharifdeen2025otpt}. 
To evaluate the generalization of our approach, we further incorporate CAS into these calibration-oriented methods.
We additionally report the ECE expectation with a triangular prior (EEP), which is computed in the same manner as AEP in Eq.~(\ref{eq:aep}). 
The results are summarized in Table~\ref{tab:ece_i}. 
Across all methods, CAS consistently achieves the highest AEP and AUC, while maintaining lower EEP.
Under TPT~\cite{shu2022testtime}, CAS maintains the lowest average EEP of 8.60\% across ImageNet and its variants.
This indicates that selective skipping guided by CAS does not amplify overconfidence or destabilize prediction margins.
For calibration-oriented methods such as C-TPT~\cite{yoon2024ctpt} and O-TPT~\cite{sharifdeen2025otpt}, CAS continues to generalize effectively.
In C-TPT~\cite{yoon2024ctpt}, CAS achieves the highest average AEP of 61.33\% and AUC of 85.67\% while preserving competitive calibration performance with an EEP of 5.43\%.
Similarly, under O-TPT~\cite{sharifdeen2025otpt}, CAS yields the best AEP of 60.03\% and AUC of 76.63\%, while the other three show comparable performance in EEP.
Overall, these results suggest that our method is not limited to accuracy-oriented TTA methods.
It also generalizes effectively to calibration-oriented methods, consistently improving robustness and reliability without sacrificing calibration performance.

\begin{table*}[t]
\centering
\caption{Ablation study of CAS across various TTA methods on both ImageNet and its variants and fine-grained benchmarks, evaluated using AUC and AEP.}
\label{tab:ablation}
\setlength{\tabcolsep}{6pt} 
\resizebox{\textwidth}{!}{ 
\begin{tabular}{@{}l l cc cc cc cc@{}}
\toprule
\multirow{2}{*}{Dataset} & \multirow{2}{*}{Strategy} 
& \multicolumn{2}{c}{{TPT}~\cite{shu2022testtime}} 
& \multicolumn{2}{c}{{R-TPT}~\cite{sheng2025illusion}}  
& \multicolumn{2}{c}{{STS}~\cite{dafnis2025testtime}} 
& \multicolumn{2}{c}{{ZERO}~\cite{farina2024frustratingly}} \\
\cmidrule(lr){3-4} \cmidrule(lr){5-6} \cmidrule(lr){7-8} \cmidrule(lr){9-10}
& & AUC{$\uparrow$} & AEP{$\uparrow$} & AUC{$\uparrow$} & AEP{$\uparrow$}  & AUC{$\uparrow$} & AEP{$\uparrow$}  & AUC{$\uparrow$} & AEP{$\uparrow$}   \\ 
\midrule

\multirow{4}{*}{
    \begin{tabular}{@{}c@{}} 
        ImageNet \\ \& its \\ variants 
    \end{tabular}
}
& MCM~\cite{ming2022delving} &60.88& 61.66 & 61.73 & 62.21  & 62.29 & 62.78 & 61.06 & 62.63 \\
& Similarity & 82.84 & 62.31 & 85.50 & 63.17  & 88.41 & 63.84 & 87.78 & 63.73 \\
& Consistency & 89.14 & 62.44 & \textbf{89.89} & \textbf{63.28} & \textbf{92.65} & \textbf{63.90} & 92.59 & \textbf{63.79} \\
& \cellcolor{casbg}CAS & \cellcolor{casbg}\textbf{89.42} & \cellcolor{casbg}\textbf{62.44} & \cellcolor{casbg}89.85 & \cellcolor{casbg}\textbf{63.28}  & \cellcolor{casbg}92.32 & \cellcolor{casbg}\textbf{63.90} & \cellcolor{casbg}\textbf{92.62} & \cellcolor{casbg}\textbf{63.79} \\
\midrule
\multirow{4}{*}{Fine-grained}
& MCM~\cite{ming2022delving} & 81.08 & 64.70 & 80.70 & 63.95  & 79.54 & 63.70 & 79.60 & 63.86 \\
& Similarity & 81.04 & 64.81 & 81.44 & 63.93 &  83.61 & 63.60 & 82.91 & 63.79 \\
& Consistency & 87.65 & 64.93 & 87.62 & 64.13 &  91.27 & \textbf{63.84} & 90.82 & 63.97 \\
& \cellcolor{casbg}CAS & \cellcolor{casbg}\textbf{87.97} & \cellcolor{casbg}\textbf{64.95} & \cellcolor{casbg}\textbf{87.87} & \cellcolor{casbg}\textbf{64.19} & \cellcolor{casbg}\textbf{91.42} & \cellcolor{casbg}63.80 & \cellcolor{casbg}\textbf{91.25} & \cellcolor{casbg}\textbf{64.02} \\
\bottomrule
\end{tabular}
}
\end{table*}
\begin{table}[t]
\centering
\caption{Accuracy (\%) of TTA methods on ImageNet with ViT-B/16 at an 85\% skip ratio. Total inference time is reported in hours (h). $\text{CLIP}^*$ denotes CLIP with 64 augmentations.}
\label{tb:speed}
\footnotesize
\setlength{\tabcolsep}{6pt} 
\resizebox{\linewidth}{!}{
\begin{tabular}{lcccccccccc}
\toprule
Metric & \multicolumn{2}{c}{$\text{CLIP}^*$} & \multicolumn{2}{c}{TPT~\cite{shu2022testtime}} & \multicolumn{2}{c}{R-TPT~\cite{sheng2025r}} & \multicolumn{2}{c}{STS~\cite{dafnis2025testtime}} & \multicolumn{2}{c}{ZERO~\cite{farina2024frustratingly}} \\
\cmidrule(lr){2-3} \cmidrule(lr){4-5} \cmidrule(lr){6-7} \cmidrule(lr){8-9} \cmidrule(lr){10-11}
 & Base & CAS & Base & CAS & Base & CAS & Base & CAS & Base & CAS \\
\midrule
Acc. (\%) & 66.72 & -- & 68.88 & \cellcolor{casbg}\textbf{69.03} & 69.36 & \cellcolor{casbg}\textbf{69.43} & 68.83 & \cellcolor{casbg}\textbf{69.20} & 69.28 & \cellcolor{casbg}\textbf{69.32} \\
Time (h)  & 1.65  & -- & 7.76  & \cellcolor{casbg}\textbf{2.55}  & 6.73  & \cellcolor{casbg}\textbf{2.40}  & 2.03  & \cellcolor{casbg}\textbf{1.68}  & 5.42  & \cellcolor{casbg}\textbf{4.96}  \\
\midrule
Speedup  & \multicolumn{2}{c}{--} & \multicolumn{2}{c}{\textbf{3.04$\times$}} & \multicolumn{2}{c}{\textbf{2.81$\times$}} & \multicolumn{2}{c}{\textbf{1.21$\times$}} & \multicolumn{2}{c}{\textbf{1.09$\times$}} \\
\bottomrule
\end{tabular}
}
\end{table}

\subsection{In-depth Analysis}
\label{sec:4.4}
\textbf{Trade-off between efficiency and accuracy.}
To evaluate the trade-off between efficiency and performance, we show the accuracy–skip ratio curves for four TTA methods~\cite{shu2022testtime,farina2024frustratingly,dafnis2025testtime,sheng2025r}, varying the skip ratio $s$ from 0 to 1 with a step size of 0.05.
The results demonstrate that CAS consistently maintains or improves accuracy compared to the full-adaptation baseline ($s=0$), even when skipping up to 85\% of samples.
This suggests that standard TTA may over-adapt certain samples, while CAS selectively identifies and bypasses harmful adaptations, thereby mitigating performance degradation.
Notably, R-TPT~\cite{sheng2025r} achieves an accuracy above 69.60\% when skipping nearly 80\% of samples on the ImageNet dataset, while STS~\cite{dafnis2025testtime} reaches 69.10\% with a skip ratio of 90\%.
In contrast, selection strategies based on Energy~\cite{liu2020energy} and MCM~\cite{ming2022delving} exhibit steep accuracy declines as the skip ratio increases.
In general, Figure~\ref{fig:trade_off} shows that CAS can effectively improve computational overhead without sacrificing accuracy, serving as a good baseline for the selective adaptation problem.

\begin{table*}[t]
\centering
\caption{Performance comparison of different AEP metric functions under TPT~\cite{shu2022testtime} framework. 
We report AUC and AEP on ImageNet and its variants.}
\label{tab:more_f}
\small
\setlength{\tabcolsep}{3pt}
\resizebox{\linewidth}{!}{
\begin{tabular}{@{}llcccccccccccc@{}}
\toprule
{Method} & {Strategy} &{ImageNet} & {ImageNet-A} & {ImageNet-V} &{ImageNet-R} &{ImageNet-K} & {Avg.}  \\

\midrule
\multirow{4}{*}{$f(s)=1$}
& Random &67.83 & 51.41 & 62.23 & 75.43 & 46.93 & 60.77 \\
& Energy~\cite{liu2020energy} &
68.03 & 51.49 & 62.53 & 76.13 & 47.13 & 61.06 \\
& MCM~\cite{ming2022delving} & 68.16 & 51.45 & 62.52 & 76.26 & 47.07 & 61.09 \\
& \cellcolor{casbg}CAS & \cellcolor{casbg}\textbf{68.91} & \cellcolor{casbg}\textbf{53.99} & \cellcolor{casbg}\textbf{63.32} & \cellcolor{casbg}\textbf{76.96} & \cellcolor{casbg}\textbf{47.80} & \cellcolor{casbg}\textbf{62.19}  \\

\midrule
\multirow{4}{*}{$f(s)=2s$}
& Random & 67.47 & 50.27 & 61.85 & 74.88 & 46.63 & 60.22 \\
& Energy~\cite{liu2020energy} & 
67.66 & 50.28 & 62.12 & 75.62 & 46.84 & 60.50\\
& MCM~\cite{ming2022delving} &67.82 & 50.17 & 62.10 & 75.76 & 46.78 & 60.53\\
& \cellcolor{casbg}CAS & \cellcolor{casbg}\textbf{68.81 } & \cellcolor{casbg}\textbf{53.28} & \cellcolor{casbg}\textbf{63.19 } & \cellcolor{casbg}\textbf{76.78 } & \cellcolor{casbg}\textbf{47.66 } & \cellcolor{casbg}\textbf{ 61.94 }  \\
\bottomrule
\end{tabular}
}
\end{table*}
\begin{table*}[t]
\centering
\caption{Performance using resized crops/horizontal flip~\cite{farina2024frustratingly} as data augmentation strategy under ZERO~\cite{farina2024frustratingly} with ViT-B/16. 
We report AUC and AEP on ImageNet and its variants.}
\label{tab:aug_i}
\small
\setlength{\tabcolsep}{1.8pt}
\resizebox{\linewidth}{!}{
\begin{tabular}{@{}llcccccccccccc@{}}
\toprule
\multirow{2}{*}{Method} & \multirow{2}{*}{Strategy} & \multicolumn{2}{c}{ImageNet} & \multicolumn{2}{c}{ImageNet-A} & \multicolumn{2}{c}{ImageNet-V} & \multicolumn{2}{c}{ImageNet-R} & \multicolumn{2}{c}{ImageNet-K} & \multicolumn{2}{c}{Avg.} \\
\cmidrule(lr){3-4} \cmidrule(lr){5-6} \cmidrule(lr){7-8} \cmidrule(lr){9-10} \cmidrule(lr){11-12} \cmidrule(l){13-14}
& & AUC{$\uparrow$} & AEP{$\uparrow$} & AUC{$\uparrow$} & AEP{$\uparrow$} & AUC{$\uparrow$} & AEP{$\uparrow$} & AUC{$\uparrow$} & AEP{$\uparrow$} & AUC{$\uparrow$} & AEP{$\uparrow$} & AUC{$\uparrow$} & AEP{$\uparrow$} \\
\midrule
\multirow{4}{*}{ZERO~\cite{farina2024frustratingly}} 
& Random & 51.18 & 68.50 & 50.66 & 55.91 & 51.33 & 63.18 & 48.78 & 76.15 & 49.99 & 47.70 & 50.39 & 62.29 \\
& Energy~\cite{liu2020energy} & 57.12 & 68.67 & 51.98 & 56.30 & 57.93 & 63.51 & 63.68 & 76.82 & 53.94 &47.81 & 56.93 &62.62 \\
& MCM~\cite{ming2022delving} & 64.30 & 68.77 & 52.96 & 56.15 & 62.89 & 63.52 & 70.56 & 76.93 & 54.80 & 47.72 & 61.10 & 62.62 \\
& \cellcolor{casbg}CAS & \cellcolor{casbg}\textbf{94.59} & \cellcolor{casbg}\textbf{69.27} & \cellcolor{casbg}\textbf{91.89} & \cellcolor{casbg}\textbf{59.54} & \cellcolor{casbg}\textbf{93.33} & \cellcolor{casbg}\textbf{64.16} & \cellcolor{casbg}\textbf{95.08} & \cellcolor{casbg}\textbf{77.34} & \cellcolor{casbg}\textbf{88.37} & \cellcolor{casbg}\textbf{48.49} & \cellcolor{casbg}\textbf{92.65} & \cellcolor{casbg}\textbf{63.76} \\
\bottomrule
\end{tabular}

}
\end{table*}
\begin{figure}[t]
    \centering

    \begin{subfigure}[b]{0.23\textwidth}
        \includegraphics[width=\textwidth]{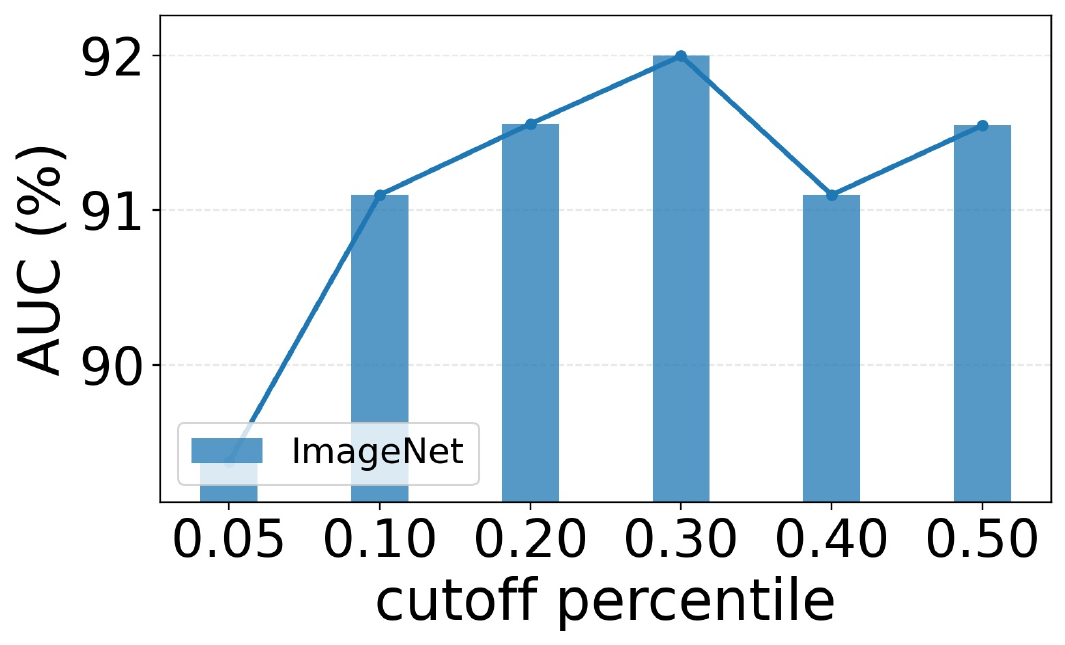}
        \caption{ImageNet}
    \end{subfigure}
    \hfill
    \begin{subfigure}[b]{0.23\textwidth}
        \includegraphics[width=\textwidth]{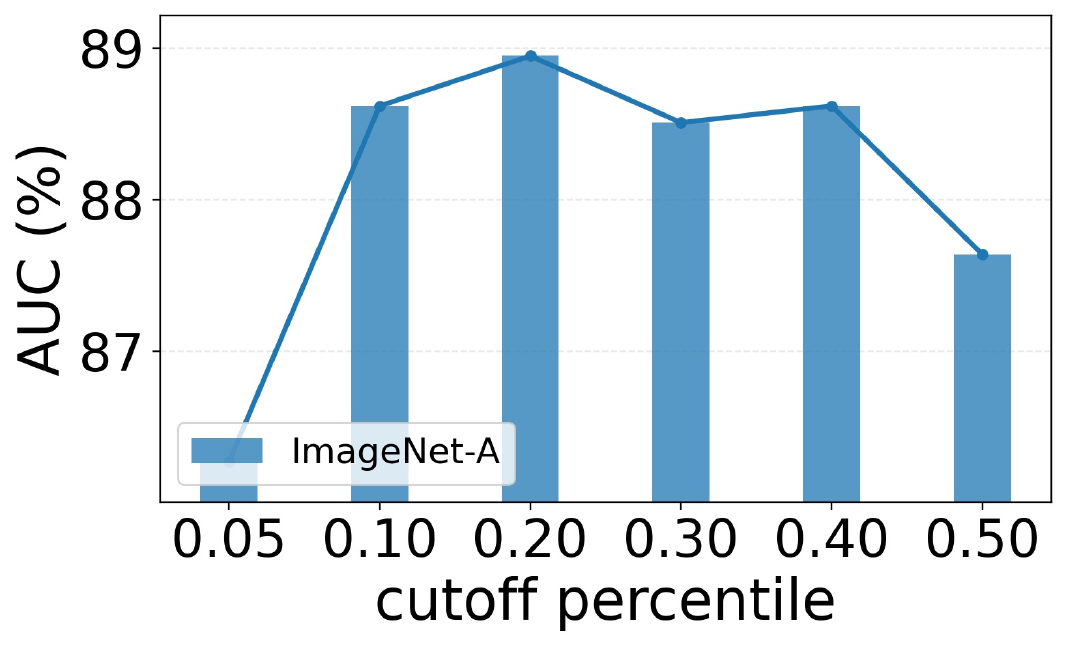}
        \caption{ImageNet-A}
    \end{subfigure}
    \hfill
    \begin{subfigure}[b]{0.23\textwidth}
        \includegraphics[width=\textwidth]{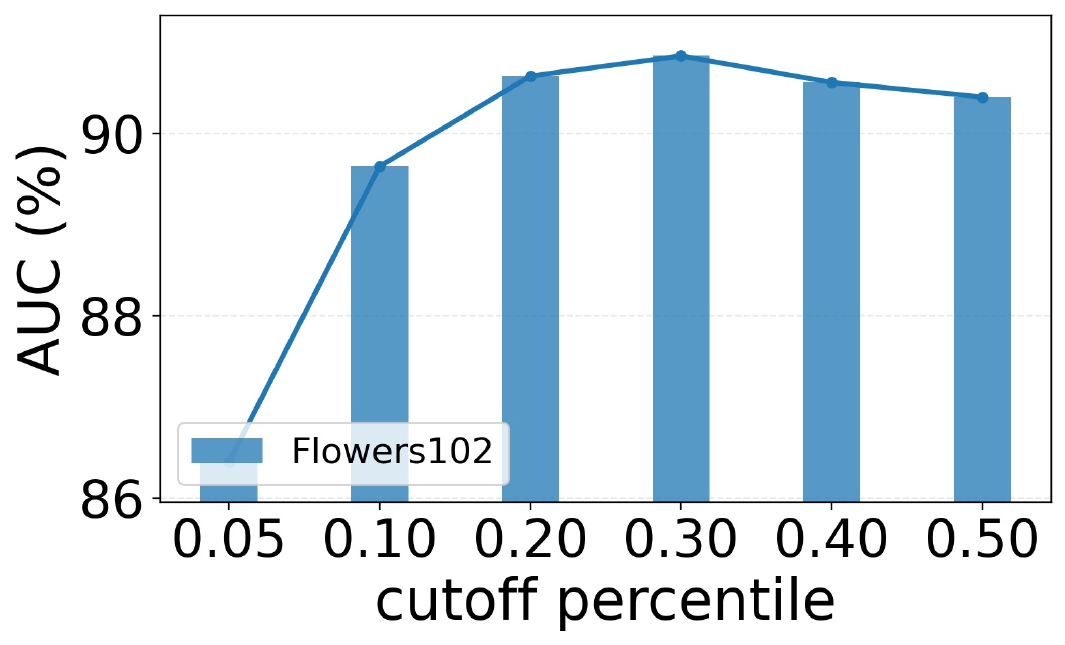}
        \caption{Flowers102}
    \end{subfigure}
    \hfill
    \begin{subfigure}[b]{0.23\textwidth}
        \includegraphics[width=\textwidth]{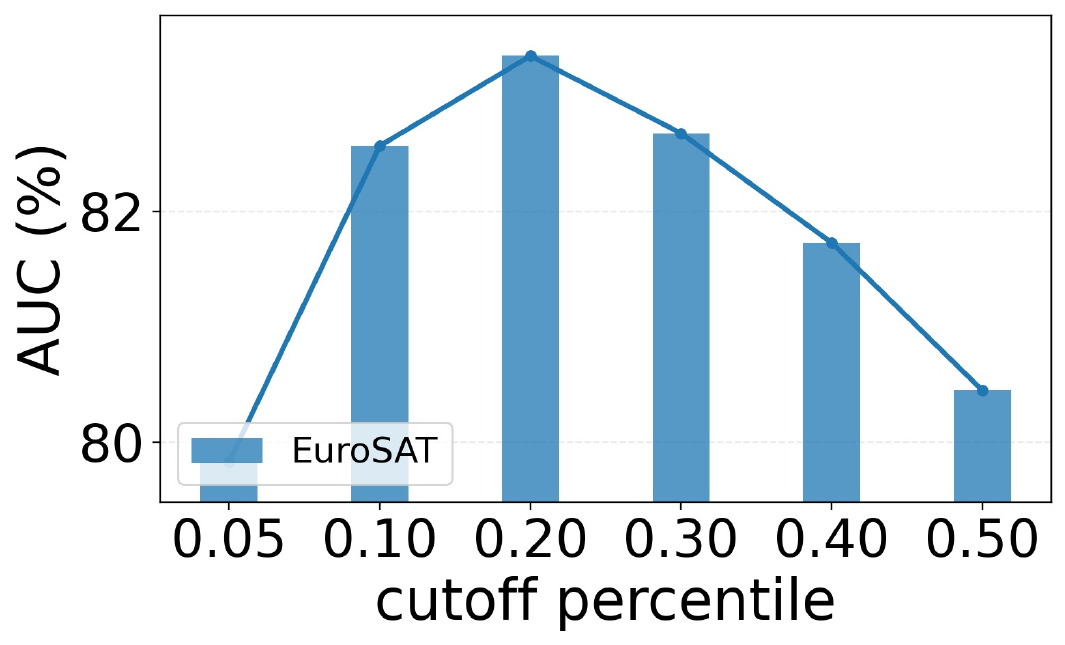}
        \caption{EuroSAT}
    \end{subfigure}

    \vspace{0.3em}
    
    \vspace{1em}

    \begin{subfigure}[b]{0.23\textwidth}
        \includegraphics[width=\textwidth]{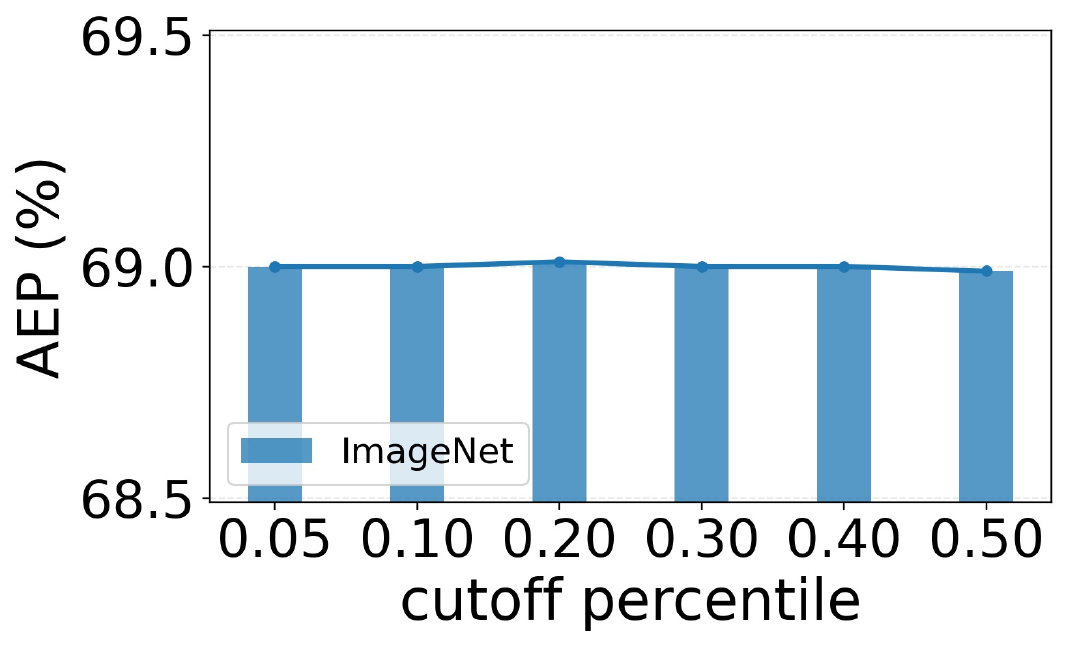}
        \caption{ImageNet}
    \end{subfigure}
    \hfill
    \begin{subfigure}[b]{0.23\textwidth}
        \includegraphics[width=\textwidth]{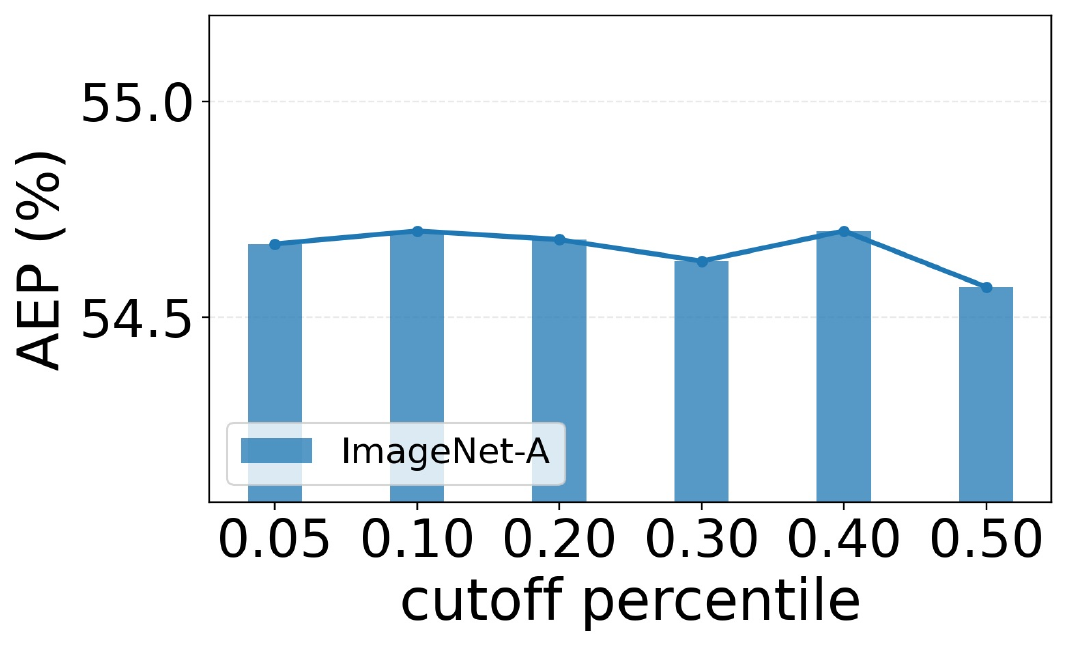}
        \caption{ImageNet-A}
    \end{subfigure}
    \hfill
    \begin{subfigure}[b]{0.23\textwidth}
        \includegraphics[width=\textwidth]{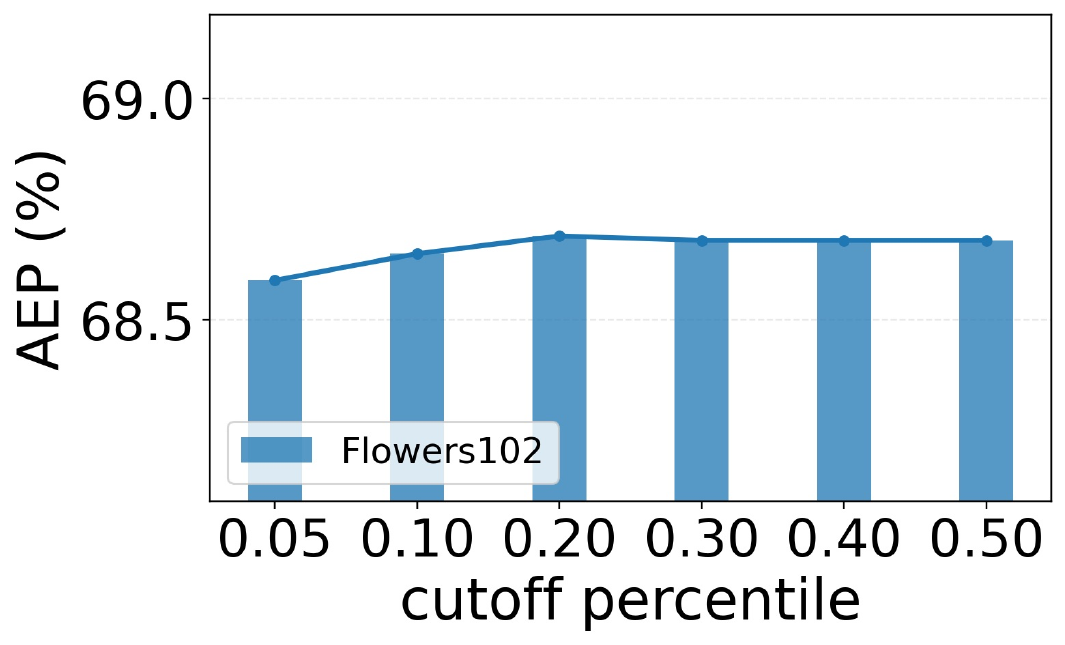}
        \caption{Flowers102}
    \end{subfigure}
    \hfill
    \begin{subfigure}[b]{0.23\textwidth}
        \includegraphics[width=\textwidth]{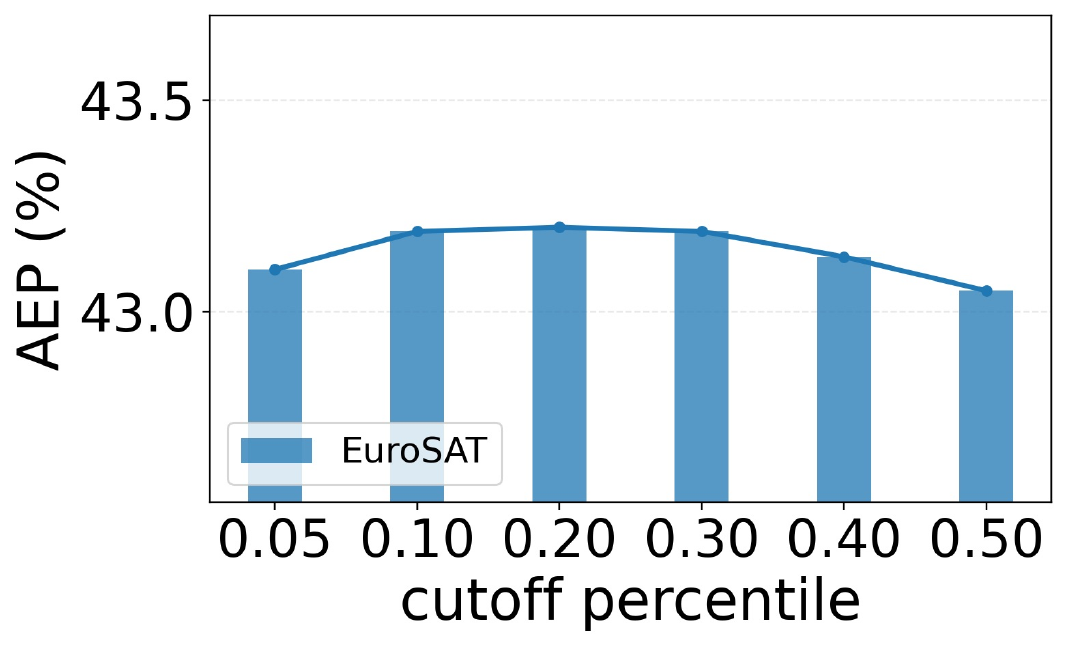}
        \caption{EuroSAT}
    \end{subfigure}

    \caption{Sensitivity analysis of CAS under different cutoff percentile $\rho$ ranging from 0.05 to 0.50 under TPT with ViT-B/16. (a)-(d) demonstrate the AUC performance on ImageNet, ImageNet-A, Flowers102, and EuroSAT, respectively, while (e)-(h) show the AEP results for the same datasets.}
    \label{fig:auc_aep_compare}
\end{figure}

\textbf{Ablation study.} 
CAS consists of augmentation prediction consistency and similarity reweighting, which correspond to the consistency and similarity components, respectively.
To analyze their individual contributions, we evaluate each component independently as a scoring function. 
The results are reported in Table~\ref{tab:ablation}.
Both consistency and similarity serve as effective skipping strategies compared to the classical MCM criterion~\cite{ming2022delving}, consistently yielding an improvement of about 30\% in AUC on ImageNet and its variants.
However, the integrated CAS baseline achieves the most stable and competitive performance overall, attaining the best or near-best AUC and AEP across all methods.
While consistency yields slightly higher gains than CAS in a few isolated cases, CAS demonstrates more stable improvements, particularly on fine-grained datasets.
Notably, CAS improves the AUC from 90.82\% to 91.25\% under ZERO~\cite{farina2024frustratingly}, further validating the effectiveness of combining both components.

\textbf{Computation cost.} 
To validate the efficiency of our baseline, we measure the total inference time of different TTA methods~\cite{shu2022testtime,farina2024frustratingly,dafnis2025testtime,sheng2025r} with and without CAS, as shown in Table~\ref{tb:speed}. 
Specifically, we report the total inference time on ImageNet using ViT-B/16.
By integrating CAS, the overall adaptation time of TPT~\cite{shu2022testtime} decreases from 7.76 to 2.55 hours. 
This delivers a $3.04\times$ speedup over full adaptation while marginally improving accuracy from $68.88\%$ to $69.03\%$.
For training-free TTA methods such as ZERO~\cite{farina2024frustratingly} and computationally efficient methods like STS~\cite{dafnis2025testtime}, CAS further reduces computational overhead by approximately 20\% and 10\%, respectively, while simultaneously improving accuracy.
These results demonstrate that CAS effectively accelerates the overall adaptation process without sacrificing TTA performance.

\textbf{Different AEP measuring functions.}
We choose a decreasing linear function because lower skip ratios may be more preferred in real-world deployments as they sacrifice less performance.
And $f(s)=2(1-s)$ was explicitly chosen because its integral evaluates exactly to 1.
To verify that our method is not dependent on a specific AEP metric function, we further evaluate CAS with two alternative AEP functions, including $f(s)=1$ and $f(s)=2s$.
As shown in Table~\ref{tab:more_f}, CAS consistently achieves the best performance under both alternative settings.
Specifically, CAS obtains the highest average AEP of 62.19\% with $f(s)=1$ and 61.94\% with $f(s)=2s$, outperforming other strategies.
These results indicate that CAS remains robust across different AEP measuring functions.

\textbf{Different data augmentations.}
To examine whether CAS depends on a specific augmentation strategy, we replace the AugMix~\cite{hendrycks2019augmix} augmentation with the random resized crops / horizontal flips used in ZERO~\cite{farina2024frustratingly}.
As shown in Table~\ref{tab:aug_i}, CAS still achieves the best performance on ImageNet and its variants.
Specifically, CAS obtains the highest average AUC of 92.65\% and average AEP of 63.76\%, outperforming other strategies by a clear margin, indicating that CAS remains effective under different data augmentation settings.

\textbf{Sensitivity to cutoff percentile $\rho$.}
To evaluate the sensitivity of CAS to cutoff percentile $\rho$, we vary it from 0.05 ($|S|=3$) to 0.50 ($|S|=32$) across four methods~\cite{shu2022testtime,farina2024frustratingly,dafnis2025testtime,sheng2025r} . 
As illustrated in Figure~\ref{fig:auc_aep_compare}, the AUC remains robust across different $\rho$. 
Initially, increasing the number of selected augmentations improves performance.
For example, on ImageNet, increasing $\rho$ by 20\% improves AUC by about 1\%.
However, when $\rho$ becomes too large, additional views with high entropy are introduced, which may harm performance and lead to slight degradation.
Similar trends are observed on ImageNet-A and fine-grained datasets. 
In contrast, AEP varies only marginally across different $\rho$, indicating that larger ratios bring limited overall benefit. 
Notably, on EuroSAT, AEP even declines as $\rho$ increases, suggesting that excessive augmentations may negatively affect performance.
Based on these results, we set the cutoff percentile $\rho$ to 0.1 as the default setting,
as this configuration maintains high detection quality while minimizing computational overhead.

\subsection{Case Study}

\begin{wrapfigure}[22]{r}{0.48\textwidth}
\centering
\vspace{-14pt}
\includegraphics[height=7cm, width=0.48\textwidth]{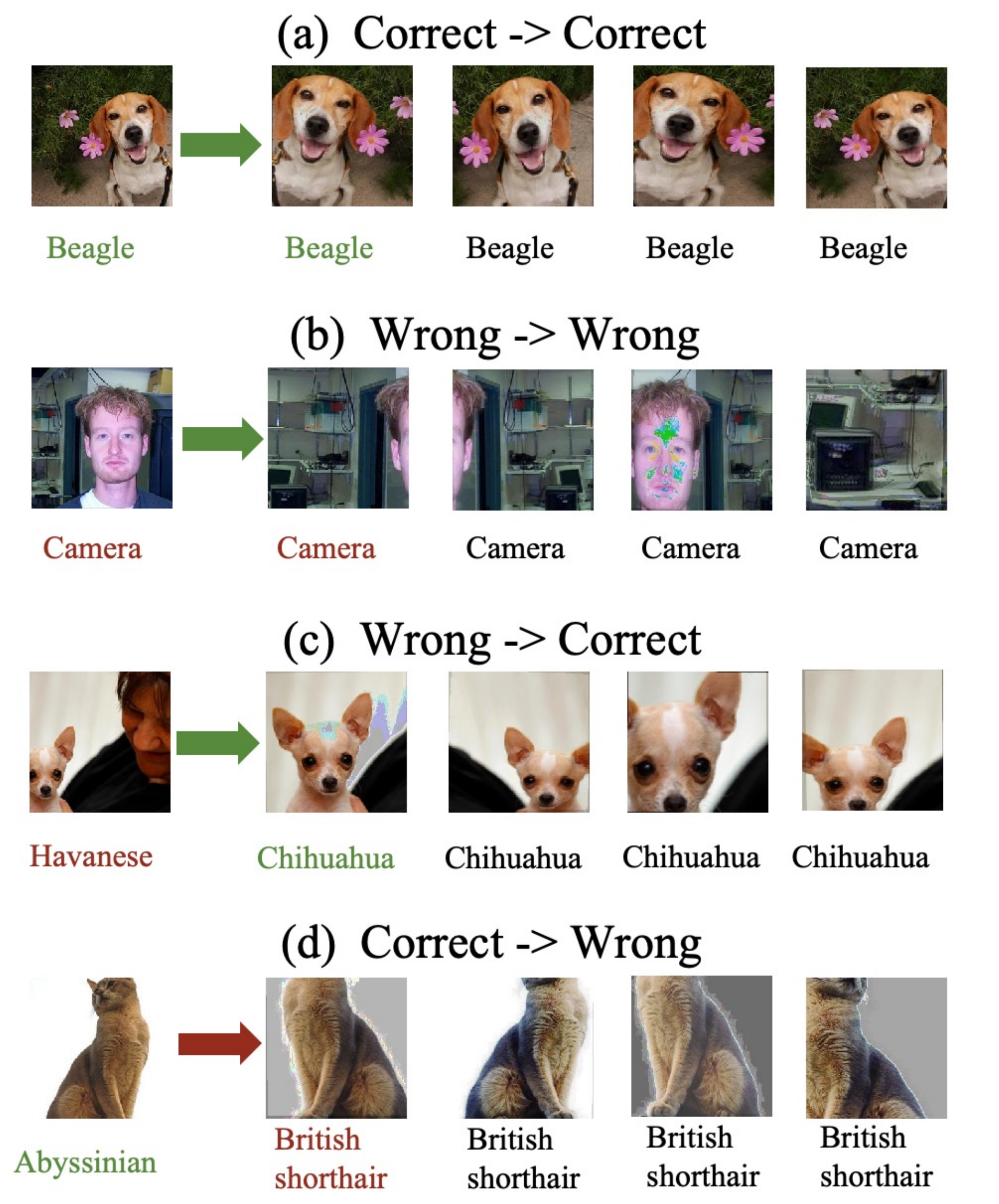}
\vspace{-5pt}
\caption{Visualization of augmented views and their corresponding predictions.}
\label{fig:visual}
\vspace{-10pt}
\end{wrapfigure}

For correct-to-correct samples, predictions are stable across augmentations, indicating that the model has learned robust and invariant representations.
Conversely, wrong-to-wrong samples yield consistently incorrect predictions, suggesting stable but biased representations that augmentation alone cannot rectify.
Meanwhile, wrong-to-correct samples lie near decision boundaries, where augmentations provide consistent corrective signals.
In contrast, correct-to-wrong samples are overly sensitive: perturbations disrupt originally correct cues, leading to performance degradation.
Overall, stable samples offer limited adaptation gains, boundary samples benefit the most from adaptation, and highly sensitive samples risk degradation from improper updates.

\section{Conclusion}
While existing TTA methods generally prioritize overall performance gains, this paper shifts the focus toward adaptation efficiency at the per-sample level.
Our main contribution is the introduction of a new selective adaptation problem, which aims to determine whether a given test sample should undergo adaptation or be skipped.
 We also introduce CAS as a simple baseline that maintains performance with reduced computational overhead, while improving end performance serves as an added benefit.
We hope this work inspires the community to further investigate this problem and build upon our baseline. 
Additionally, we encourage the exploration of new directions, such as extending selective adaptation to tasks beyond image classification.


\section*{Acknowledgements}
We thank Dr. Lijun Sheng for his critical discussions, and the anonymous reviewers for their constructive comments and helpful suggestions that improved this paper.
This work was funded by the National Natural Science Foundation of China under Grants 62276256 and U2441251, Beijing Natural Science Foundation Z260008, and National Key Research and Development Program of China 2026ZD1500301.

%
%
\bibliographystyle{splncs04}
\bibliography{eccv}

\clearpage
\appendix

\section{Algorithm}
We provide the pseudo-code for the proposed Cross-Augmentation Similarity (CAS) in Algorithm~\ref{alg:cas}. 
For a given test image $x$, the CAS score is computed by evaluating the prediction consistency and distribution similarity across its high-quality augmented views. 
Samples with a high CAS score can skip the adaptation process and rely directly on zero-shot predictions.

\begin{algorithm}[H]
\caption{CAS Algorithm}
\label{alg:cas}
\footnotesize
\textbf{Input:} Test image $x$, pretrained VLM $f_\theta$, augmentation function $\mathcal{A}(\cdot)$, augmentation number $(N-1)$, cutoff percentile $\rho$, threshold $\gamma$. \\
\textbf{Output:} ${y}(x)$

\begin{algorithmic}[1]
\State $p_\text{zs}(\mathcal{A}_i(x))\gets \text{softmax}(f_\theta(\mathcal{A}_i(x)))$
\State $H(\mathcal{A}_i(x)) \gets - p_\text{zs}(\mathcal{A}_i(x)) \log p_\text{zs}(\mathcal{A}_i(x))$

\State $k \gets \lfloor N\rho \rfloor$
\State $\mathcal{S} \gets$ indices of $k$ smallest $H(\mathcal{A}_i(x))$


\State $c_i \leftarrow \mathbb{I}\!\left({y}_{zs}(\mathcal{A}_i(x))={y}_{zs}(\mathcal{A}_0(x))\right)$

\State $s_i \leftarrow \cos\!\left(p_{zs}(\mathcal{A}_i(x)),\, p_{zs}(\mathcal{A}_0(x))\right)$

\State $W \leftarrow \sum_{i\in S} s_i$
\State $\alpha_{\text{CAS}}(x) \leftarrow \frac{1}{W} \sum_{i\in S} c_i \cdot s_i$

\If{$\alpha_\text{CAS}(x) < \gamma$} 
    \State Perform adaptation
    \State \Return ${y}_\textbf{adapt}(x)$
\Else
    \State \Return ${y}_\textbf{zs}(x)$
\EndIf
\end{algorithmic}
\end{algorithm}

\section{Results on ResNet-50}
\textbf{Performance on ImageNet and its variants}
We evaluate the performance of CAS on ImageNet and its variants using the ResNet-50~\cite{he2016deep} backbone, as summarized in Table~\ref{tab:rn50_results}.
As demonstrated, CAS consistently outperforms all competing selection strategies across various TTA methods with respect to both AUC and AEP metrics.
Specifically, when integrated into the ZERO~\cite{farina2024frustratingly}, CAS achieves the best overall performance, attaining an average AUC of 87.96\% and an AEP of 47.54\%.
This represents a notable improvement over the second-best baseline, MCM~\cite{ming2022delving}, which achieves an average AUC of 54.88\%, demonstrating stronger detection ability of CAS for selective adaptation.
On the ImageNet dataset, CAS integrated with ZERO reaches an AUC of 91.86\%, surpassing MCM's 61.44\%.
Even across challenging datasets like ImageNet-A and ImageNet-R, CAS maintains robust performance, peaking at an AUC of 83.28\% under TPT and 89.36\% under STS.
These results show that CAS generalizes well to other architectures, effectively separating beneficial from harmful adaptations. 
\begin{table*}[t]
\centering
\caption{Performance comparison of different selection strategies under various TTA methods with RN50. We report AUC and AEP on ImageNet and its variants. The best results under each TTA method are highlighted in \textbf{bold}.}
\label{tab:rn50_results}
\setlength{\tabcolsep}{8pt}
\resizebox{\linewidth}{!}{
\begin{tabular}{@{}llcccccccccccc@{}}
\toprule
\multirow{2}{*}{Method} & \multirow{2}{*}{Strategy} & \multicolumn{2}{c}{ImageNet} & \multicolumn{2}{c}{ImageNet-A} & \multicolumn{2}{c}{ImageNet-V} & \multicolumn{2}{c}{ImageNet-R} & \multicolumn{2}{c}{ImageNet-K} & \multicolumn{2}{c}{Avg.} \\
\cmidrule(lr){3-4} \cmidrule(lr){5-6} \cmidrule(lr){7-8} \cmidrule(lr){9-10} \cmidrule(lr){11-12} \cmidrule(l){13-14}
& & AUC{$\uparrow$} & AEP{$\uparrow$} & AUC{$\uparrow$} & AEP{$\uparrow$} & AUC{$\uparrow$} & AEP{$\uparrow$} & AUC{$\uparrow$} & AEP{$\uparrow$} & AUC{$\uparrow$} & AEP{$\uparrow$} & AUC{$\uparrow$} & AEP{$\uparrow$} \\
\midrule
\multirow{4}{*}{TPT~\cite{shu2022testtime}} 
& Random & 50.04 & 59.90 & 50.26 & 24.90 & 49.87 & 53.58 & 50.96 & 58.17 & 50.04 & 34.54 & 50.23 & 46.22 \\
& Energy~\cite{liu2020energy} & 56.37 & 60.13 & 50.55 & 24.92 & 54.68 & 53.99 & 59.38 & 58.51 & 51.94 & 34.63 & 54.58 & 46.44 \\
& MCM~\cite{ming2022delving} & 61.19& 60.22& 	47.49& 	24.76& 	58.05& 	53.98& 	62.30& 	58.54& 	49.35& 	34.56& 	55.68& 46.41 \\
& \cellcolor{casbg}CAS & \cellcolor{casbg}\textbf{89.45} & \cellcolor{casbg}\textbf{60.72} & \cellcolor{casbg}\textbf{83.28} & \cellcolor{casbg}\textbf{26.28} & \cellcolor{casbg}\textbf{88.05} & \cellcolor{casbg}\textbf{54.59} & \cellcolor{casbg}\textbf{88.81} & \cellcolor{casbg}\textbf{59.04} & \cellcolor{casbg}\textbf{80.85} & \cellcolor{casbg}\textbf{35.12} & \cellcolor{casbg}\textbf{86.09} & \cellcolor{casbg}\textbf{47.15} \\
\midrule
\multirow{4}{*}{R-TPT~\cite{sheng2025illusion}} 
& Random & 50.64 & 59.97 & 51.24 & 26.22 & 50.11 & 53.62 & 50.80 & 57.31 & 49.94 & 33.85 & 50.55 & 46.19 \\
& Energy~\cite{liu2020energy} & 56.59 & 60.16 & 48.67 & 25.87 & 55.71 & 54.12 & 59.42 & 57.58 & 52.91 & 34.02 & 54.66 & 46.35 \\
& MCM~\cite{ming2022delving} & 62.02 &60.26 &	46.95 &	25.78 &	59.74 &	54.09 &	61.61 &	57.52 &	49.48 &	33.85 &	55.96 &	46.30 \\
& \cellcolor{casbg}CAS & \cellcolor{casbg}\textbf{89.50} & \cellcolor{casbg}\textbf{60.82} & \cellcolor{casbg}\textbf{82.87} & \cellcolor{casbg}\textbf{28.10} & \cellcolor{casbg}\textbf{87.98} & \cellcolor{casbg}\textbf{54.72} & \cellcolor{casbg}\textbf{88.24} & \cellcolor{casbg}\textbf{58.04} & \cellcolor{casbg}\textbf{80.31} & \cellcolor{casbg}\textbf{34.56} & \cellcolor{casbg}\textbf{85.78} & \cellcolor{casbg}\textbf{47.25} \\
\midrule
\multirow{4}{*}{STS~\cite{dafnis2025testtime}} 
& Random & 50.26 & 59.23 & 49.08 & 28.64 & 49.89 & 53.08 & 50.52 & 57.10 & 50.39 & 34.36 & 50.03 & 46.48 \\
& Energy~\cite{liu2020energy} & 57.15 & 59.39 & 49.40 & 28.64 & 55.53 &53.50 & 59.98 & 57.48 & 52.41 & 34.42 & 54.89 & 46.69 \\
& MCM~\cite{ming2022delving} & 61.99&	59.33&47.42&	28.41&	58.74&	53.34&	61.37&	57.18&	47.87&	34.11&	55.48&	46.47 \\
& \cellcolor{casbg}CAS & \cellcolor{casbg}\textbf{91.43} & \cellcolor{casbg}\textbf{59.72} & \cellcolor{casbg}\textbf{83.29} & \cellcolor{casbg}\textbf{31.56} & \cellcolor{casbg}\textbf{89.70} & \cellcolor{casbg}\textbf{53.96} & \cellcolor{casbg}\textbf{89.36} & \cellcolor{casbg}\textbf{57.64} & \cellcolor{casbg}\textbf{83.22} & \cellcolor{casbg}\textbf{34.91} & \cellcolor{casbg}\textbf{87.40} & \cellcolor{casbg}\textbf{47.56} \\
\midrule
\multirow{4}{*}{ZERO~\cite{farina2024frustratingly}} 
& Random & 50.23 & 59.69 & 50.45 & 27.58 & 50.55 & 53.48 & 51.08 & 57.32 & 50.39 & 34.35 & 50.54 & 46.48 \\
& Energy~\cite{liu2020energy} & 56.24 & 59.84 & 48.72 & 27.37 & 54.36 & 53.86 & 59.58 & 57.58 & 52.05 & 34.42 & 54.19 & 46.61 \\
& MCM~\cite{ming2022delving} & 61.44 &	59.86 &46.12 &	27.09 &	57.76 &	53.79 &61.34	 &57.38	 &47.73	 &34.12	 &54.88 &46.45 \\
& \cellcolor{casbg}CAS & \cellcolor{casbg}\textbf{91.86} & \cellcolor{casbg}\textbf{60.38} & \cellcolor{casbg}\textbf{84.33} & \cellcolor{casbg}\textbf{29.98} & \cellcolor{casbg}\textbf{90.36} & \cellcolor{casbg}\textbf{54.52} & \cellcolor{casbg}\textbf{89.80} & \cellcolor{casbg}\textbf{57.89} & \cellcolor{casbg}\textbf{83.44} & \cellcolor{casbg}\textbf{34.93} & \cellcolor{casbg}\textbf{87.96} & \cellcolor{casbg}\textbf{47.54} \\
\bottomrule
\end{tabular}
}

\end{table*}

\begin{table*}[t]
\centering
\caption{Performance comparison of different selection strategies under various TTA methods with RN50. We report AUC and AEP on fine-grained and downstream datasets. The best results under each TTA method are highlighted in \textbf{bold}.}
\label{tab:rn50_fg}
\renewcommand{\arraystretch}{0.8}
\setlength{\tabcolsep}{6pt}
\resizebox{1\linewidth}{!}{
\begin{tabular}{@{}llccccccccccccc@{}}
\toprule
Method & Strategy & Metric & Flow. & DTD & Pets & UCF & Cal. & Air. & Euro. & Cars & Food & SUN & Avg. \\
\midrule
\multirow{8}{*}{TPT~\cite{shu2022testtime}}
& \multirow{2}{*}{Random} & AUC & 49.05 & 47.29 & 50.85 & 47.07 & 49.54 & 47.08 & 49.70 & 48.99 & 50.23 & 51.13 & 49.09 \\
&  & AEP & 62.15 & 40.92 & 84.09 & 59.99 & 87.18 & 16.77 & 26.66 & 57.43 & 74.67 & 60.59 & 57.05 \\
\cmidrule(lr){2-14}
& \multirow{2}{*}{Energy~\cite{liu2020energy}} & AUC & 56.70 & 58.58 & 59.40 & 51.22 & 65.13 & 46.87 & 18.87 & 51.02 & 62.63 & 56.24 & 52.67 \\
&  & AEP & 
62.00 & 41.42 & 83.99 & 60.16 & 88.02 & 16.81 & 21.93 & 57.53 & 74.85 & 60.71 & 56.74\\
\cmidrule(lr){2-14}
& \multirow{2}{*}{MCM~\cite{ming2022delving}} & AUC & 59.77& 	62.59& 	65.40	& 57.53	& 76.43	& 41.22& 	31.60& 	56.96& 	70.62& 	63.18& 	58.53\\
&  & AEP & 62.10& 	41.37& 	84.07& 	60.11	& \textbf{88.09}& 	16.38& 	24.62& 	57.77	& 74.89& 	60.86& 	57.03 \\
\cmidrule(lr){2-14}
& \cellcolor{casbg} & \cellcolor{casbg}AUC & \cellcolor{casbg}\textbf{84.58} & \cellcolor{casbg}\textbf{82.78} & \cellcolor{casbg}\textbf{96.22} & \cellcolor{casbg}\textbf{89.57} & \cellcolor{casbg}\textbf{96.95} & \cellcolor{casbg}\textbf{66.91} & \cellcolor{casbg}\textbf{72.64} & \cellcolor{casbg}\textbf{83.55} & \cellcolor{casbg}\textbf{91.87} & \cellcolor{casbg}\textbf{89.05} & \cellcolor{casbg}\textbf{85.41} \\
& \cellcolor{casbg}\multirow{-2}{*}{CAS} & \cellcolor{casbg}AEP & \cellcolor{casbg}\textbf{62.41} & \cellcolor{casbg}\textbf{41.48} & \cellcolor{casbg}\textbf{84.52} & \cellcolor{casbg}\textbf{60.67} & \cellcolor{casbg}88.07 & \cellcolor{casbg}\textbf{17.41} & \cellcolor{casbg}\textbf{27.84} & \cellcolor{casbg}\textbf{58.37} & \cellcolor{casbg}\textbf{75.06} & \cellcolor{casbg}\textbf{61.37} & \cellcolor{casbg}\textbf{57.72} \\
\midrule

\multirow{8}{*}{R-TPT~\cite{sheng2025illusion}}
& \multirow{2}{*}{Random} & AUC & 47.50 & 46.52 & 49.78 & 46.79 & 52.54 & 48.74 & 49.63 & 49.29 & 50.38 & 51.10 & 49.23 \\
&  & AEP &\textbf{ 61.44} & 40.53 & 83.88 & 59.08 & 86.08 & 17.06 & 21.86 & 57.22 & \textbf{73.64} & 60.19 & 56.10 \\
\cmidrule(lr){2-14}
& \multirow{2}{*}{Energy~\cite{liu2020energy}} & AUC & 56.27 & 56.38 & 60.95 & 52.54 & 64.51 & 47.27 & 18.30 & 51.37 & 62.53 & 55.95 & 52.61 \\
&  & AEP & 61.03 & 40.86 & 83.74 & 59.21 & \textbf{86.55} & 17.13 & 16.05 & 57.18 & 73.56 & 60.23 & 55.55\\
\cmidrule(lr){2-14}
& \multirow{2}{*}{MCM~\cite{ming2022delving}} & AUC &60.56	& 63.16& 	66.79& 	60.06& 	77.07	& 40.93& 	31.47& 	57.62	& 71.52& 	63.39& 	59.26\\
&  & AEP & 61.05& 	41.04& 	83.79& 	59.09& 	86.50& 	16.47	& 19.65& 	57.39& 	73.52& 	60.33& 	55.88 \\
\cmidrule(lr){2-14}
& \cellcolor{casbg} & \cellcolor{casbg}AUC & \cellcolor{casbg}\textbf{83.83} & \cellcolor{casbg}\textbf{82.42} & \cellcolor{casbg}\textbf{96.36} & \cellcolor{casbg}\textbf{88.57} & \cellcolor{casbg}\textbf{96.73} & \cellcolor{casbg}\textbf{68.04} & \cellcolor{casbg}\textbf{70.00} & \cellcolor{casbg}\textbf{83.85} & \cellcolor{casbg}\textbf{92.35} & \cellcolor{casbg}\textbf{88.96} & \cellcolor{casbg}\textbf{85.11} \\
& \cellcolor{casbg}\multirow{-2}{*}{CAS} & \cellcolor{casbg}AEP & \cellcolor{casbg}61.33 & \cellcolor{casbg}\textbf{41.09} & \cellcolor{casbg}\textbf{84.20} & \cellcolor{casbg}\textbf{59.63} & \cellcolor{casbg}86.49 & \cellcolor{casbg}\textbf{17.71} & \cellcolor{casbg}\textbf{22.99} & \cellcolor{casbg}\textbf{58.15} & \cellcolor{casbg}73.58 & \cellcolor{casbg}\textbf{60.81} & \cellcolor{casbg}\textbf{56.60} \\
\midrule

\multirow{8}{*}{STS~\cite{dafnis2025testtime}}
& \multirow{2}{*}{Random} & AUC & 47.33 & 48.75 & 49.07 & 46.56 & 51.40 & 49.85 & 49.10 & 48.32 & 49.65 & 50.76 & 49.08 \\
&  & AEP & \textbf{59.15} & 39.43 & 83.29 & 59.01 & 86.41 & 16.93 & 22.60 & 56.82 & \textbf{72.22 }& 59.51 & 55.54 \\
\cmidrule(lr){2-14}
& \multirow{2}{*}{Energy~\cite{liu2020energy}} & AUC & 55.87 & 56.95 & 60.46 & 52.02 & 66.18 & 44.29 & 17.34 & 52.14 & 62.08 & 55.69 & 52.30 \\
&  & AEP & 58.29 & 39.58 & 83.06 & 59.20 &\textbf{ 87.02}& 16.66 & 16.70 & 56.94 & 71.91 & 59.51 & 54.89  \\
\cmidrule(lr){2-14}
& \multirow{2}{*}{MCM~\cite{ming2022delving}} & AUC & 58.82	& 62.36	& 65.82	& 60.21	& 78.22	& 38.00& 	31.86& 	58.45	& 69.66	& 62.20& 	58.56\\
&  & AEP & 58.19& \textbf{39.61}& 	83.05& 	59.00& 	86.98	& 16.18& 	20.28	& 56.98& 	71.64& 	59.54	& 55.15 \\
\cmidrule(lr){2-14}
& \cellcolor{casbg} & \cellcolor{casbg}AUC & \cellcolor{casbg}\textbf{89.97} & \cellcolor{casbg}\textbf{87.75} & \cellcolor{casbg}\textbf{96.76} & \cellcolor{casbg}\textbf{90.22} & \cellcolor{casbg}\textbf{97.08} & \cellcolor{casbg}\textbf{72.34} & \cellcolor{casbg}\textbf{69.96} & \cellcolor{casbg}\textbf{88.03} & \cellcolor{casbg}\textbf{93.30} & \cellcolor{casbg}\textbf{91.00} & \cellcolor{casbg}\textbf{87.64} \\
& \cellcolor{casbg}\multirow{-2}{*}{CAS} & \cellcolor{casbg}AEP & \cellcolor{casbg}58.29 & \cellcolor{casbg}39.44 & \cellcolor{casbg}\textbf{83.38} & \cellcolor{casbg}\textbf{59.50} & \cellcolor{casbg}86.80 & \cellcolor{casbg}\textbf{17.47} & \cellcolor{casbg}\textbf{23.25} & \cellcolor{casbg}\textbf{57.53} & \cellcolor{casbg}\textbf{71.53} & \cellcolor{casbg}\textbf{59.82} & \cellcolor{casbg}\textbf{55.70} \\
\midrule

\multirow{8}{*}{ZERO~\cite{farina2024frustratingly}}
& \multirow{2}{*}{Random} & AUC & 47.78 & 50.69 & 48.81 & 45.69 & 50.49 & 49.58 & 49.56 & 48.55 & 50.31 & 50.76 & 49.22 \\
&  & AEP & \textbf{59.78 }& 39.41 & 83.78 & 58.77 & 86.21 & 16.98 & 22.44 & 57.21 &\textbf{ 72.83} & 60.04 & 55.75 \\
\cmidrule(lr){2-14}
& \multirow{2}{*}{Energy~\cite{liu2020energy}} & AUC & 57.17 & 56.79 & 61.01 & 50.89 & 66.40 & 46.70 & 18.70 & 52.48 & 62.30 & 55.20 & 52.76 \\
&  & AEP & 59.16 & 39.56 & 83.81 & 58.98 &\textbf{ 86.85} & 16.94 & 16.81 & 57.38 & 72.62 & 60.05 & 55.22 \\
\cmidrule(lr){2-14}
& \multirow{2}{*}{MCM~\cite{ming2022delving}} & AUC & 59.51& 	60.18& 	66.29& 	59.24& 	77.86	& 38.34& 	32.36& 	57.80	& 70.24& 	61.84& 	58.37 \\
&  & AEP & 58.99& \textbf{39.59}& 	83.80& 	58.83& 	86.79& 	16.40	& 20.17& 	57.44& 	72.43& 	60.12& 	55.46 \\
\cmidrule(lr){2-14}
& \cellcolor{casbg} & \cellcolor{casbg}AUC & \cellcolor{casbg}\textbf{89.82} & \cellcolor{casbg}\textbf{85.65} & \cellcolor{casbg}\textbf{96.34} & \cellcolor{casbg}\textbf{90.46} & \cellcolor{casbg}\textbf{97.05} & \cellcolor{casbg}\textbf{73.41} & \cellcolor{casbg}\textbf{71.15} & \cellcolor{casbg}\textbf{87.16} & \cellcolor{casbg}\textbf{93.19} & \cellcolor{casbg}\textbf{91.12} & \cellcolor{casbg}\textbf{87.54} \\
& \cellcolor{casbg}\multirow{-2}{*}{CAS} & \cellcolor{casbg}AEP & \cellcolor{casbg}59.15 & \cellcolor{casbg}39.37 & \cellcolor{casbg}\textbf{84.14} & \cellcolor{casbg}\textbf{59.25} & \cellcolor{casbg}86.57 & \cellcolor{casbg}\textbf{17.55} & \cellcolor{casbg}\textbf{23.49} & \cellcolor{casbg}\textbf{58.10} & \cellcolor{casbg}72.38 & \cellcolor{casbg}\textbf{60.52} & \cellcolor{casbg}\textbf{56.05 }\\
\bottomrule
\end{tabular}
}
\end{table*}

\textbf{Performance on fine-grained datasets.}
We further evaluate CAS on fine-grained datasets, with the results summarized in Table~\ref{tab:rn50_fg}.
CAS consistently achieves the highest average performance across all evaluated TTA methods, outperforming Random Energy~\cite{liu2020energy} and MCM~\cite{ming2022delving} strategies. 
For instance, CAS attains an average AUC of 87.64\% under STS~\cite{dafnis2025testtime}, which represents a notable margin over MCM~\cite{ming2022delving} (58.56\%) and Energy~\cite{liu2020energy} (52.30\%), illustrating the superior capability of CAS in filtering out ineffective adaptations. 
Furthermore, CAS demonstrates remarkable robustness across diverse benchmarks, achieving peak AUCs of 97.08\% on Caltech101 and 93.30\% on Food101 when integrated with STS.
Overall, CAS maintains consistently high average performance across all baselines, highlighting its strong adaptability and superior balance between efficiency and accuracy on fine-grained datasets.

\section{Results on Other TTA Methods}
The main paper focuses on training-based TTA methods for CLIP~\cite{radford2021learning}.
To further examine the generality of CAS, we apply it to the standard model-based TTA method MEMO~\cite{zhang2022memo}, as well as the training-free method MTA~\cite{zanella2024test}.
As shown in the following part, CAS remains effective under both methods, further demonstrating its broad applicability as a selective adaptation strategy.

\begin{table}[t]
    \centering
    \caption{Comparison of AUC and AEP under MEMO~\cite{zhang2022memo} with different skipping strategies on CIFAR-10 and ImageNet-R. Bold numbers indicate the best performance.}
    \label{tab:memo}
    
    \renewcommand{\arraystretch}{1} 
    \setlength{\tabcolsep}{8pt} 
    
    \begin{tabular}{ll cccc}
        \toprule
        \textbf{Dataset} & \textbf{Metric} & \textbf{Random} & \textbf{MCM} & \textbf{Energy} & \textbf{CAS} \\ 
        \midrule

        \multicolumn{6}{l}{\textit{Backbone: ResNet-26 \cite{he2016deep}}} \\
        CIFAR-10 & AUC & 52.12 & 87.22& 84.38 & \textbf{97.98} \\
                 & AEP &92.09 & 92.55& 92.51 & \textbf{92.66} \\
        CIFAR-10-C & AUC & 50.24& 76.52& 73.69 & \textbf{95.10} \\
                 & AEP &79.43& 79.94& 79.84 & \textbf{80.40} \\
        
        \midrule
        \multicolumn{6}{l}{\textit{Backbone: ResNet-50 \cite{he2016deep}}} \\
        ImageNet-R & AUC & 49.93 & 51.77 & 50.52 & \textbf{81.86} \\
                   & AEP & 39.59 &39.73 & 39.74 & \textbf{41.08} \\
        \bottomrule
    \end{tabular}
\end{table}
\begin{figure}[ht]
    \centering
    
    \begin{subfigure}[b]{0.32\textwidth}
        \centering
        \includegraphics[width=\textwidth]{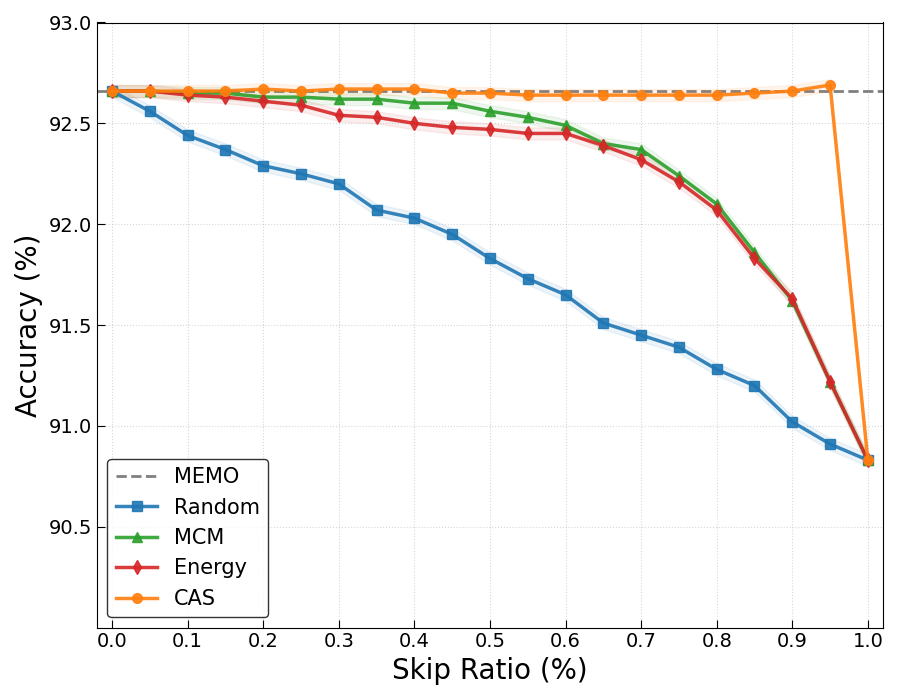}
        \caption{CIFAR-10}
        \label{fig:sub1}
    \end{subfigure}
    \hfill 
    \begin{subfigure}[b]{0.32\textwidth}
        \centering
        \includegraphics[width=\textwidth]{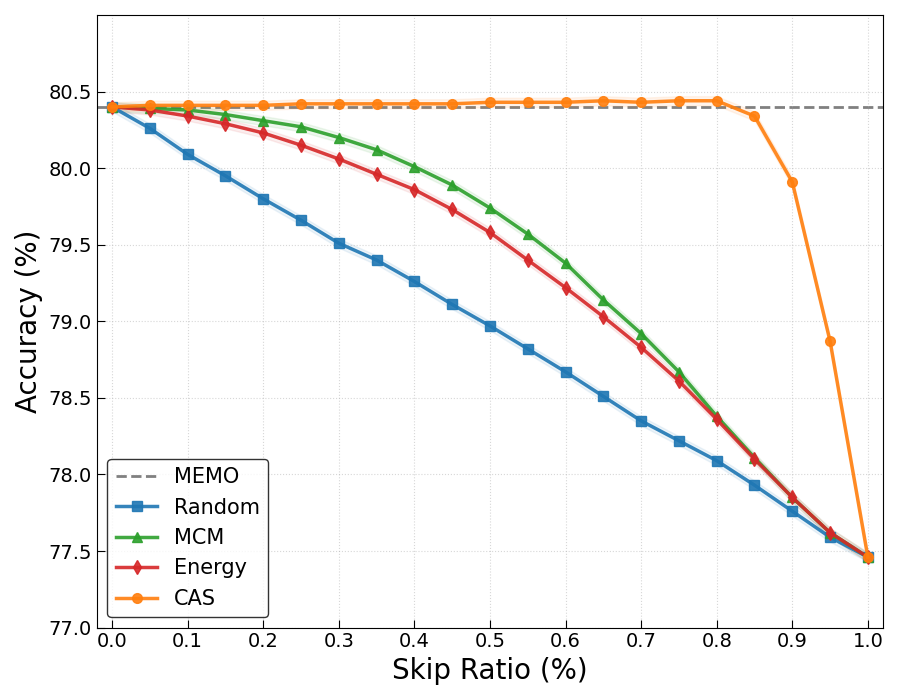}
        \caption{CIFAR-10-C}
        \label{fig:sub2}
    \end{subfigure}
    \hfill
    \begin{subfigure}[b]{0.32\textwidth}
        \centering
        \includegraphics[width=\textwidth]{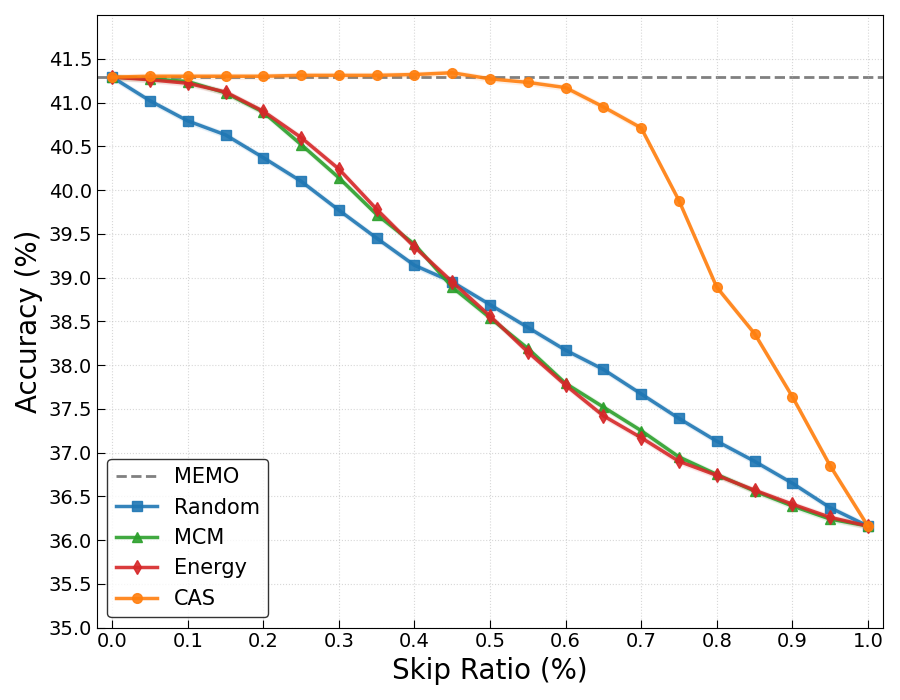}
        \caption{ImageNet-R}
        \label{fig:sub3}
    \end{subfigure}
    
    \caption{Accuracy (\%) versus skip ratio (\%) under different skipping strategies for MEMO.
CAS consistently maintains higher accuracy even when skipping a large proportion of adaptation processes.
Compared with Random, MCM, and Energy, CAS demonstrates a superior efficiency-accuracy trade-off.}
    \label{fig:memofour}
\end{figure}
\subsection{CAS for MEMO~\cite{zhang2022memo}}
To further demonstrate the generality of our baseline, we incorporate CAS as a selective adaptation strategy into MEMO~\cite{zhang2022memo}.
We evaluate it on CIFAR-10~\cite{Krizhevsky09} and CIFAR-10-C~\cite{hendrycks2019benchmarking} with ResNet-26~\cite{he2016deep}, and on ImageNet-R~\cite{hendrycks2021many} with ResNet-50~\cite{he2016deep}.
As summarized in Table~\ref{tab:memo}, CAS consistently outperforms baseline strategies, including Random, Energy~\cite{liu2020energy}, and MCM~\cite{ming2022delving}, across both AUC and AEP metrics.
On CIFAR-10, CAS achieves an AUC of 97.98\%, significantly surpassing the second strongest baseline, MCM~\cite{ming2022delving} (87.22\%).
A similar pattern is observed on CIFAR-10-C, where CAS again obtains the best results, achieving the highest AUC of 95.10\% and AEP of 80.40\%.
On the more challenging ImageNet-R dataset, CAS maintains its advantage, achieving an AUC of 81.86\%, outperforming the second-best method by 30.09\%.
These results further show that CAS generalizes well across different TTA frameworks, highlighting its broad applicability to selective adaptation.

To evaluate the trade-off between efficiency and accuracy, we plot the accuracy--skip ratio curves of different selection strategies under MEMO~\cite{zhang2022memo}, which is illustrated in Fig.~\ref{fig:memofour}.
CAS consistently outperforms Random, MCM, and Energy across almost the entire skip range, maintaining accuracy close to the full-adaptation baseline even when substantial samples are skipped.
By contrast, the other strategies show much larger performance drops as the skip ratio increases.
These results indicate that CAS identifies samples whose adaptation can be skipped more reliably, leading to better efficiency without sacrificing accuracy.

\subsection{CAS for MTA~\cite{zanella2024test}}
\textbf{Performance on ImageNet and its variants.} 
We evaluate the performance of CAS under the training-free method MTA~\cite{zanella2024test} framework on ImageNet and its variants, as summarized in Table~\ref{tab:mta_imagenet}. 
CAS consistently outperforms all selection strategies across all evaluation metrics. 
Specifically, CAS achieves an average AUC of 93.36\%, providing a substantial improvement over the second strongest baseline, MCM, which attains an AUC of 59.80\%. 
On the challenging ImageNet-A dataset, CAS reaches an AUC of 92.12\%, demonstrating its strong capability in identifying ineffective adaptations even in training-free scenarios. 
Across all evaluated variants, its performance remains highly stable, highlighting its ability to effectively distinguish beneficial updates from harmful ones.

\textbf{Performance on fine-grained datasets.} 
We further evaluate CAS on fine-grained datasets under MTA~\cite{zanella2024test}, with results summarized in Table~\ref{tab:mta_fg}.
CAS consistently achieves the highest average performance across all datasets. 
Under the MTA framework, CAS attains a peak average AUC of 92.30\%, significantly surpassing MCM (66.50\%) and Energy (61.03\%) by a wide margin.
Notably, CAS demonstrates robustness across different domains, achieving a high AUC of 99.13\% on Caltech101 and 97.85\% on Pets.
Overall, CAS delivers consistently robust performance across fine-grained classification tasks, demonstrating its superior trade-off between efficiency and accuracy.
\begin{table*}[ht]
\centering
\caption{Performance comparison of different selection strategies under the training-free MTA framework on ImageNet and its variants. We report AUC and AEP. The best results are highlighted in \textbf{bold}.}
\label{tab:mta_imagenet}
\setlength{\tabcolsep}{8pt}
\resizebox{\linewidth}{!}{
\begin{tabular}{@{}llcccccccccccc@{}}
\toprule
\multirow{2}{*}{Method} & \multirow{2}{*}{Strategy} & \multicolumn{2}{c}{ImageNet} & \multicolumn{2}{c}{ImageNet-A} & \multicolumn{2}{c}{ImageNet-V} & \multicolumn{2}{c}{ImageNet-R} & \multicolumn{2}{c}{ImageNet-K} & \multicolumn{2}{c}{Avg.} \\
\cmidrule(lr){3-4} \cmidrule(lr){5-6} \cmidrule(lr){7-8} \cmidrule(lr){9-10} \cmidrule(lr){11-12} \cmidrule(l){13-14}
& & AUC{$\uparrow$} & AEP{$\uparrow$} & AUC{$\uparrow$} & AEP{$\uparrow$} & AUC{$\uparrow$} & AEP{$\uparrow$} & AUC{$\uparrow$} & AEP{$\uparrow$} & AUC{$\uparrow$} & AEP{$\uparrow$} & AUC{$\uparrow$} & AEP{$\uparrow$} \\
\midrule
\multirow{4}{*}{MTA}
& Random & 50.85 & 68.44 & 49.20 & 53.84 & 51.71 & 62.79 & 49.05 & 75.93 & 50.45 & 47.73 & 50.25 & 61.75 \\
& Energy~\cite{liu2020energy} & 57.43 & 68.64 & 50.51 & 54.18 & 57.52 &62.99& 63.11 & 76.47 & 53.06 & 47.79 & 56.33 & 62.01 \\
& MCM~\cite{ming2022delving} & 63.98&	68.74&	50.51&	54.00&	62.57&63.04&	69.18&	76.55&	52.77&	47.73&	59.80&	62.01 \\
& \cellcolor{casbg}CAS & \cellcolor{casbg}\textbf{95.08} & \cellcolor{casbg}\textbf{69.24} & \cellcolor{casbg}\textbf{92.12} & \cellcolor{casbg}\textbf{56.83} & \cellcolor{casbg}\textbf{94.01} & \cellcolor{casbg}\textbf{63.58} & \cellcolor{casbg}\textbf{95.59} & \cellcolor{casbg}\textbf{76.96} & \cellcolor{casbg}\textbf{89.98} & \cellcolor{casbg}\textbf{48.47} & \cellcolor{casbg}\textbf{93.36} & \cellcolor{casbg}\textbf{63.02} \\
\bottomrule
\end{tabular}
}

\vspace{2em}

\centering
\caption{Performance comparison of different selection strategies under the training-free MTA framework on fine-grained and downstream datasets. We report AUC and AEP. The best results are highlighted in \textbf{bold}.}
\label{tab:mta_fg}
\renewcommand{\arraystretch}{0.8}
\setlength{\tabcolsep}{6pt}
\resizebox{1\linewidth}{!}{
\begin{tabular}{@{}llccccccccccccc@{}}
\toprule
Method & Strategy & Metric & Flow. & DTD & Pets & UCF & Cal. & Air. & Euro. & Cars & Food & SUN & Avg. \\
\midrule
\multirow{8}{*}{MTA}
& \multirow{2}{*}{Random} & AUC & 50.46 & 51.68 & 55.67 & 49.73 & 48.59 & 53.04 & 49.88 & 50.74 & 50.38 & 50.09 & 51.03 \\
&  & AEP & 67.41 & 45.32 & \textbf{88.11} & 66.75 & 94.18 & 24.60 & 42.35 & 67.01 & 84.19 & 64.36 & 64.43 \\
\cmidrule(lr){2-14}
& \multirow{2}{*}{Energy~\cite{liu2020energy}} & AUC & 60.38 & 54.51 & 67.94 & 60.63 & 69.36 & 47.94 & 67.74 & 54.43 & 68.64 & 58.68 & 61.03 \\
&  & AEP & 67.35 & 45.42 & 87.99 & 67.17 &\textbf{ 94.47} & 24.34 & \textbf{42.82} & 67.07 & 84.36 & 64.65 & 64.56 \\
\cmidrule(lr){2-14}
& \multirow{2}{*}{MCM~\cite{ming2022delving}} & AUC & 63.78	&67.73	&73.06&	68.02&	79.01&	47.91&	58.97&	63.21&	78.63&	64.64&	66.50 \\
&  & AEP & 67.40&45.71&87.97&67.19&94.41&	24.21&	42.71&	67.28&	84.38&	64.78&	64.60\\
\cmidrule(lr){2-14}
& \cellcolor{casbg} & \cellcolor{casbg}AUC & \cellcolor{casbg}\textbf{93.62} & \cellcolor{casbg}\textbf{90.97} & \cellcolor{casbg}\textbf{97.85} & \cellcolor{casbg}\textbf{94.15} & \cellcolor{casbg}\textbf{99.13} & \cellcolor{casbg}\textbf{79.32} & \cellcolor{casbg}\textbf{82.80} & \cellcolor{casbg}\textbf{93.14} & \cellcolor{casbg}\textbf{97.36} & \cellcolor{casbg}\textbf{94.67} & \cellcolor{casbg}\textbf{92.30} \\
& \cellcolor{casbg}\multirow{-2}{*}{CAS} & \cellcolor{casbg}AEP & \cellcolor{casbg}\textbf{67.52} & \cellcolor{casbg}\textbf{45.91} & \cellcolor{casbg}87.94 & \cellcolor{casbg}\textbf{67.55} & \cellcolor{casbg}94.32 & \cellcolor{casbg}\textbf{24.66} & \cellcolor{casbg}42.56 & \cellcolor{casbg}\textbf{67.60} & \cellcolor{casbg}\textbf{84.43} & \cellcolor{casbg}\textbf{65.21} & \cellcolor{casbg}\textbf{64.77} \\
\bottomrule
\end{tabular}
}
\end{table*}

\section{More Baselines}
We explored additional baseline strategies, including Max Logits, Max Softmax, and GL-MCM~\cite{miyai2025glmcm}.
However, we only reported Energy~\cite{liu2020energy} and MCM~\cite{ming2022delving} in the main text due to their comparable performance and space constraints. 
Table~\ref{tab:more_baselines} summarizes the detailed results for these previously evaluated baselines. 
Specifically, CAS achieves an average AUC of 89.42\%, outperforming the second-best baseline by a margin of 22.63\%. 
Furthermore, CAS yields an average AEP of 62.44\%, further validating its superior capability in maintaining TTA performance in the selective adaptation problem.
\begin{table*}[h]
\centering
\caption{Performance comparison of different baselines under TPT~\cite{shu2022testtime} with ViT-B/16. 
We report AUC and AEP on ImageNet and its variants.}
\label{tab:more_baselines}
\setlength{\tabcolsep}{1.5pt}
\resizebox{\linewidth}{!}{
\begin{tabular}{@{}llcccccccccccc@{}}
\toprule
\multirow{2}{*}{Method} & \multirow{2}{*}{Strategy} & \multicolumn{2}{c}{ImageNet} & \multicolumn{2}{c}{ImageNet-A} & \multicolumn{2}{c}{ImageNet-V} & \multicolumn{2}{c}{ImageNet-R} & \multicolumn{2}{c}{ImageNet-K} & \multicolumn{2}{c}{Avg.} \\
\cmidrule(lr){3-4} \cmidrule(lr){5-6} \cmidrule(lr){7-8} \cmidrule(lr){9-10} \cmidrule(lr){11-12} \cmidrule(l){13-14}
& & AUC{$\uparrow$} & AEP{$\uparrow$} & AUC{$\uparrow$} & AEP{$\uparrow$} & AUC{$\uparrow$} & AEP{$\uparrow$} & AUC{$\uparrow$} & AEP{$\uparrow$} & AUC{$\uparrow$} & AEP{$\uparrow$} & AUC{$\uparrow$} & AEP{$\uparrow$} \\
\midrule
\multirow{4}{*}{TPT~\cite{shu2022testtime}} 
& Max Logits & 59.24 & 68.44 & 52.71 & 52.69 & 59.20 & 62.96 & 65.85 & 76.67 & 54.94 & 47.42 & 58.39 & 61.64 \\
& Max Softmax & 72.85 & 68.64 & 57.47 & 52.88 & 68.85 & 63.02 & 75.07 & 76.75 & 59.72 & 47.38 & 66.79 & 61.73 \\
& GL-MCM~\cite{miyai2025glmcm} & 62.67 & 68.50 & 56.82 & 53.03 & 60.90 & 62.94 & 68.23 & 76.69 & 54.29 & 47.47 & 60.58 & 61.73 \\
& \cellcolor{casbg}CAS & \cellcolor{casbg}\textbf{91.10} & \cellcolor{casbg}\textbf{ 69.00} & \cellcolor{casbg}\textbf{ 88.62} & \cellcolor{casbg}\textbf{54.70 } & \cellcolor{casbg}\textbf{89.69} & \cellcolor{casbg}\textbf{63.45} & \cellcolor{casbg}\textbf{93.45} & \cellcolor{casbg}\textbf{77.14 } & \cellcolor{casbg}\textbf{84.24} & \cellcolor{casbg}\textbf{47.93} & \cellcolor{casbg}\textbf{89.42} & \cellcolor{casbg}\textbf{62.44} \\

\bottomrule
\end{tabular}
}
\end{table*}

\section{Different VLMs}
To demonstrate the generalizability of our approach across different VLMs, we further report the performance of CAS on the SigLIP~\cite{zhai2023sigmoid} backbone. 
As shown in Table~\ref{tab:siglip}, CAS consistently maintains strong detection capabilities, achieving an average AUC of 93.03\% across ImageNet and its OOD variants.
\begin{table*}[h]
\centering
\caption{Performance comparison of different VLMs under TPT~\cite{shu2022testtime} framework. 
We report AUC and AEP on ImageNet and its variants.}
\label{tab:siglip}
\setlength{\tabcolsep}{8pt}
\resizebox{\linewidth}{!}{
\begin{tabular}{@{}llcccccccccccc@{}}
\toprule
\multirow{2}{*}{Method} & \multirow{2}{*}{Strategy} & \multicolumn{2}{c}{ImageNet} & \multicolumn{2}{c}{ImageNet-A} & \multicolumn{2}{c}{ImageNet-V} & \multicolumn{2}{c}{ImageNet-R} & \multicolumn{2}{c}{ImageNet-K} & \multicolumn{2}{c}{Avg.} \\
\cmidrule(lr){3-4} \cmidrule(lr){5-6} \cmidrule(lr){7-8} \cmidrule(lr){9-10} \cmidrule(lr){11-12} \cmidrule(l){13-14}
& & AUC{$\uparrow$} & AEP{$\uparrow$} & AUC{$\uparrow$} & AEP{$\uparrow$} & AUC{$\uparrow$} & AEP{$\uparrow$} & AUC{$\uparrow$} & AEP{$\uparrow$} & AUC{$\uparrow$} & AEP{$\uparrow$} & AUC{$\uparrow$} & AEP{$\uparrow$} \\
\midrule
\multirow{4}{*}{SigLIP~\cite{zhai2023sigmoid}}
& Random & 50.77 & 76.23 & 50.05 & 46.15 & 51.00 & 68.98 & 50.54 & 89.73 & 48.30 & 66.89 & 50.13 & 69.60\\
& Energy~\cite{liu2020energy} & 60.20 & 76.26 & 51.56 & 46.17 & 57.60 & 69.00 & 61.34 & 89.75 & 53.42 & 66.94 & 56.82 & 69.63 \\
& MCM~\cite{ming2022delving} & 67.16 & 76.32 & 51.77 & 46.20 & 62.48 & 69.07 & 74.68 & 89.82 & 62.48 & 66.99 & 63.71 & 69.68 \\
& \cellcolor{casbg}CAS & \cellcolor{casbg}\textbf{95.45 } & \cellcolor{casbg}\textbf{76.46 } & \cellcolor{casbg}\textbf{85.78} & \cellcolor{casbg}\textbf{46.70} & \cellcolor{casbg}\textbf{93.30} & \cellcolor{casbg}\textbf{69.23} & \cellcolor{casbg}\textbf{ 97.12 } & \cellcolor{casbg}\textbf{89.90} & \cellcolor{casbg}\textbf{93.52} & \cellcolor{casbg}\textbf{67.07} & \cellcolor{casbg}\textbf{93.03} & \cellcolor{casbg}\textbf{69.87} \\
\bottomrule
\end{tabular}
}
\end{table*}

\section{Impact of the Number of Test-Time Augmentations}
\begin{figure*}[ht]
    \centering
    \begin{subfigure}[t]{0.48\textwidth}
        \centering
        \includegraphics[width=\linewidth]{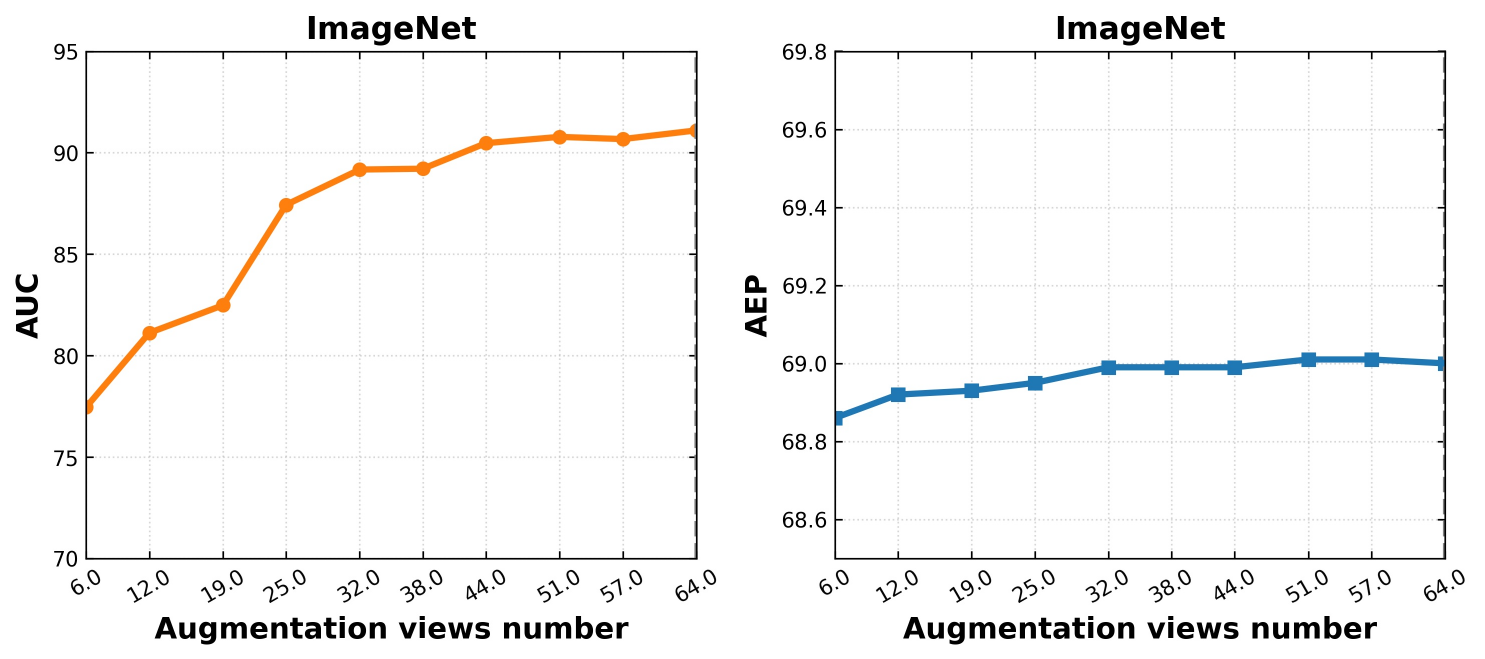}
        \caption{ImageNet}
    \end{subfigure}
    \hfill
    \begin{subfigure}[t]{0.48\textwidth}
        \centering
        \includegraphics[width=\linewidth]{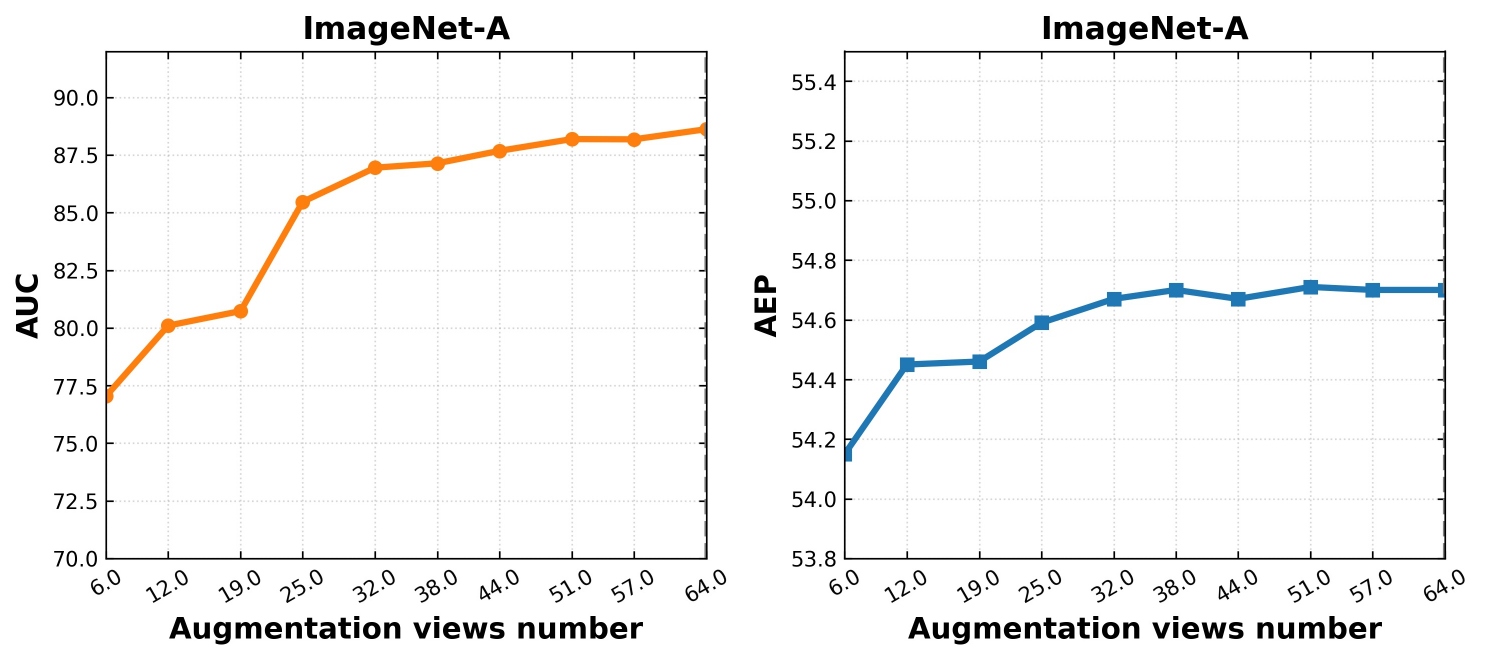}
        \caption{ImageNet-A}
    \end{subfigure}

    \vspace{0.8em}

    \begin{subfigure}[t]{0.48\textwidth}
        \centering
        \includegraphics[width=\linewidth]{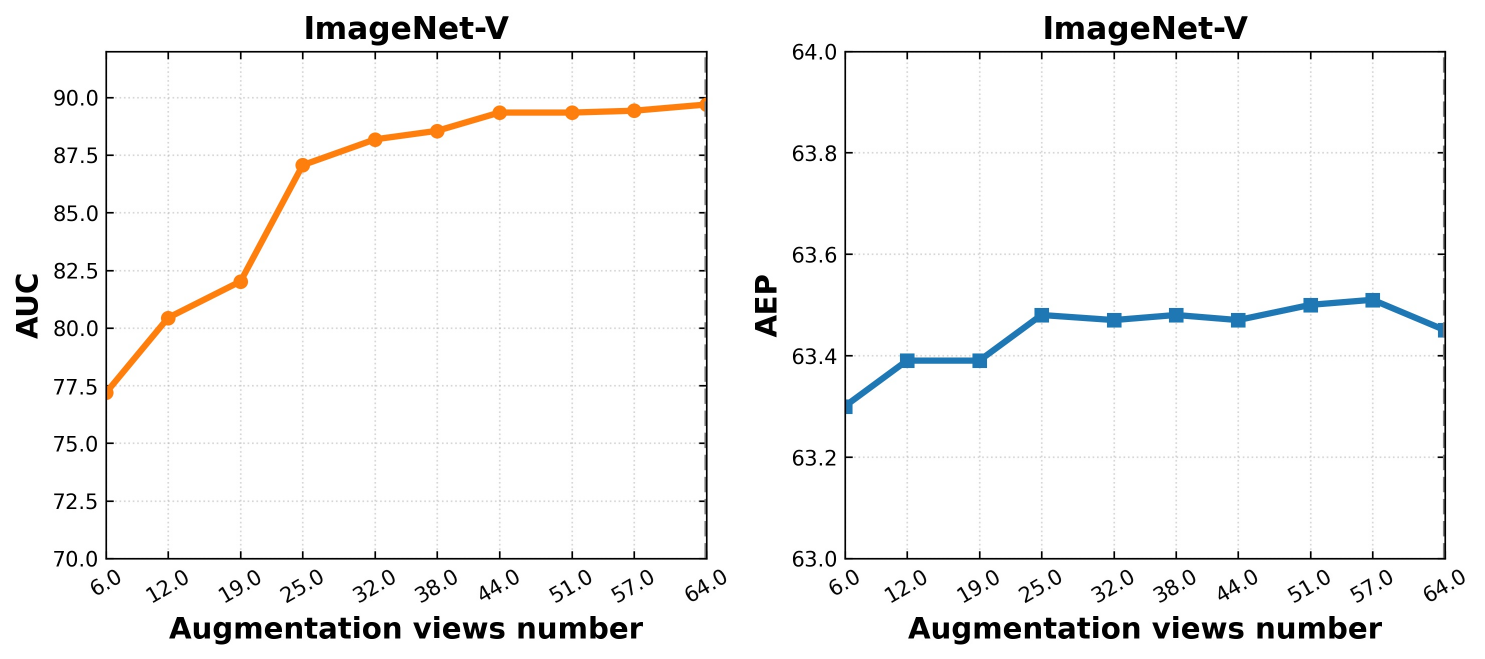}
        \caption{ImageNet-V2}
    \end{subfigure}
    \hfill
    \begin{subfigure}[t]{0.48\textwidth}
        \centering
        \includegraphics[width=\linewidth]{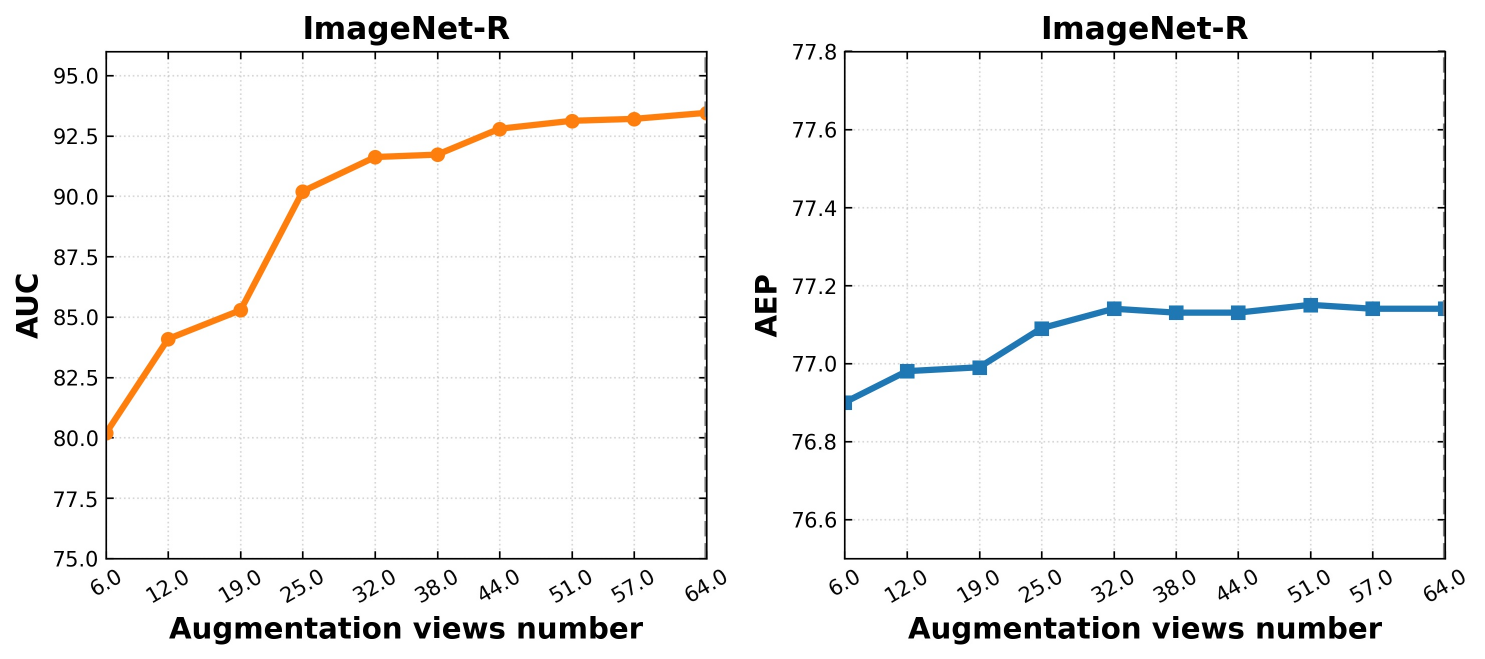}
        \caption{ImageNet-R}
    \end{subfigure}

    \vspace{0.8em}

    \begin{subfigure}[t]{0.48\textwidth}
        \centering
        \includegraphics[width=\linewidth]{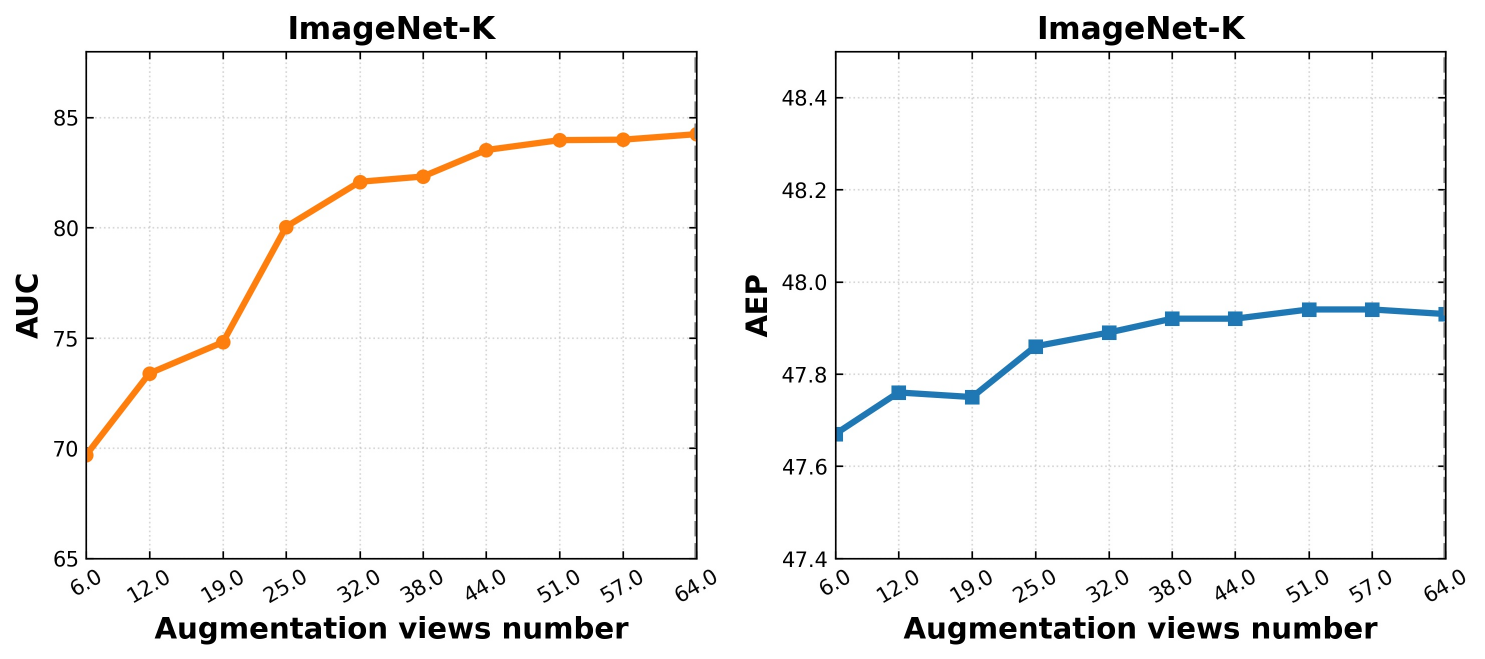}
        \caption{ImageNet-Sketch}
    \end{subfigure}
    \hfill
    \begin{subfigure}[t]{0.48\textwidth}
        \centering
        \includegraphics[width=\linewidth]{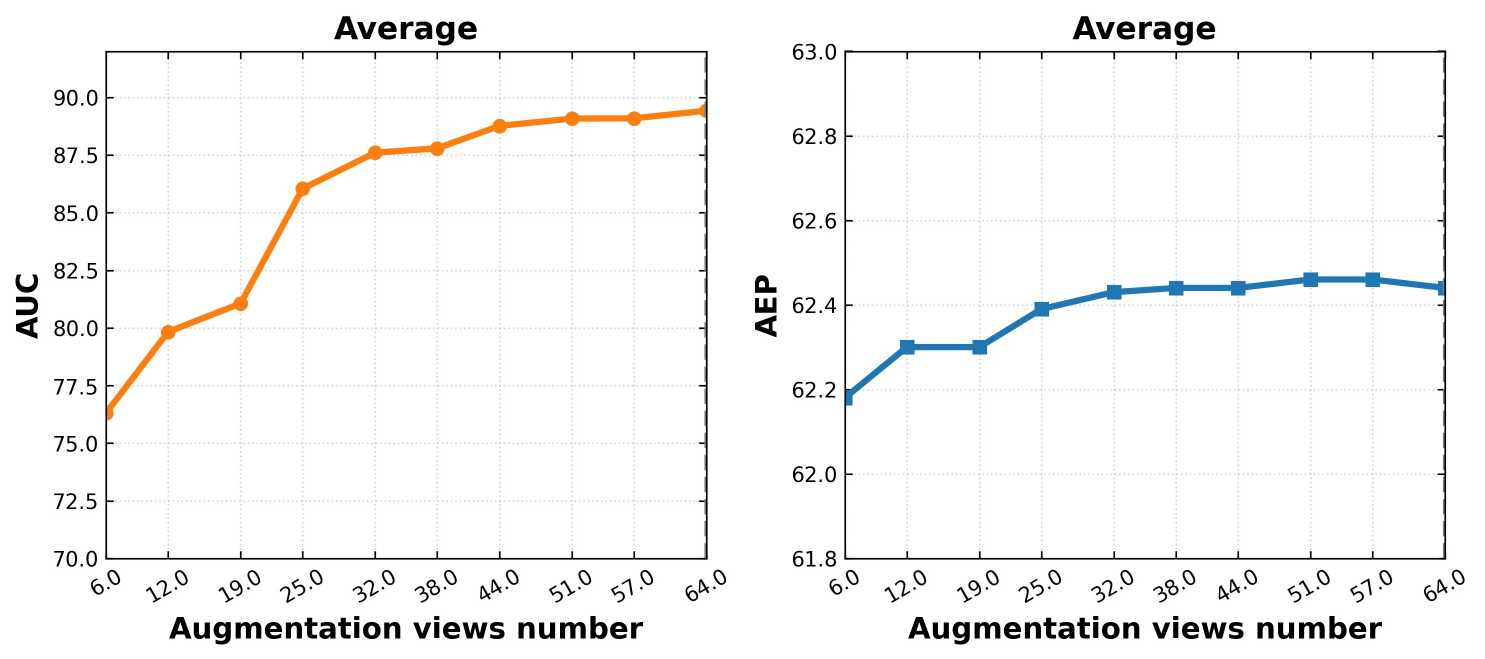}
        \caption{Average}
    \end{subfigure}

    \caption{
    Impact of reducing augmentation views on ImageNet and its variants.
    Even when using substantially fewer augmentations than the default 64 views, both AUC and AEP remain largely stable across datasets, indicating that our method does not rely on excessive augmentations.
    }
    \label{fig:aug_ratio}
\end{figure*}

To assess the sensitivity of CAS to the number of test-time augmentations, we examine how its performance changes under different augmentation budgets.
While existing TTA methods typically rely on a default of 64 views to ensure stable prediction, this high volume incurs significant computational overhead.
In this study, we evaluate the stability of CAS by reducing the number of augmentation views from the original 64 to smaller values of 6.
Figure~\ref{fig:aug_ratio} illustrates the impact of view reduction on AUC and AEP across various datasets. We observe that while performance initially increases with the number of views, it rapidly plateaus at a relatively low view count.
Notably, both AUC and AEP remain largely stable even when using fewer augmentations compared to the default setting. 
This trend indicates that the proposed selective adaptation mechanism is highly robust. It does not rely on an excessive augmentation budget to maintain performance. 

\section{Adaptation Behaviors across Different TTA Methods}
\begin{figure*}[t]
    \centering

    \begin{subfigure}[t]{0.235\textwidth}
        \centering
        \includegraphics[width=\linewidth]{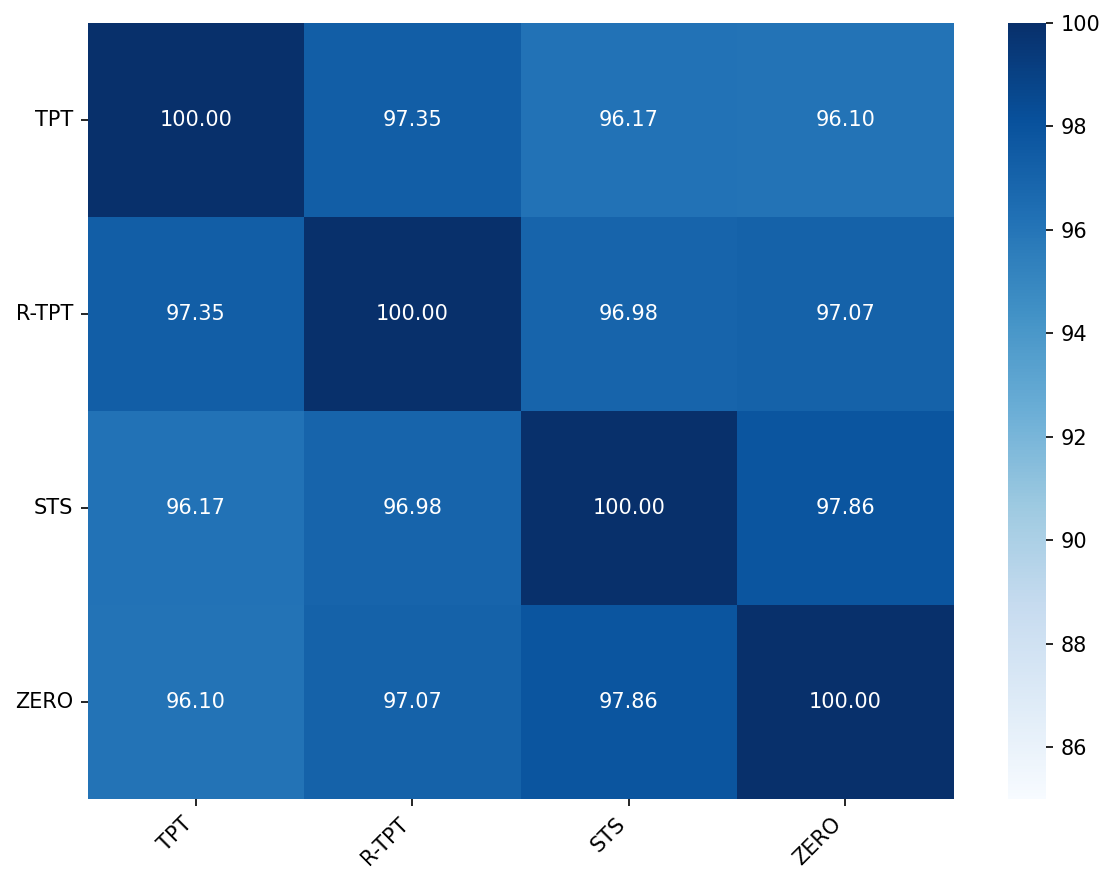}
        \caption{All datasets}
    \end{subfigure}
    \hfill
    \begin{subfigure}[t]{0.235\textwidth}
        \centering
        \includegraphics[width=\linewidth]{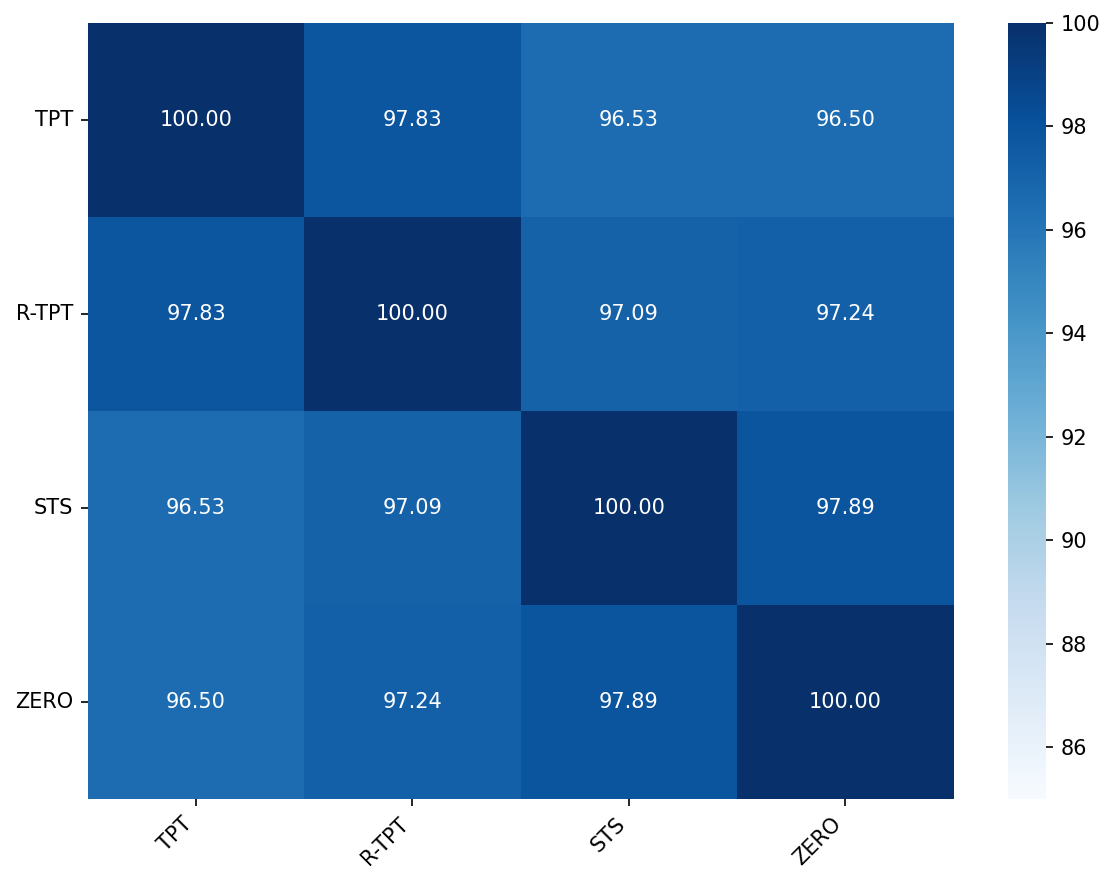}
        \caption{ImageNet}
    \end{subfigure}
    \hfill
    \begin{subfigure}[t]{0.235\textwidth}
        \centering
        \includegraphics[width=\linewidth]{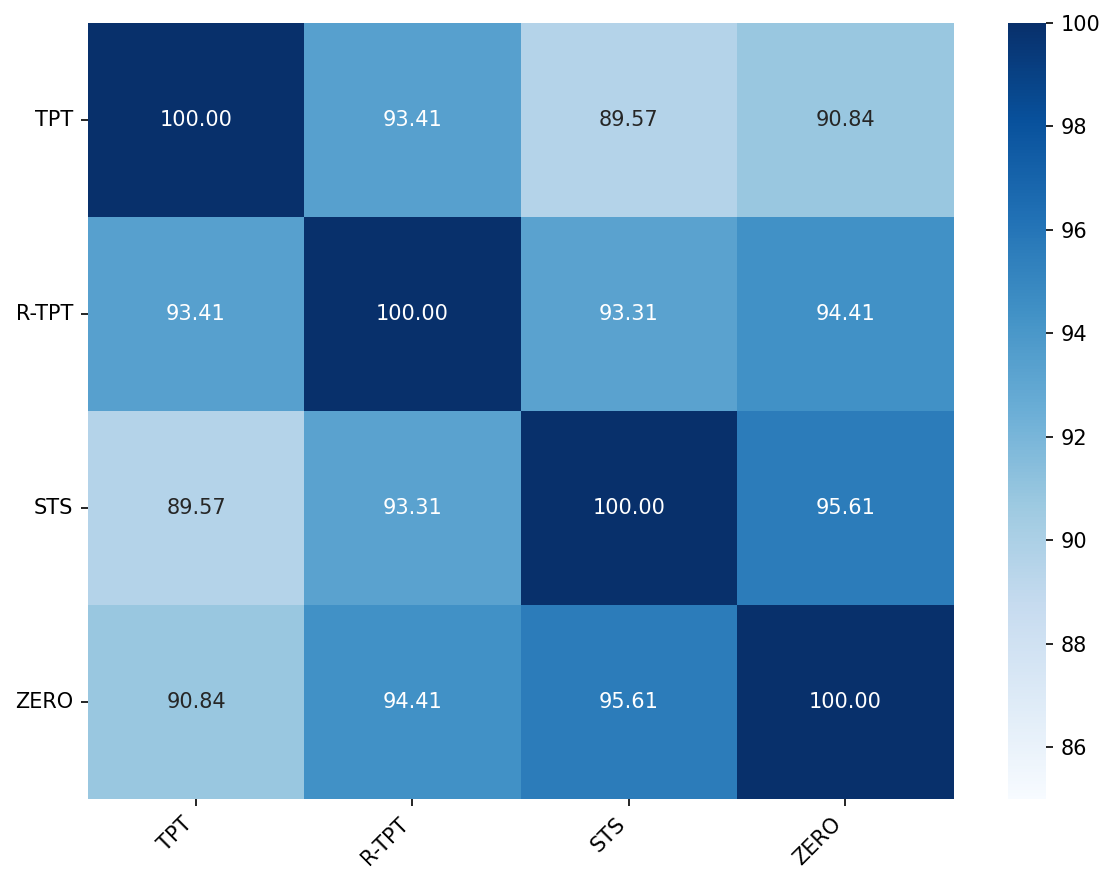}
        \caption{ImageNet-A}
    \end{subfigure}
    \hfill
    \begin{subfigure}[t]{0.235\textwidth}
        \centering
        \includegraphics[width=\linewidth]{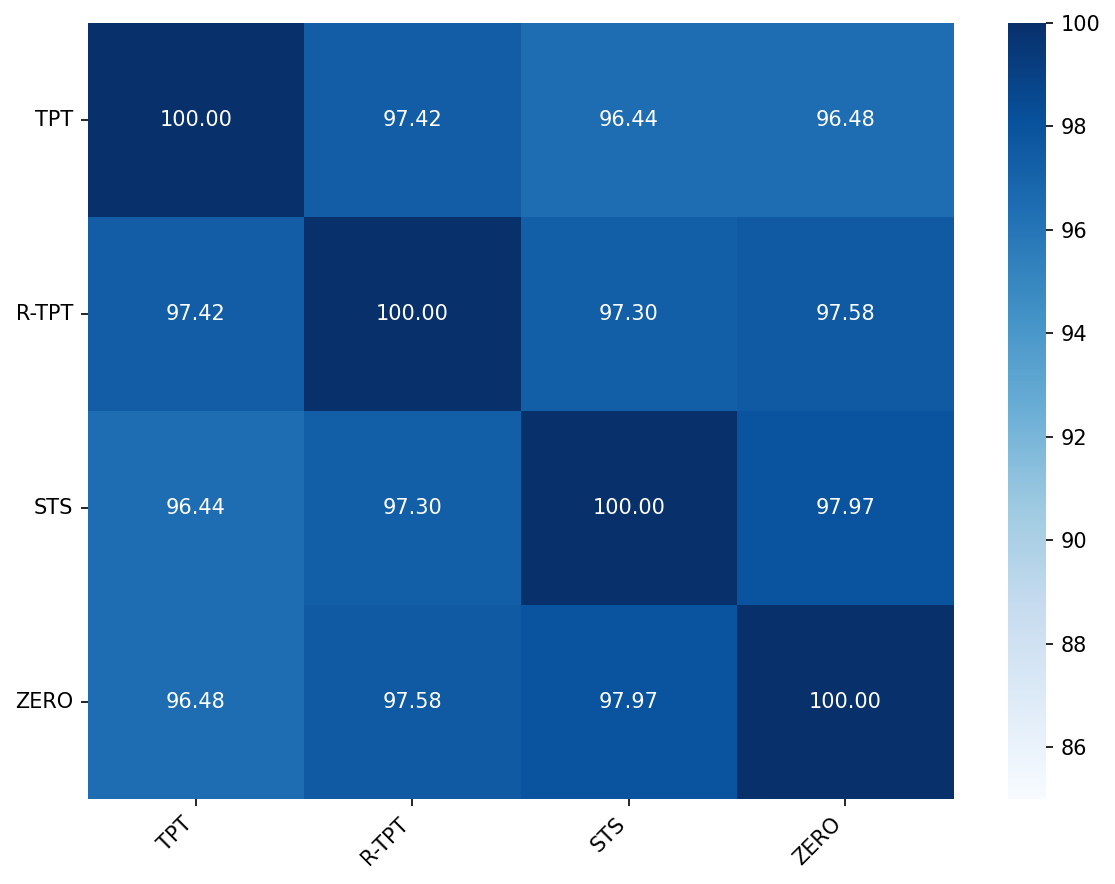}
        \caption{ImageNet-R}
    \end{subfigure}

    \vspace{0.8em}

    \begin{subfigure}[t]{0.235\textwidth}
        \centering
        \includegraphics[width=\linewidth]{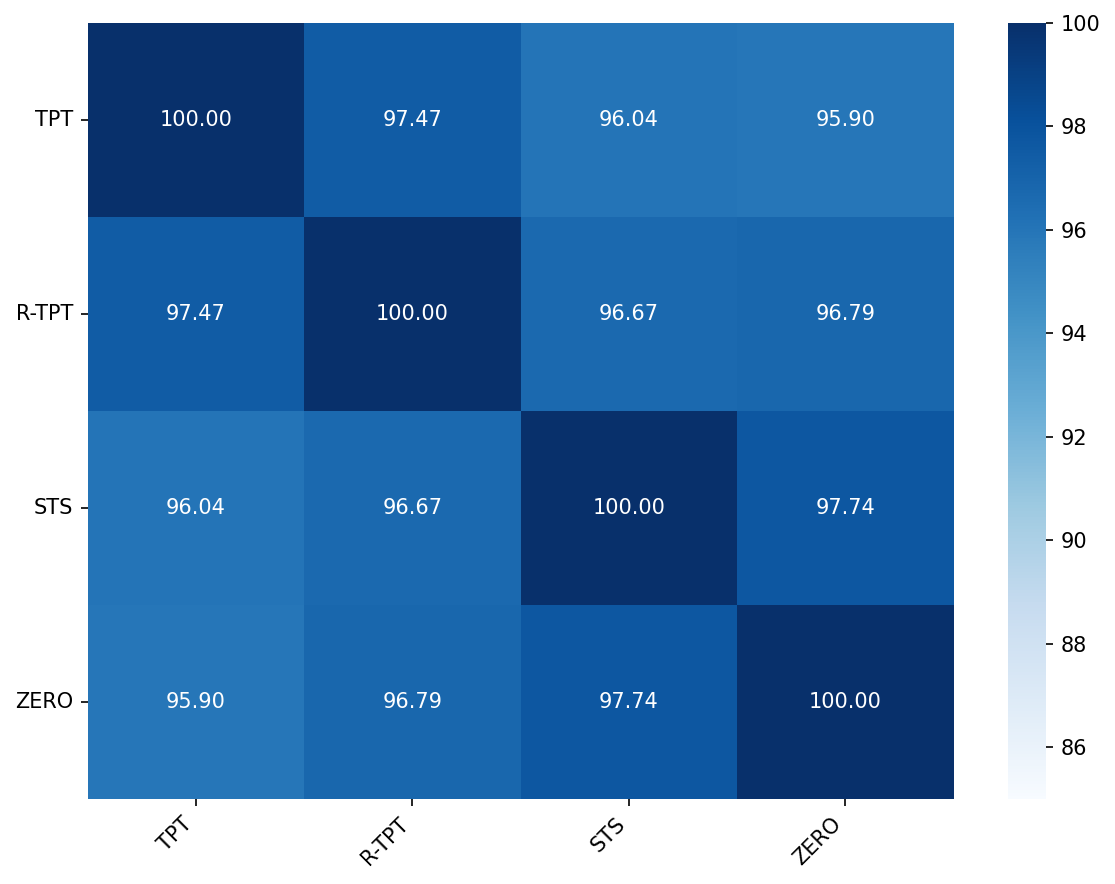}
        \caption{ImageNet-V}
    \end{subfigure}
    \hfill
    \begin{subfigure}[t]{0.235\textwidth}
        \centering
        \includegraphics[width=\linewidth]{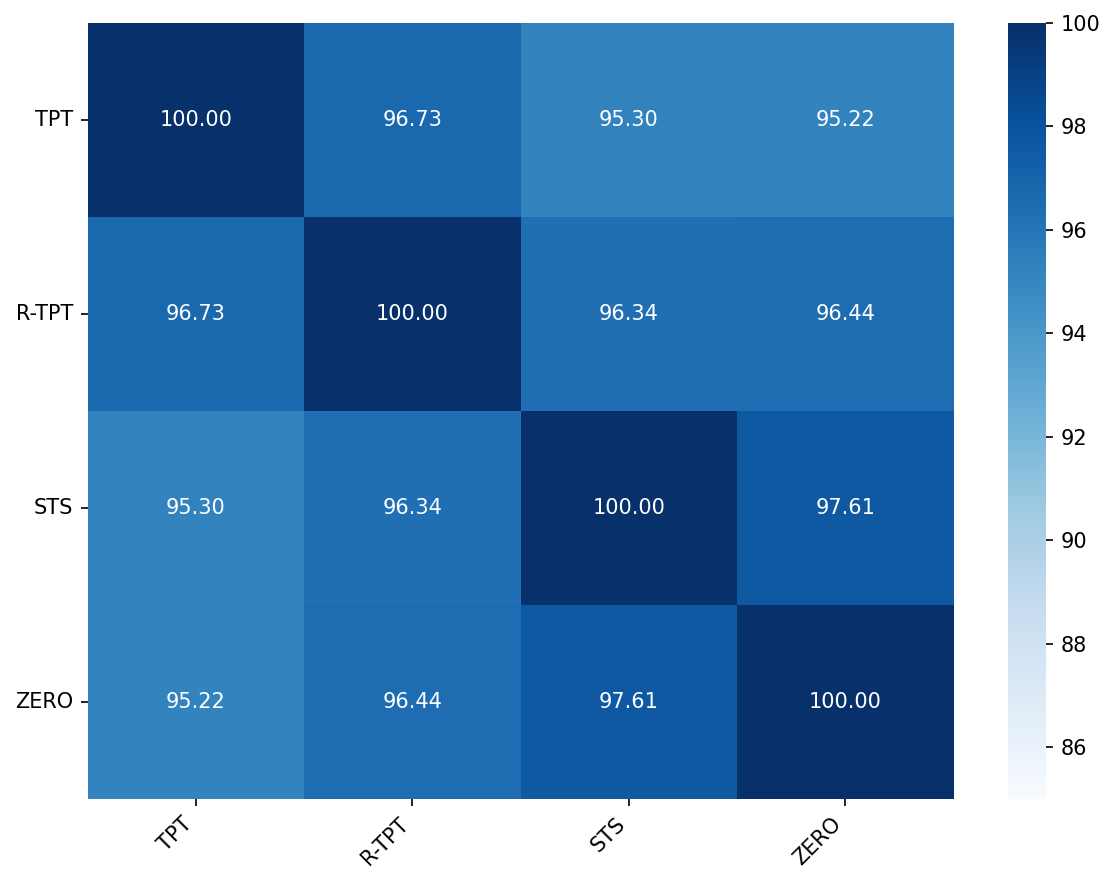}
        \caption{ImageNet-K}
    \end{subfigure}
    \hfill
    \begin{subfigure}[t]{0.235\textwidth}
        \centering
        \includegraphics[width=\linewidth]{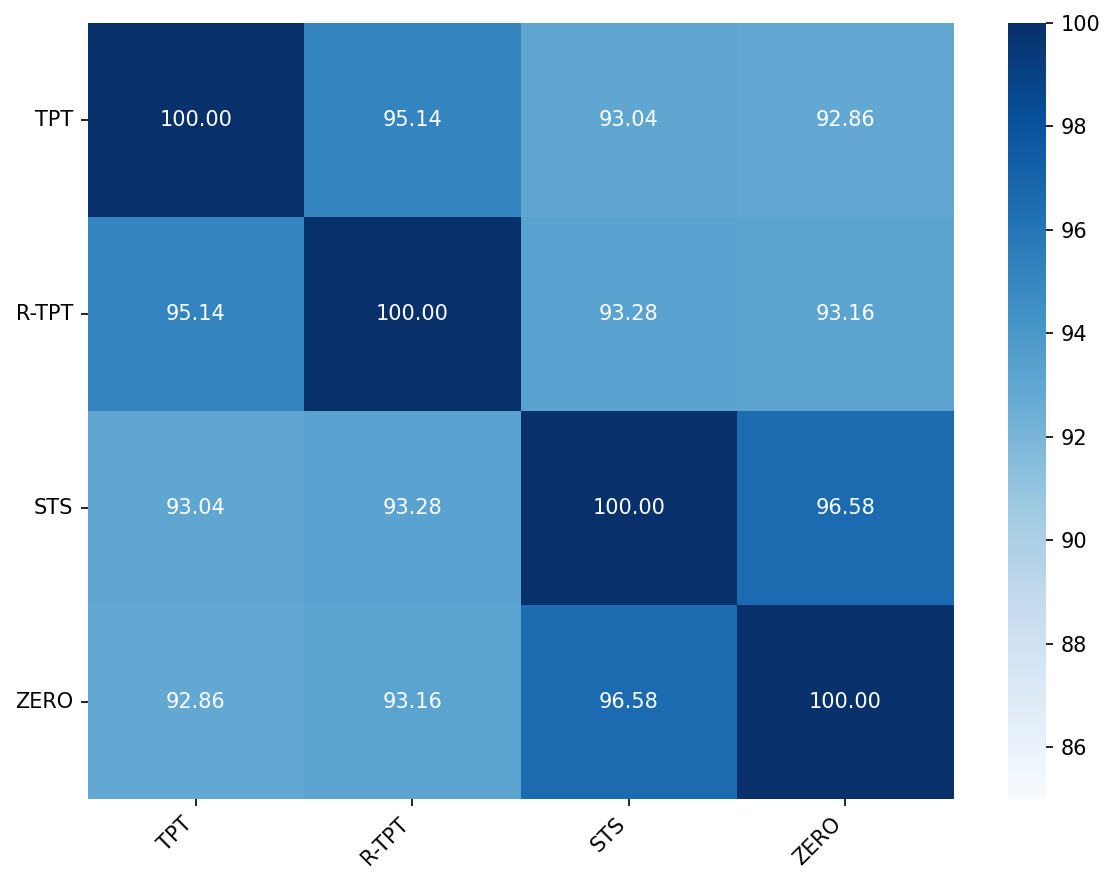}
        \caption{Aircraft}
    \end{subfigure}
    \hfill
    \begin{subfigure}[t]{0.235\textwidth}
        \centering
        \includegraphics[width=\linewidth]{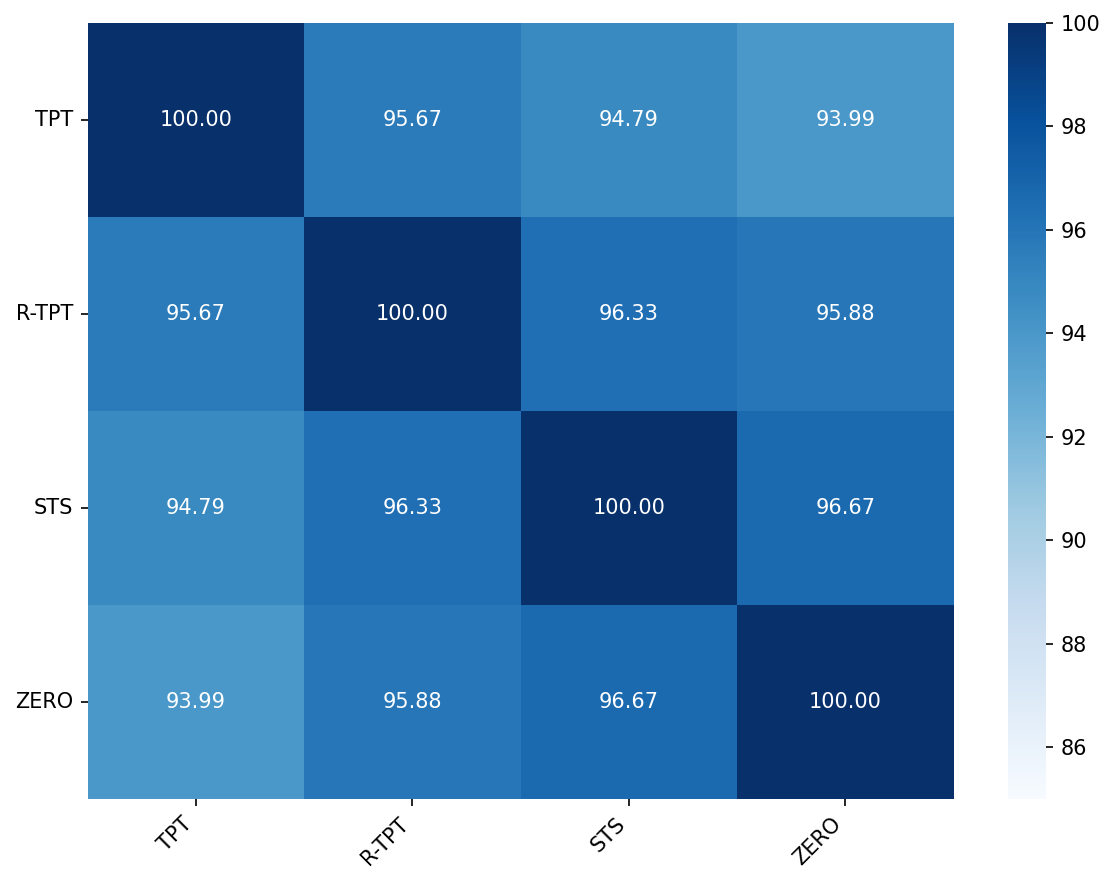}
        \caption{Cars}
    \end{subfigure}

    \vspace{0.8em}

    \begin{subfigure}[t]{0.235\textwidth}
        \centering
        \includegraphics[width=\linewidth]{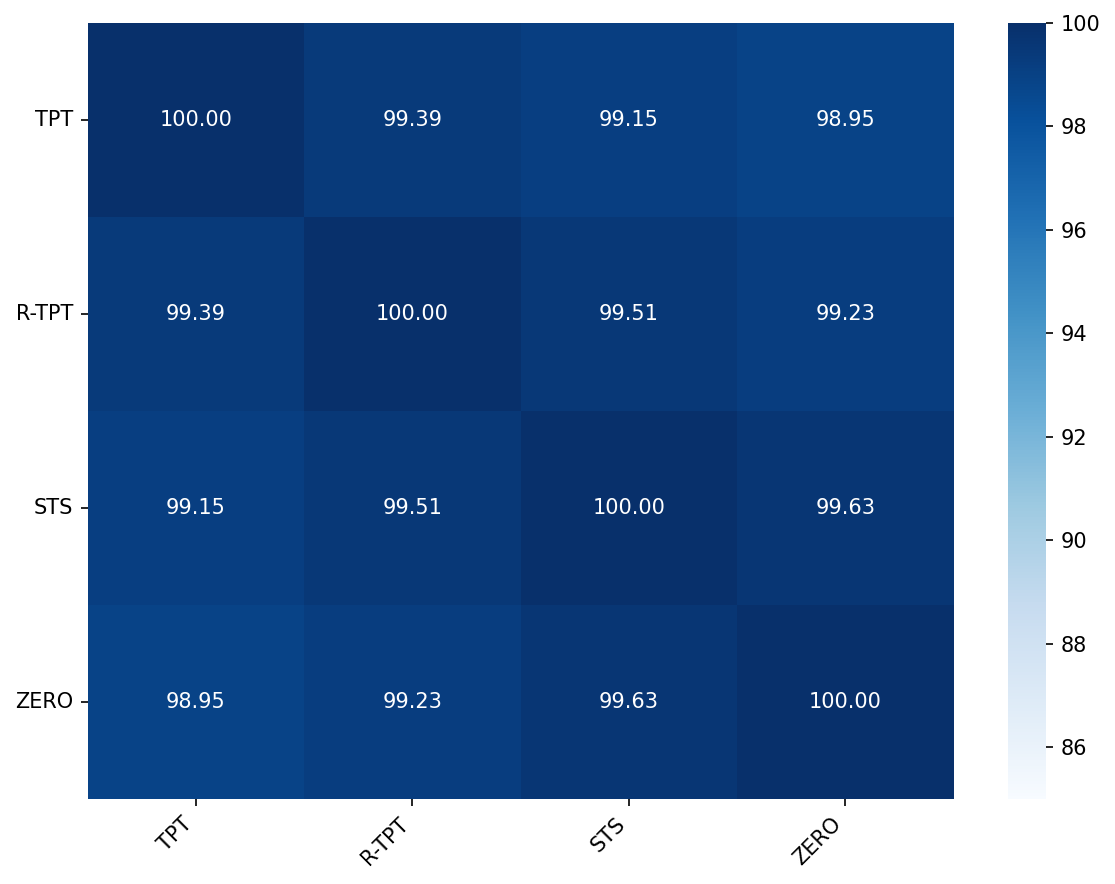}
        \caption{Caltech101}
    \end{subfigure}
    \hfill
    \begin{subfigure}[t]{0.235\textwidth}
        \centering
        \includegraphics[width=\linewidth]{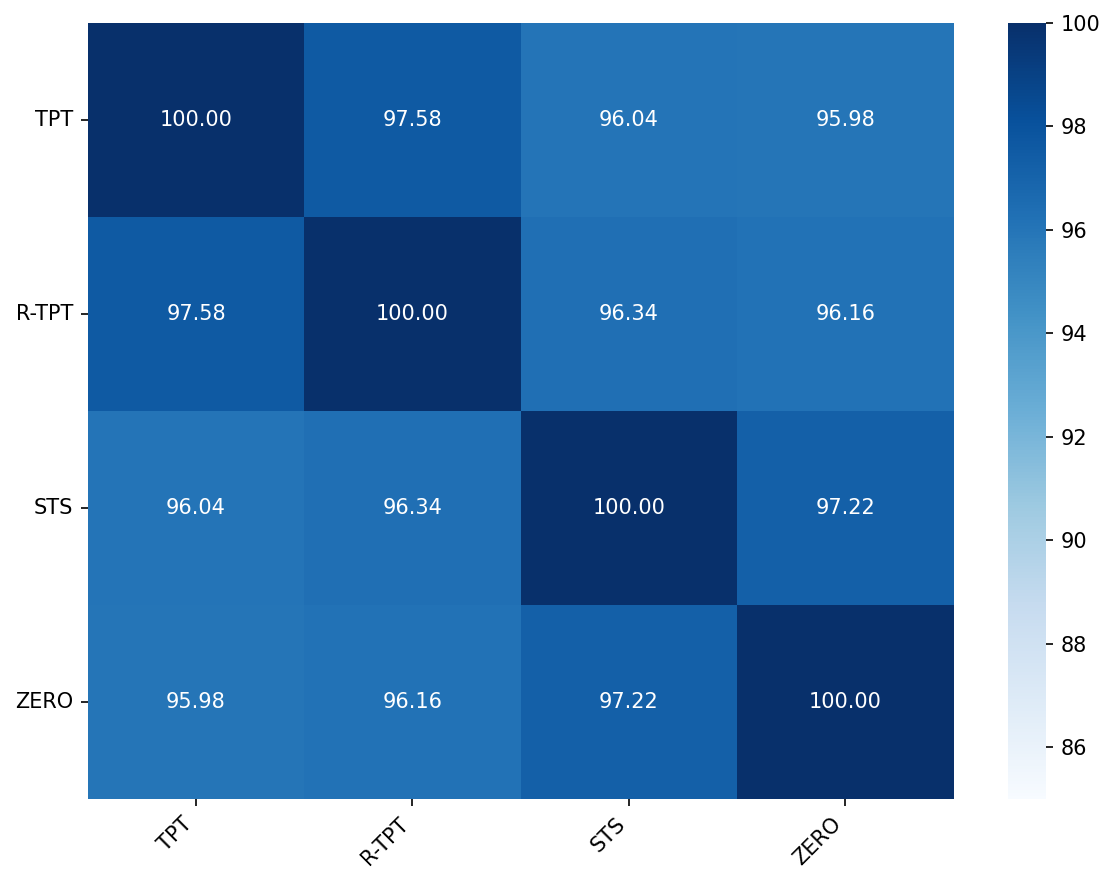}
        \caption{DTD}
    \end{subfigure}
    \hfill
    \begin{subfigure}[t]{0.235\textwidth}
        \centering
        \includegraphics[width=\linewidth]{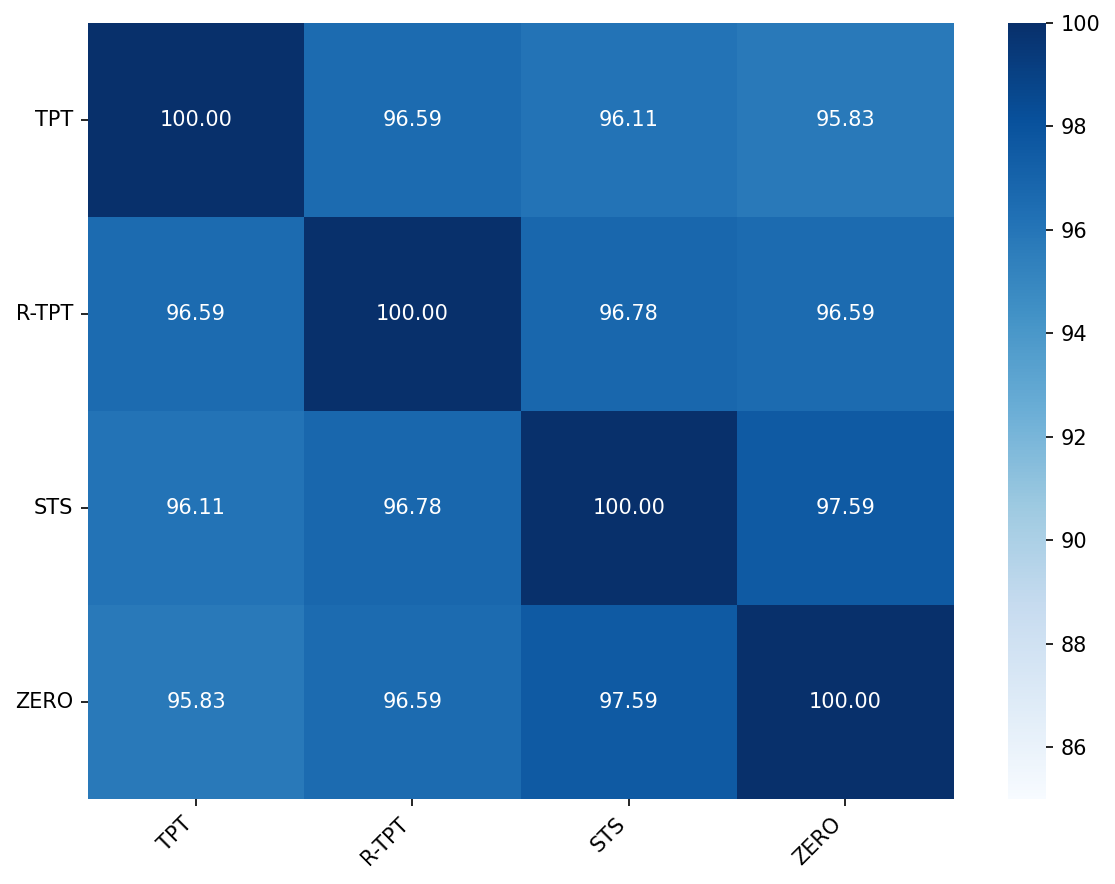}
        \caption{EuroSAT}
    \end{subfigure}
    \hfill
    \begin{subfigure}[t]{0.235\textwidth}
        \centering
        \includegraphics[width=\linewidth]{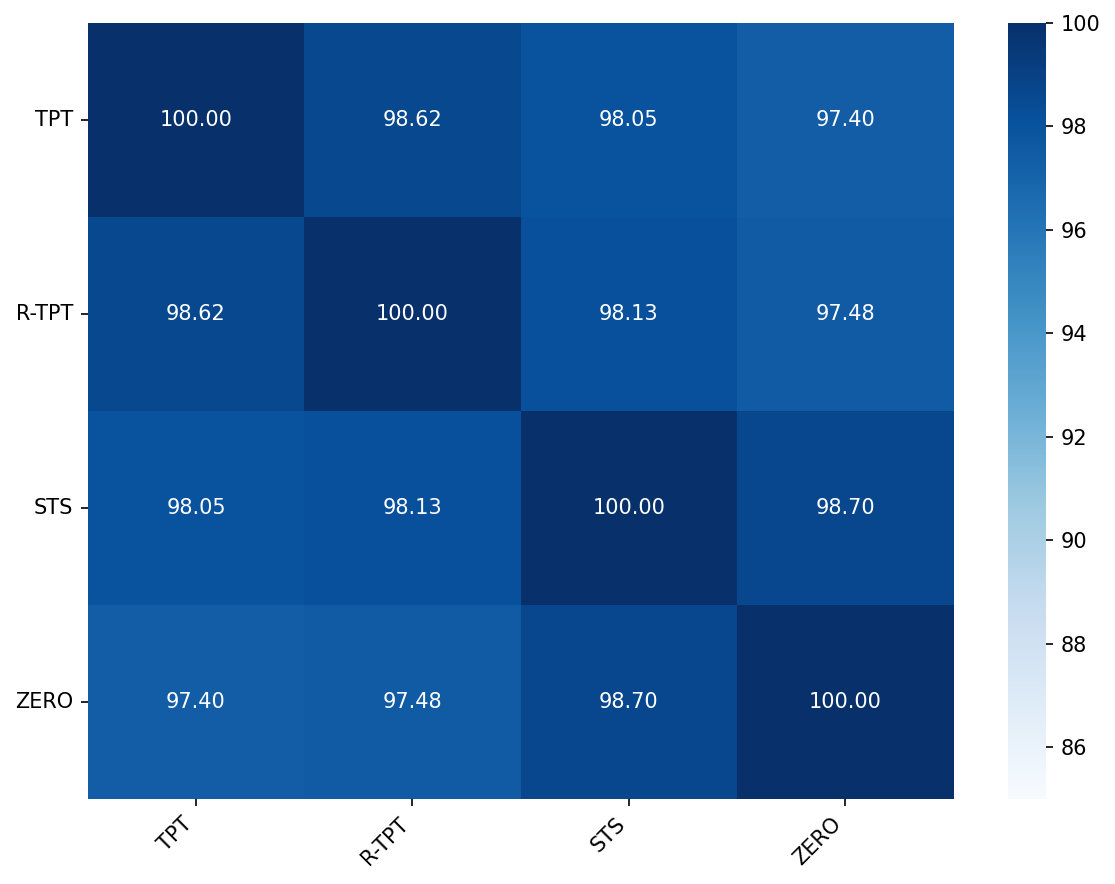}
        \caption{Flower102}
    \end{subfigure}

    \vspace{0.8em}

    \begin{subfigure}[t]{0.235\textwidth}
        \centering
        \includegraphics[width=\linewidth]{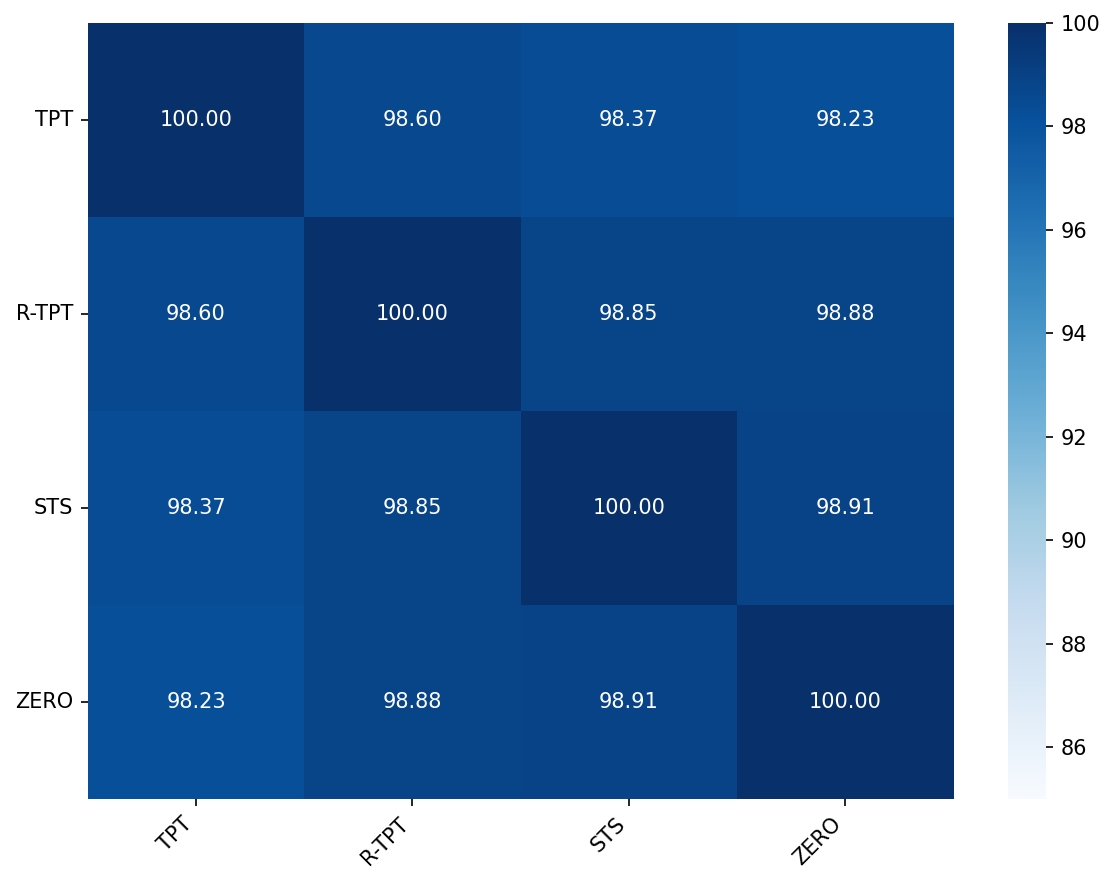}
        \caption{Food101}
    \end{subfigure}
    \hfill
    \begin{subfigure}[t]{0.235\textwidth}
        \centering
        \includegraphics[width=\linewidth]{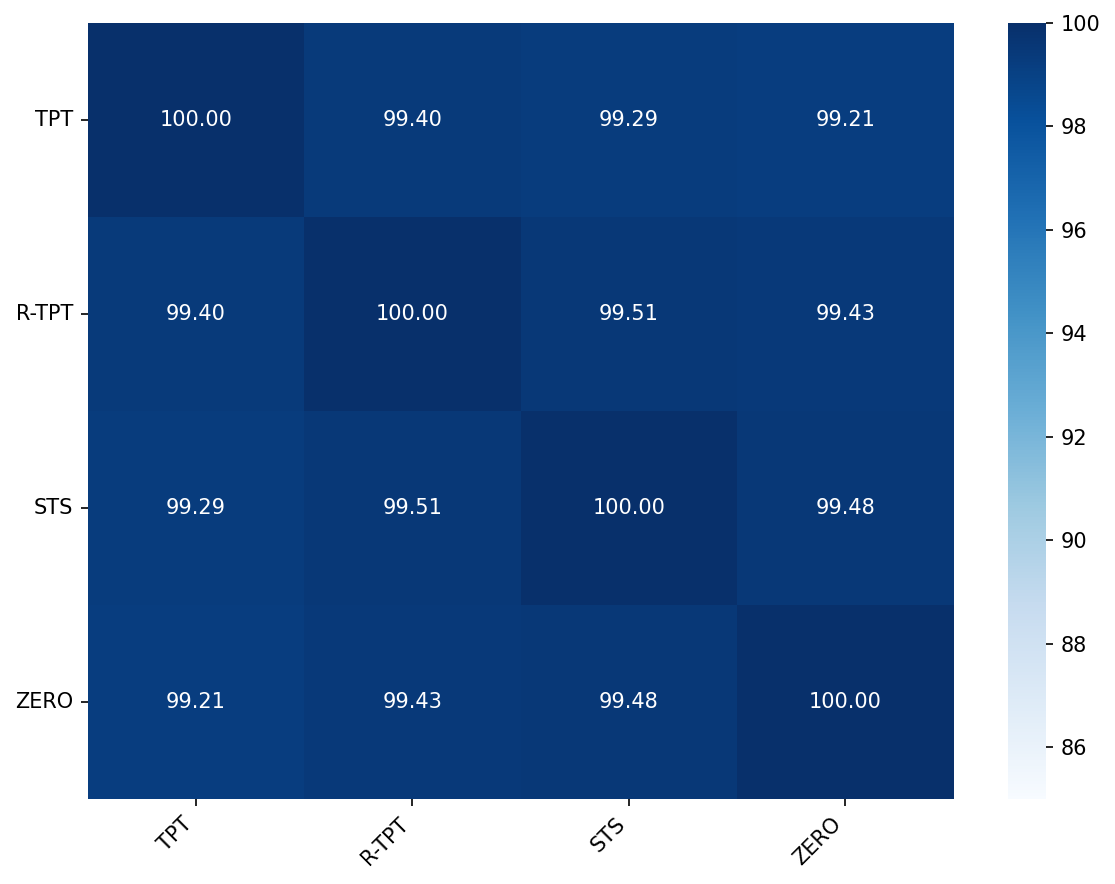}
        \caption{Pets}
    \end{subfigure}
    \hfill
    \begin{subfigure}[t]{0.235\textwidth}
        \centering
        \includegraphics[width=\linewidth]{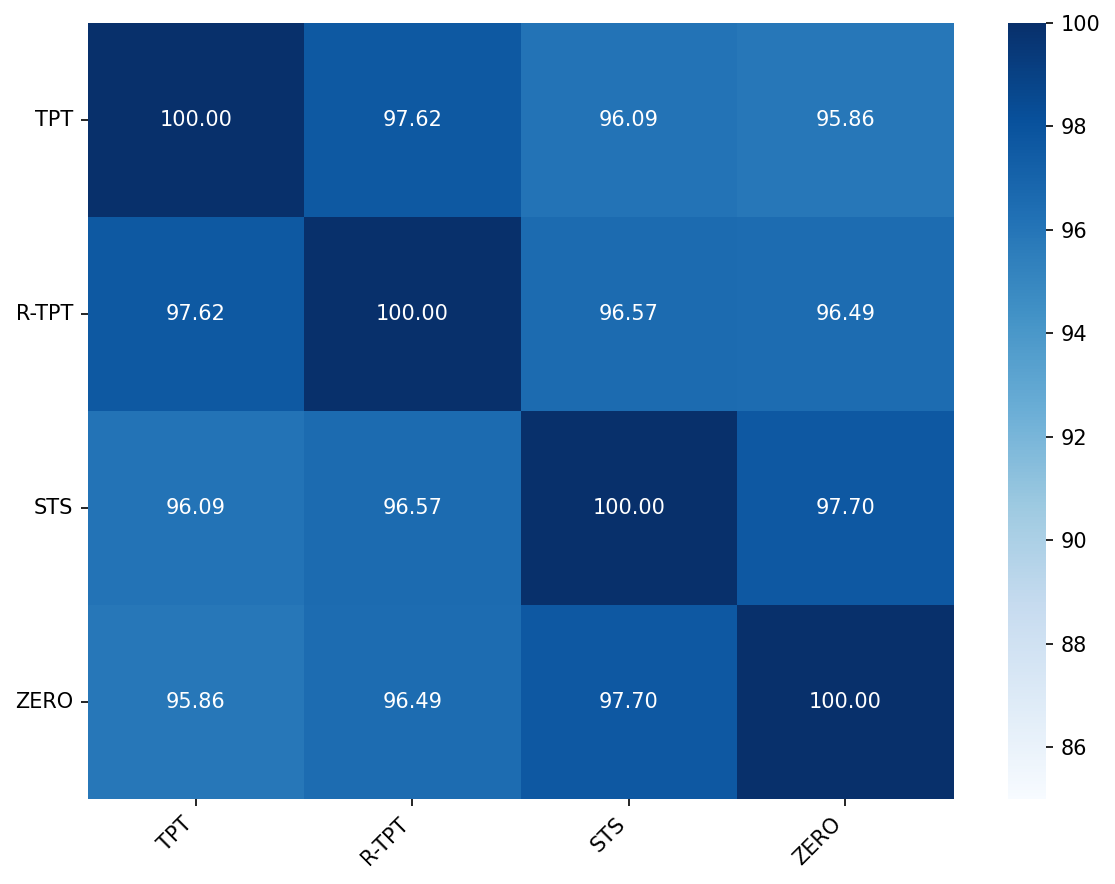}
        \caption{SUN397}
    \end{subfigure}
    \hfill
    \begin{subfigure}[t]{0.235\textwidth}
        \centering
        \includegraphics[width=\linewidth]{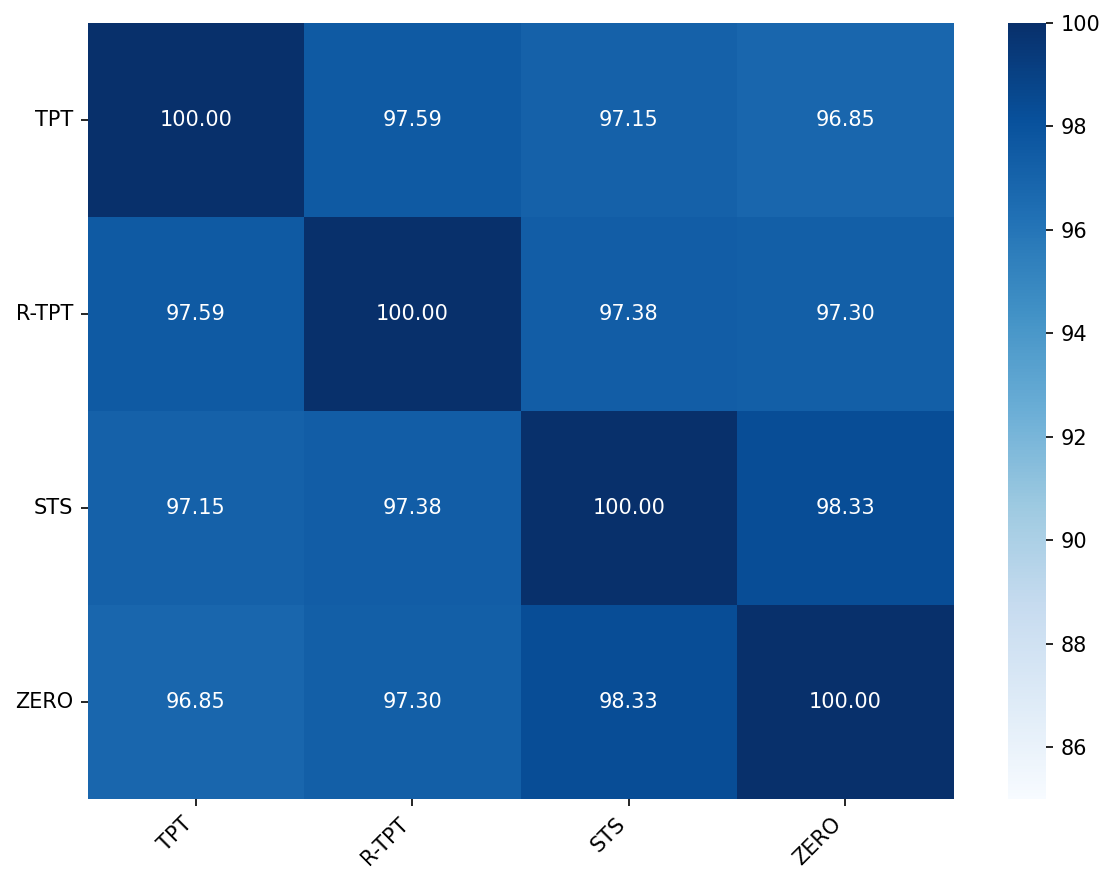}
        \caption{UCF101}
    \end{subfigure}
    \caption{
The figure presents the Hamming similarity of effective and ineffective adaptation cases across different TTA methods. 
Higher similarity indicates stronger agreement among methods on whether a sample requires adaptation.
}
    \label{fig:hamming_all_4X4}
\end{figure*}

To examine the similarity of selective adaptation behaviors across different TTA methods, we analyze the detection ground truth of effective and ineffective cases in TPT~\cite{shu2022testtime}, R-TPT~\cite{sheng2025r}, STS~\cite{dafnis2025testtime}, and ZERO~\cite{farina2024frustratingly} using Hamming distance.
For each method, Wrong to Correct is labeled as beneficial (0), while Correct to Correct, Wrong to Wrong, and Correct to Wrong are treated as ineffective (1).
Given two methods $M_1$ and $M_2$, the Hamming similarity is defined as the percentage of test samples for which the two methods assign the same binary label. 
The corresponding Hamming Similarity is defined as $1 - \text{Distance}$.
As shown in Fig.~\ref{fig:hamming_all_4X4}, the Hamming similarity across all evaluated benchmarks consistently exceeds 90\%, indicating that different TTA methods exhibit highly consistent selective adaptation behaviors. 
In particular, the ineffective cases are largely shared across methods, meaning that most samples are consistently identified as not requiring adaptation. 
This observation suggests that the necessity of adaptation is largely determined by the sample itself rather than the specific TTA algorithm.
Consequently, many adaptation operations performed by existing methods are redundant, highlighting the importance of selectively applying adaptation only to samples that can truly benefit from it.

\section{Statistics of Adaptation Cases}
\begin{figure*}[p]
    \centering
    \begin{subfigure}[t]{0.24\textwidth}
        \centering
        \includegraphics[width=\linewidth]{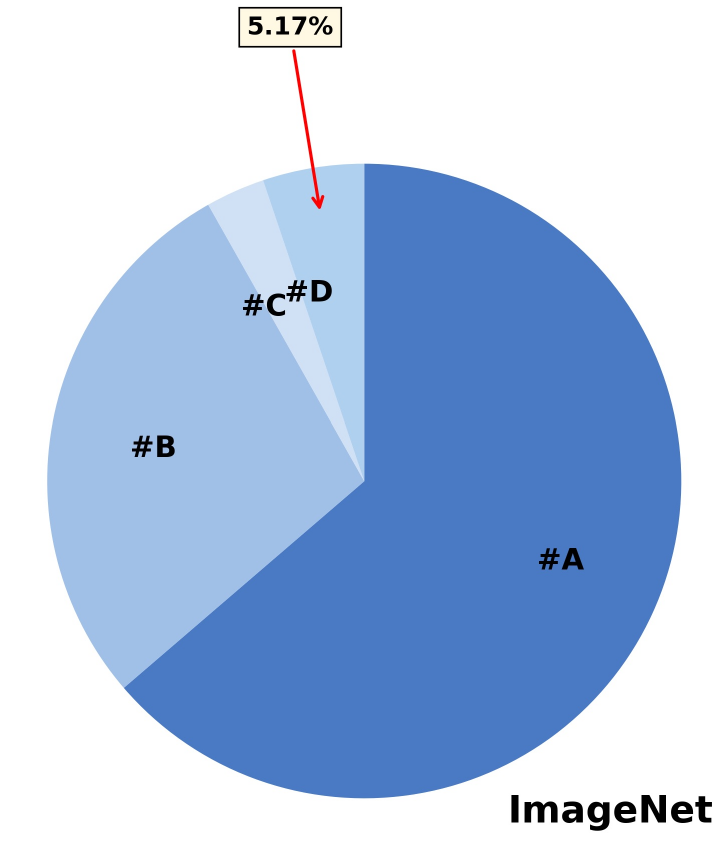}
        \caption{ImageNet}
    \end{subfigure}
    \hfill
    \begin{subfigure}[t]{0.24\textwidth}
        \centering
        \includegraphics[width=\linewidth]{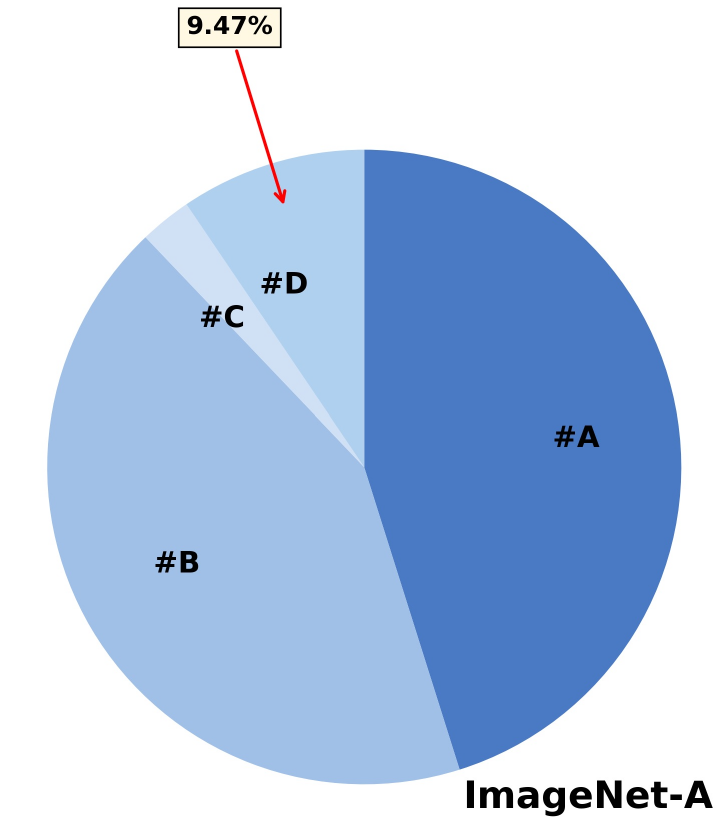}
        \caption{ImageNet-A}
    \end{subfigure}
    \hfill
    \begin{subfigure}[t]{0.24\textwidth}
        \centering
        \includegraphics[width=\linewidth]{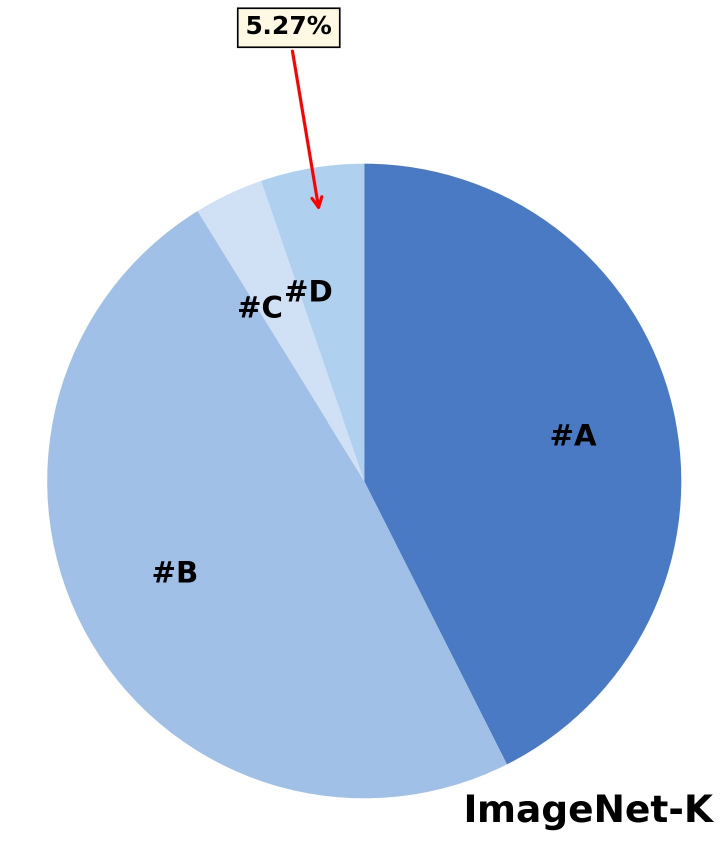}
        \caption{ImageNet-K}
    \end{subfigure}
    \hfill
    \begin{subfigure}[t]{0.24\textwidth}
        \centering
        \includegraphics[width=\linewidth]{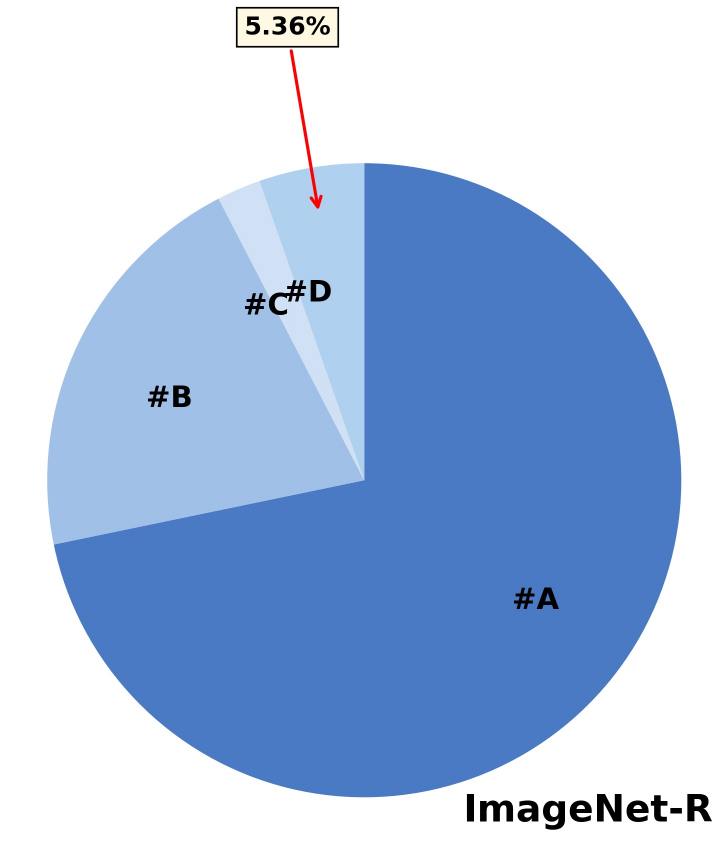}
        \caption{ImageNet-R}
    \end{subfigure}

    \vspace{0.5em}

    \begin{subfigure}[t]{0.24\textwidth}
        \centering
        \includegraphics[width=\linewidth]{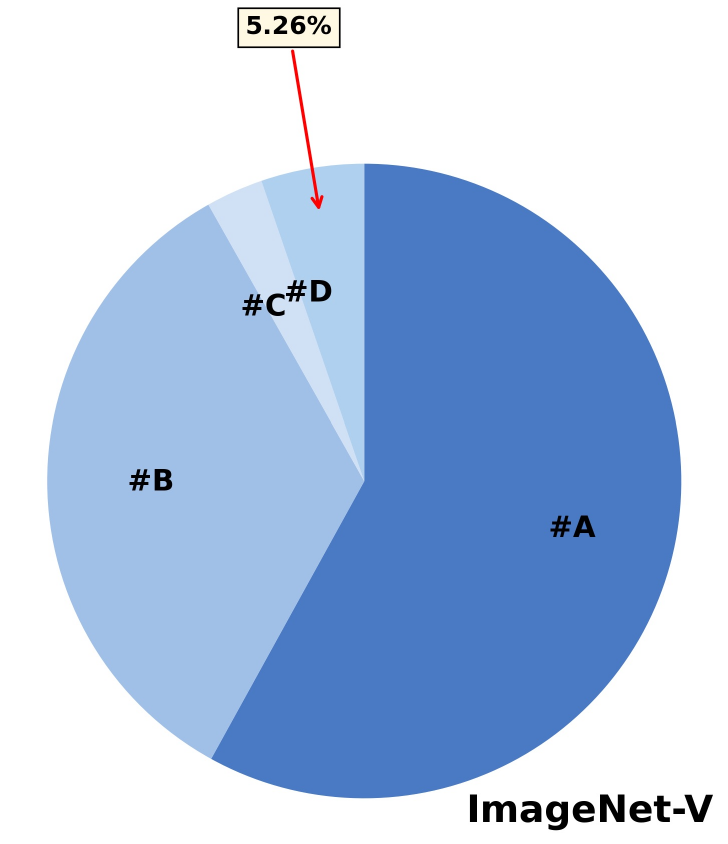}
        \caption{ImageNet-V}
    \end{subfigure}
    \hfill
    \begin{subfigure}[t]{0.24\textwidth}
        \centering
        \includegraphics[width=\linewidth]{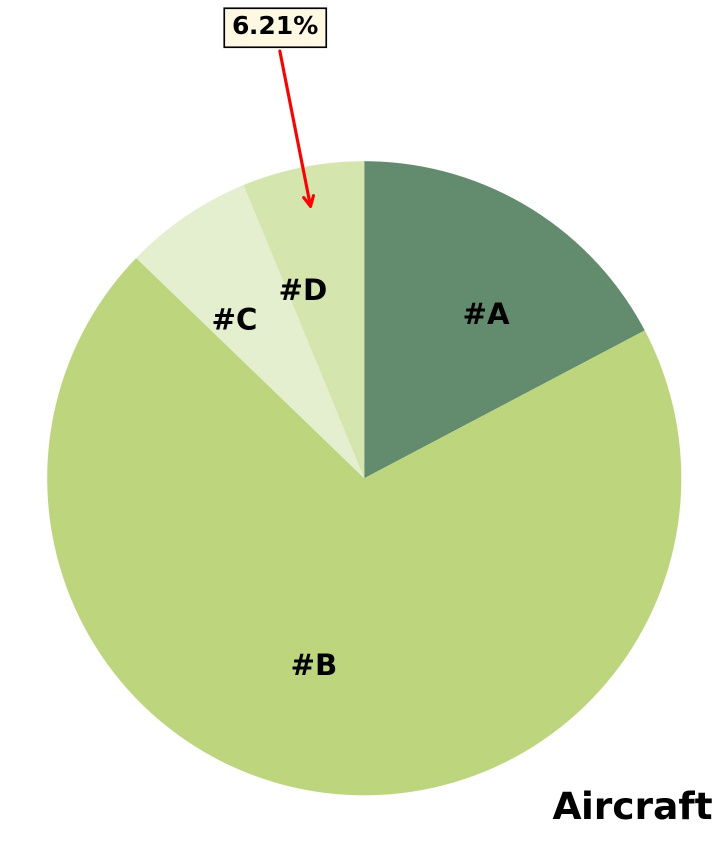}
        \caption{Aircraft}
    \end{subfigure}
    \hfill
    \begin{subfigure}[t]{0.24\textwidth}
        \centering
        \includegraphics[width=\linewidth]{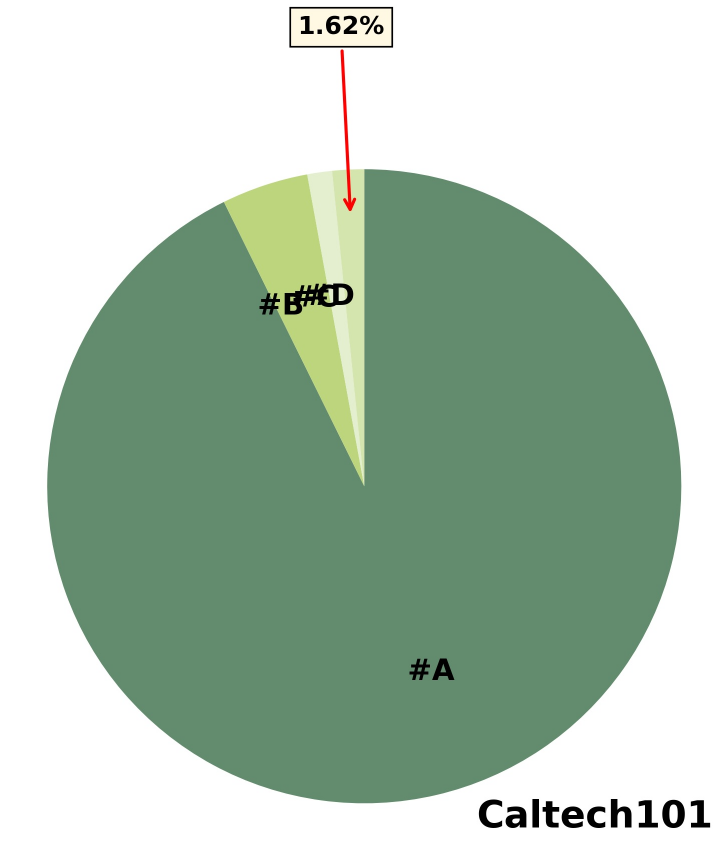}
        \caption{Caltech101}
    \end{subfigure}
    \hfill
    \begin{subfigure}[t]{0.24\textwidth}
        \centering
        \includegraphics[width=\linewidth]{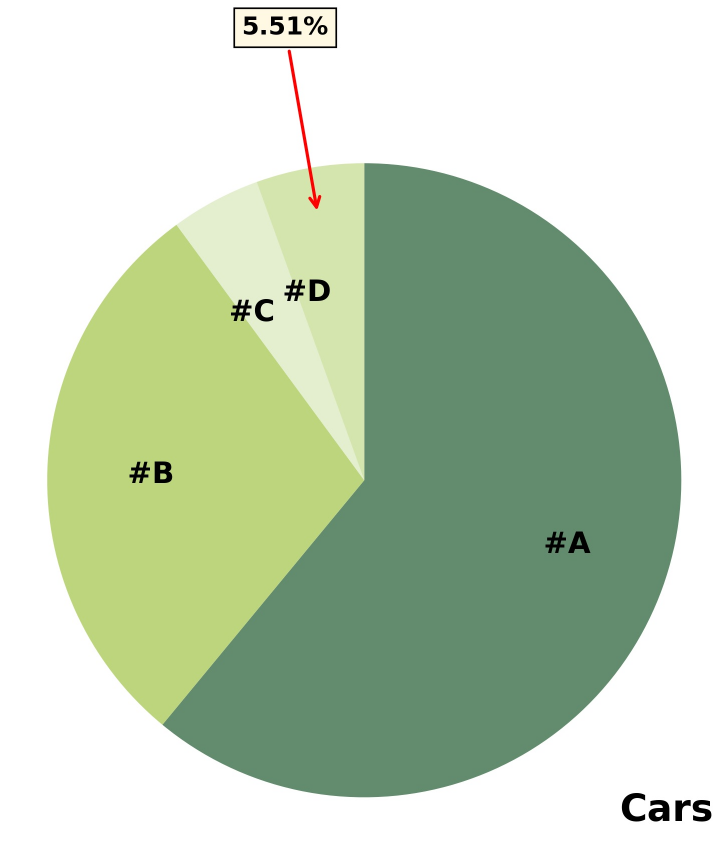}
        \caption{Cars}
    \end{subfigure}

    \vspace{0.5em}

    \begin{subfigure}[t]{0.24\textwidth}
        \centering
        \includegraphics[width=\linewidth]{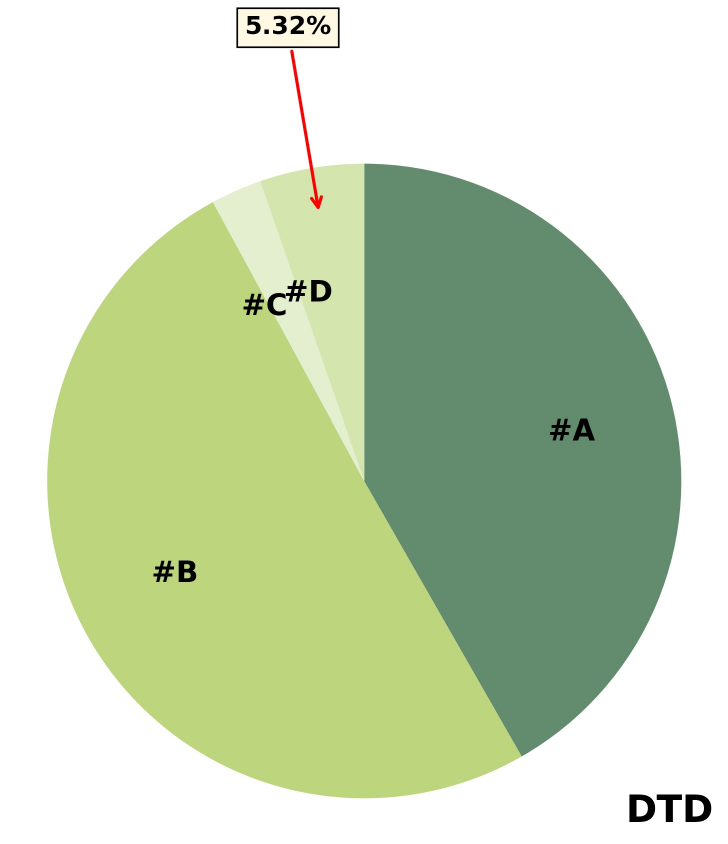}
        \caption{DTD}
    \end{subfigure}
    \hfill
    \begin{subfigure}[t]{0.24\textwidth}
        \centering
        \includegraphics[width=\linewidth]{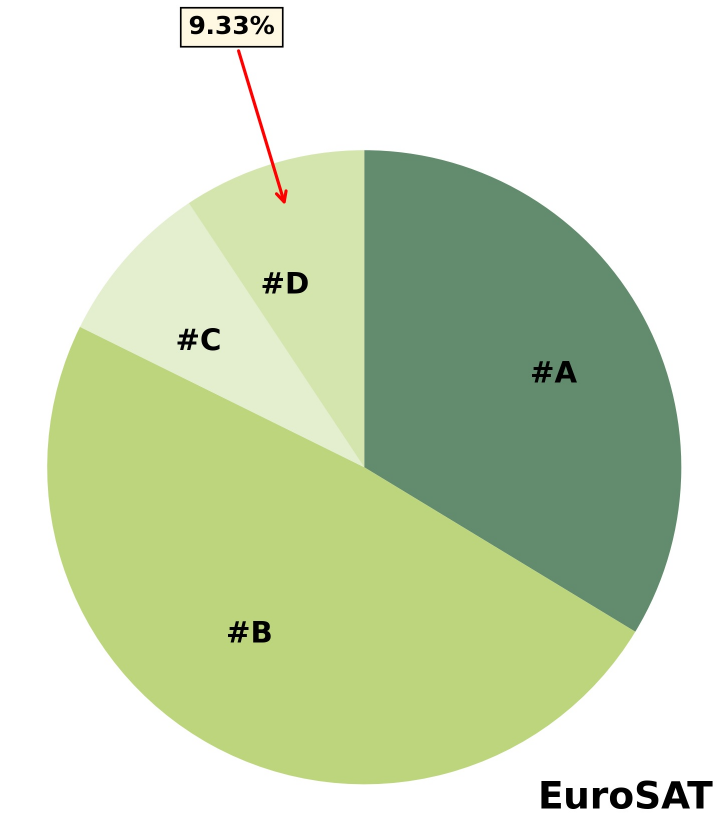}
        \caption{EuroSAT}
    \end{subfigure}
    \hfill
    \begin{subfigure}[t]{0.24\textwidth}
        \centering
        \includegraphics[width=\linewidth]{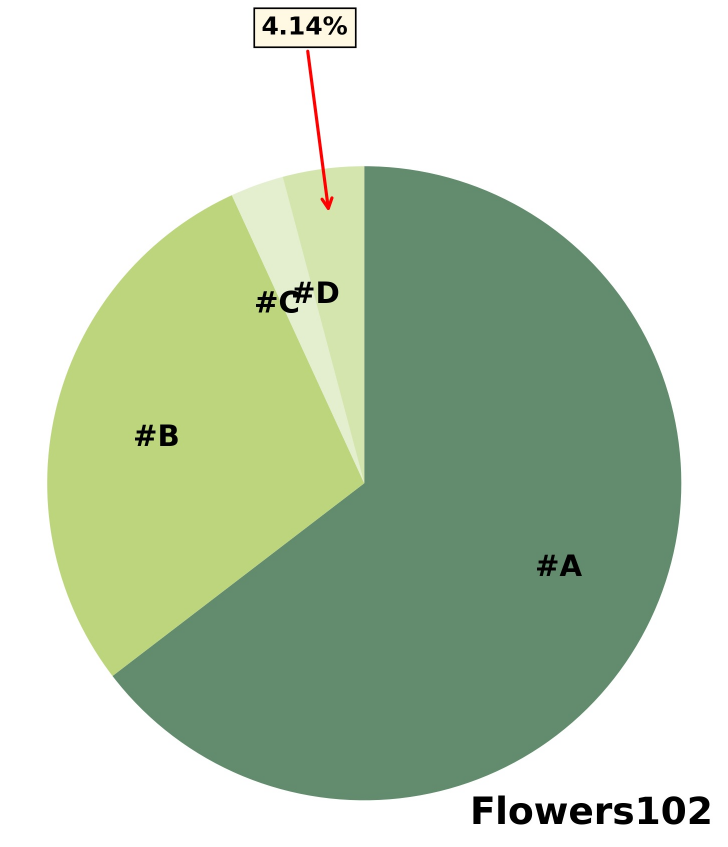}
        \caption{Flowers102}
    \end{subfigure}
    \hfill
    \begin{subfigure}[t]{0.24\textwidth}
        \centering
        \includegraphics[width=\linewidth]{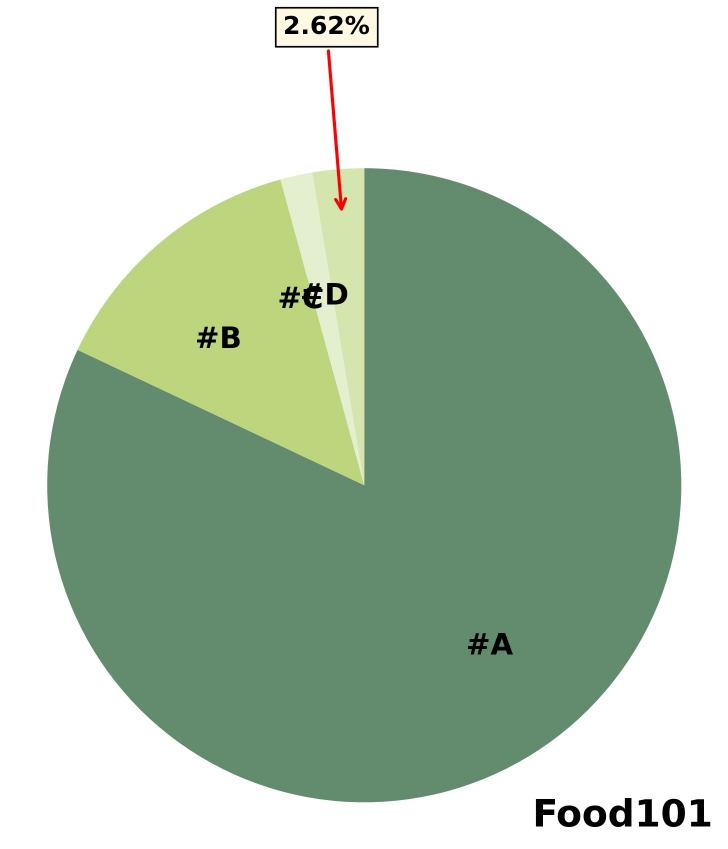}
        \caption{Food101}
    \end{subfigure}

    \vspace{0.5em}

    \begin{subfigure}[t]{0.24\textwidth}
        \centering
        \includegraphics[width=\linewidth]{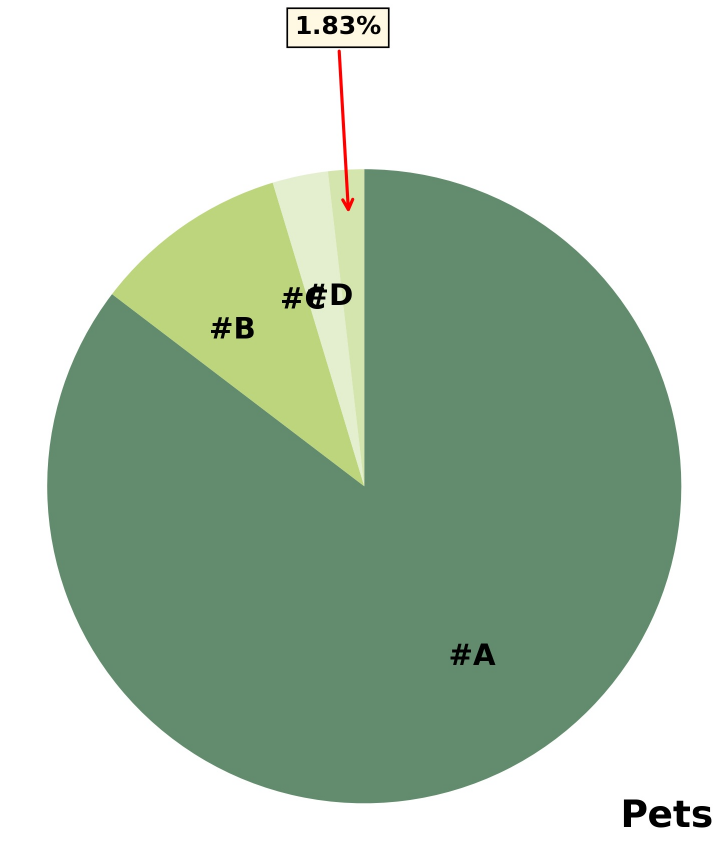}
        \caption{Pets}
    \end{subfigure}
    \hfill
    \begin{subfigure}[t]{0.24\textwidth}
        \centering
        \includegraphics[width=\linewidth]{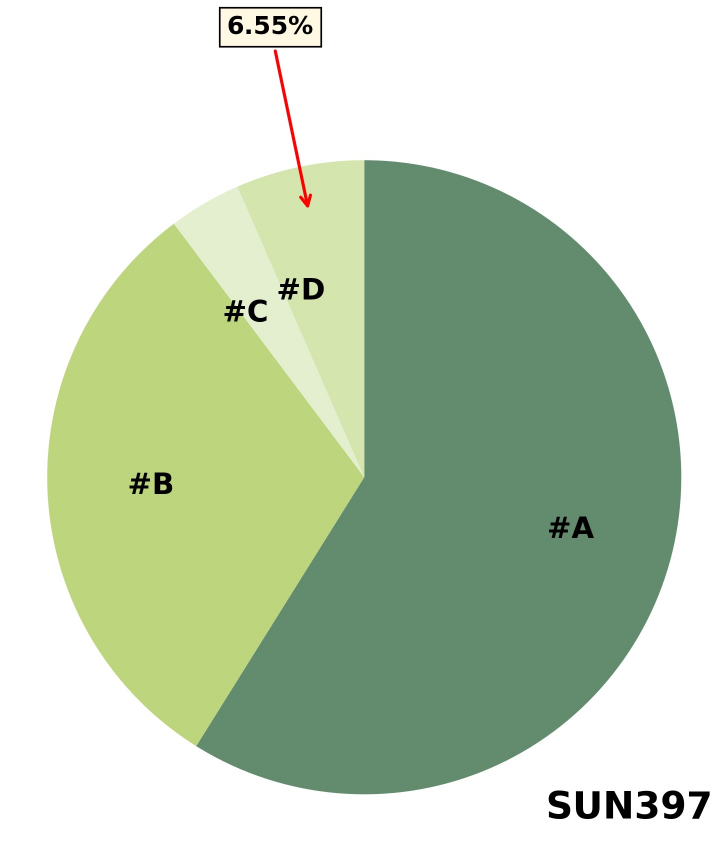}
        \caption{SUN397}
    \end{subfigure}
    \hfill
    \begin{subfigure}[t]{0.24\textwidth}
        \centering
        \includegraphics[width=\linewidth]{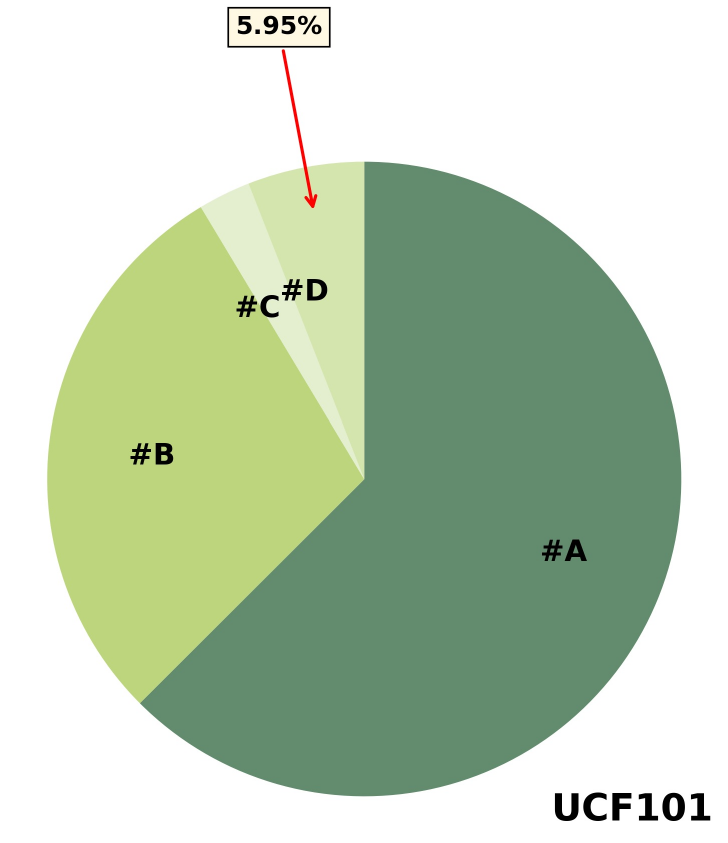}
        \caption{UCF101}
    \end{subfigure}
    \hfill
    \begin{subfigure}[t]{0.24\textwidth}
        \centering
        \includegraphics[width=\linewidth]{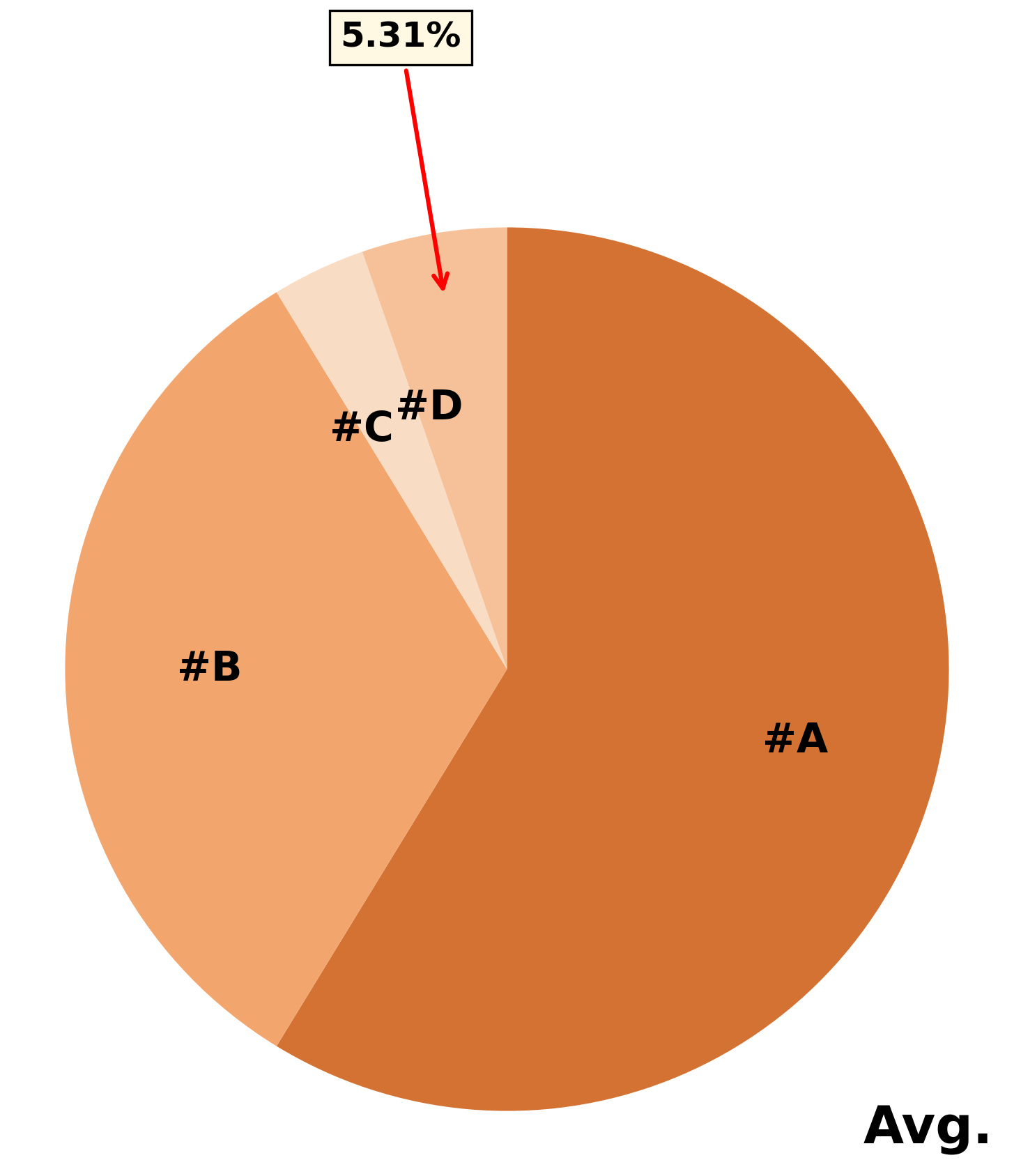} 
        \caption{Average}
    \end{subfigure}

    \caption{
The figure illustrates the distribution of four adaptation cases across various datasets under TPT~\cite{shu2022testtime}, including \#A (Correct to Correct), \#B (Wrong to Wrong), \#C (Correct to Wrong), and \#D (Wrong to Correct).
The percentage of the beneficial case \#D is highlighted.
}
    \label{fig:tpt_case_pie_all}
\end{figure*}
To further understand the behavior of test-time adaptation, we analyze the distribution of four adaptation cases across different datasets under TPT. 
Figure~\ref{fig:tpt_case_pie_all} presents the distribution of four adaptation cases across 15 benchmarks. 
Cases \#A (Correct to Correct) and \#B (Wrong to Wrong) are categorized as negligible cases, where adaptation does not change the prediction outcome. 
Case \#C (Correct to Wrong) represents harmful adaptation, while Case \#D (Wrong to Correct) corresponds to beneficial adaptation. 
Notably, only Case \#D reflects genuinely effective adaptation, whereas the remaining three cases are ineffective and do not require adaptation. 
As shown in the figure, the majority of samples fall into the negligible cases (\#A and \#B) across all datasets, while harmful and beneficial transitions occur only in a small fraction of instances. 
In particular, beneficial adaptations account for only a small percentage, typically around 5\% or lower, indicating that most adaptation processes are unnecessary and contribute little to performance improvement.

\begin{figure}[!th]
    \centering
    \begin{subfigure}[b]{0.38\textwidth}
        \centering
        \includegraphics[width=\textwidth]{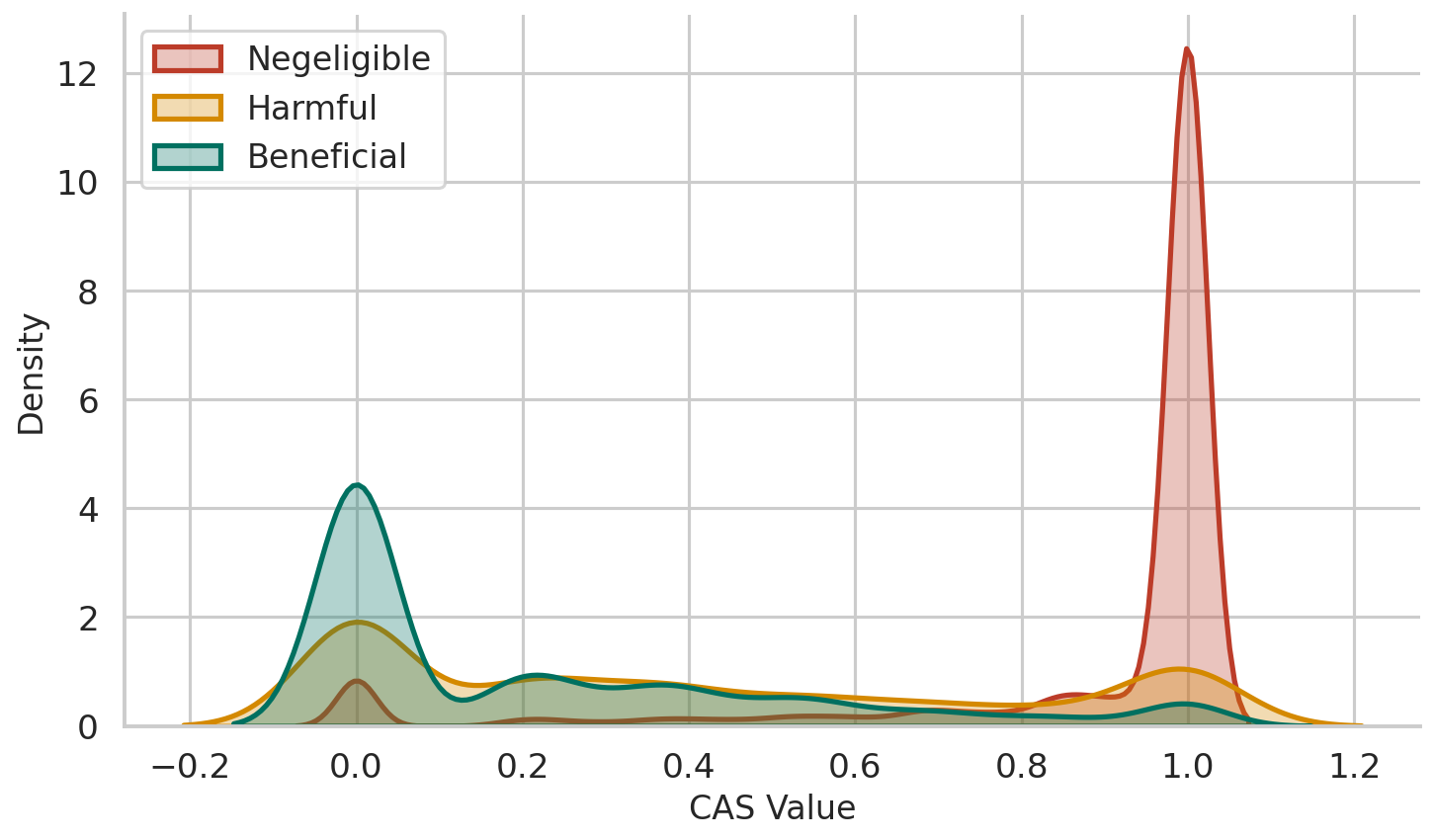}
        \caption{ImageNet}
    \end{subfigure}
   \hspace{10pt}
    \begin{subfigure}[b]{0.38\textwidth}
        \centering
        \includegraphics[width=\textwidth]{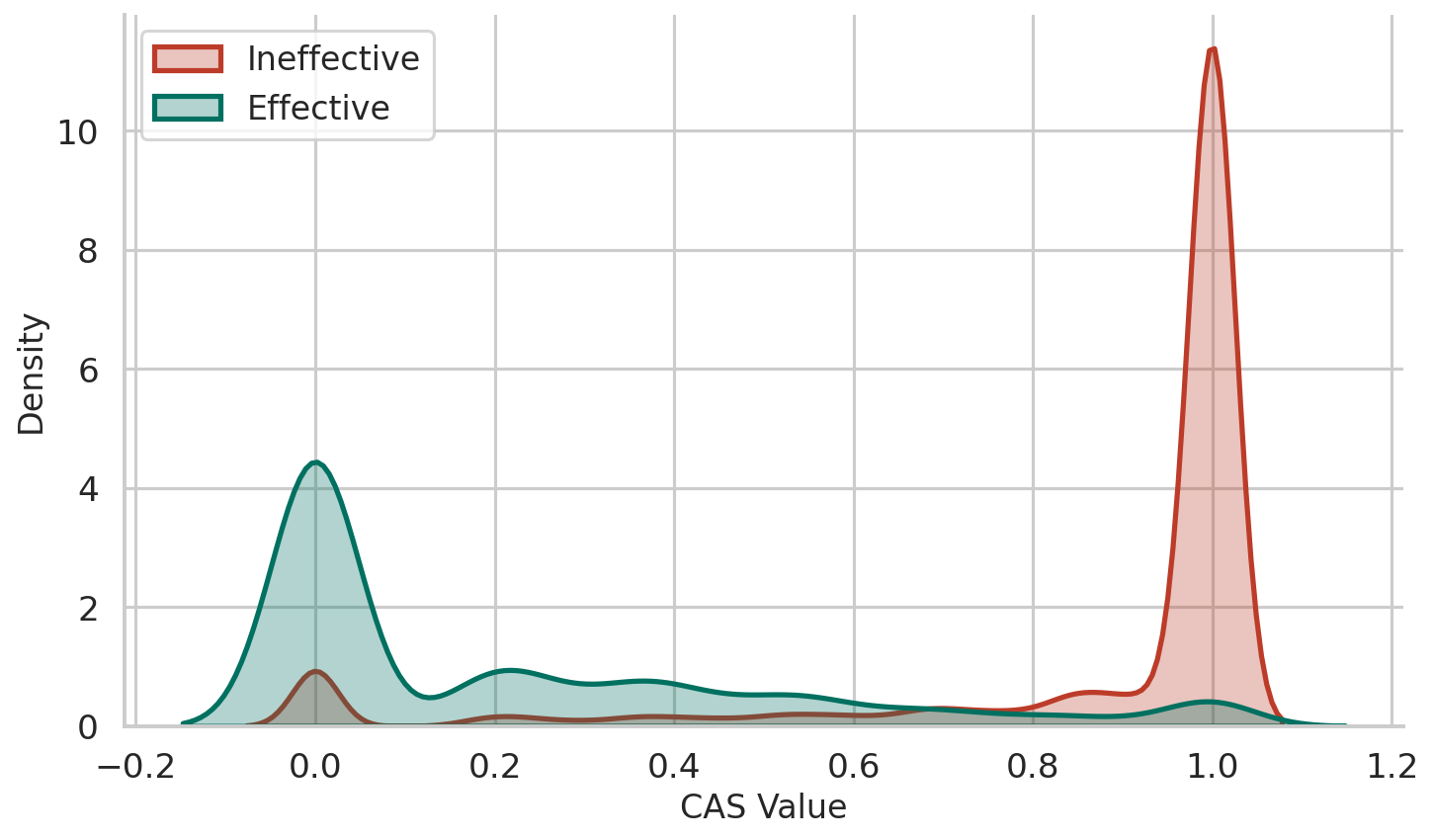}
        \caption{ImageNet}
    \end{subfigure}

    \vspace{10pt} 

    \begin{subfigure}[b]{0.38\textwidth}
        \centering
        \includegraphics[width=\textwidth]{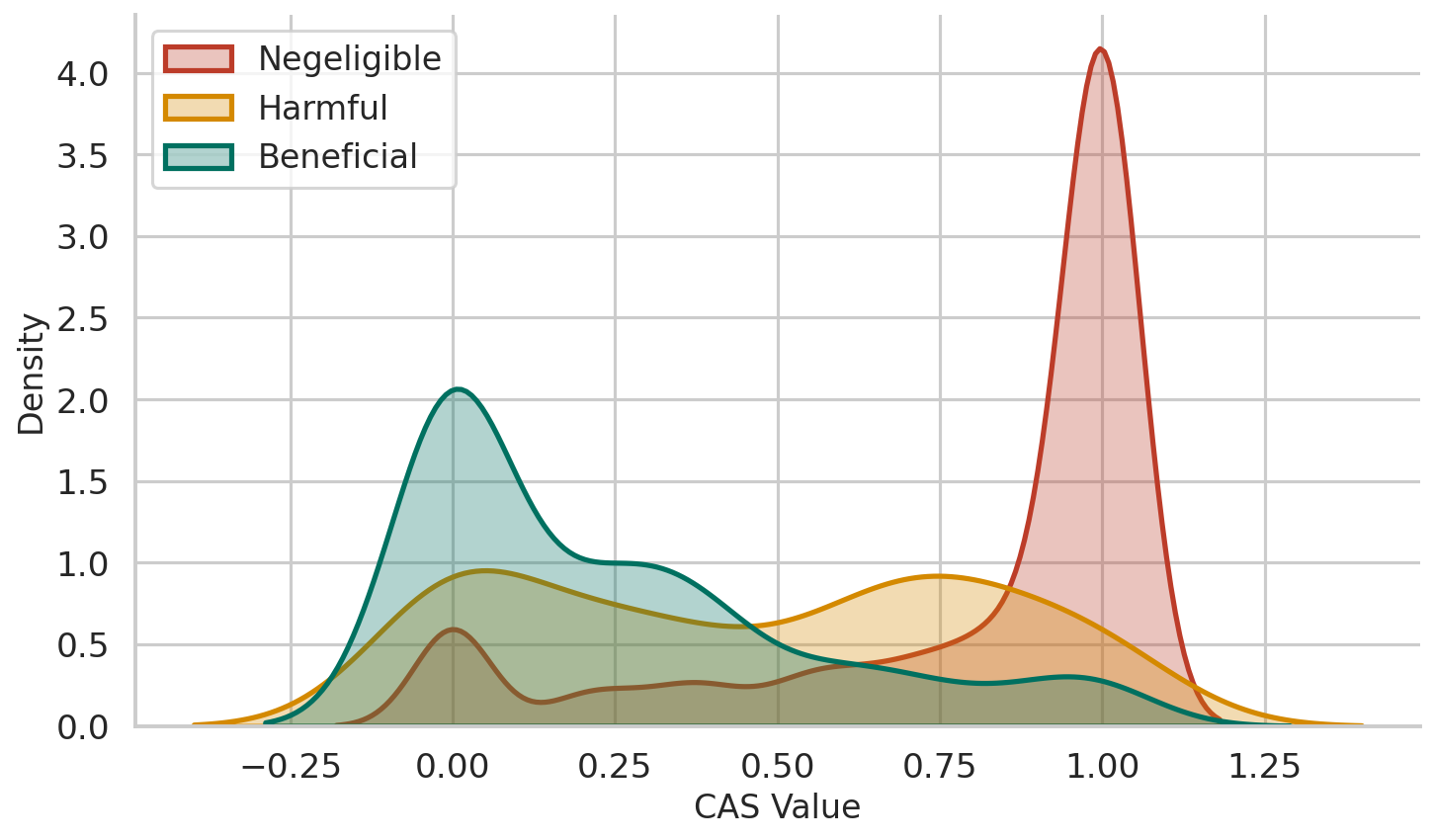}
        \caption{DTD}
    \end{subfigure}
    \hspace{10pt}
    \begin{subfigure}[b]{0.38\textwidth}
        \centering
        \includegraphics[width=\textwidth]{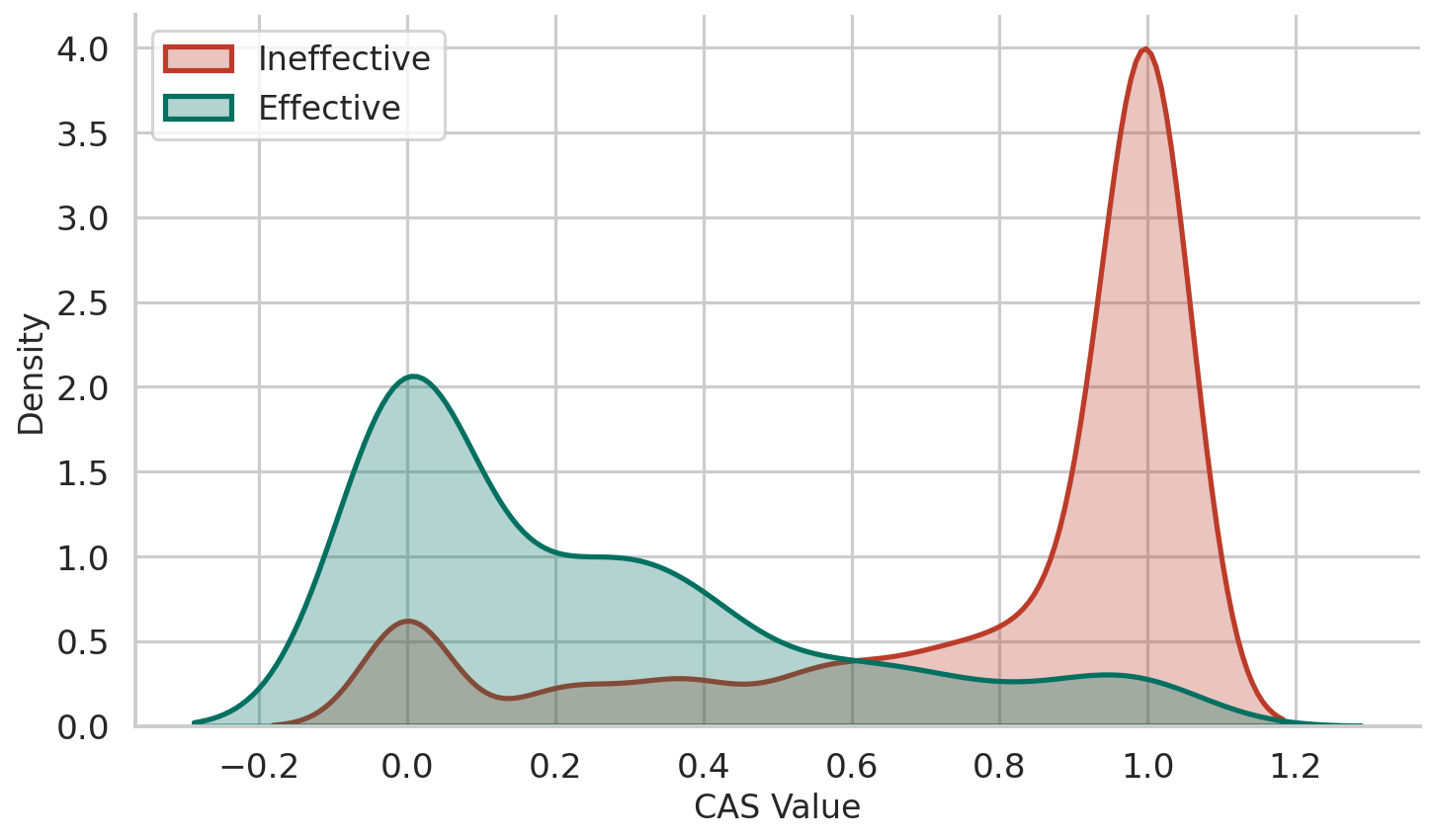}
        \caption{DTD}
    \end{subfigure}

    \vspace{10pt}

    \begin{subfigure}[b]{0.38\textwidth}
        \centering
        \includegraphics[width=\textwidth]{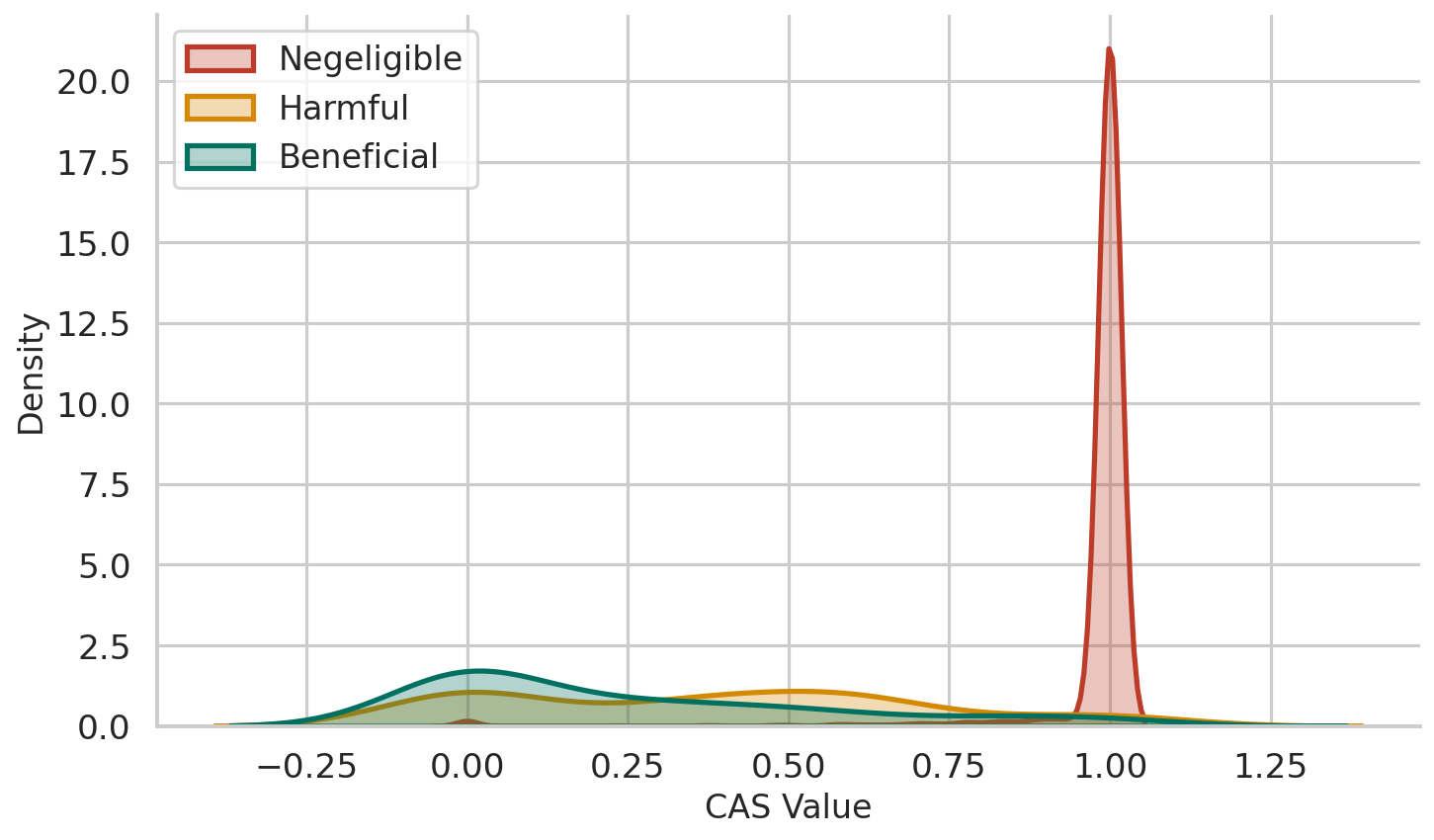}
        \caption{Caltech101}
    \end{subfigure}
    \hspace{10pt}
    \begin{subfigure}[b]{0.38\textwidth}
        \centering
        \includegraphics[width=\textwidth]{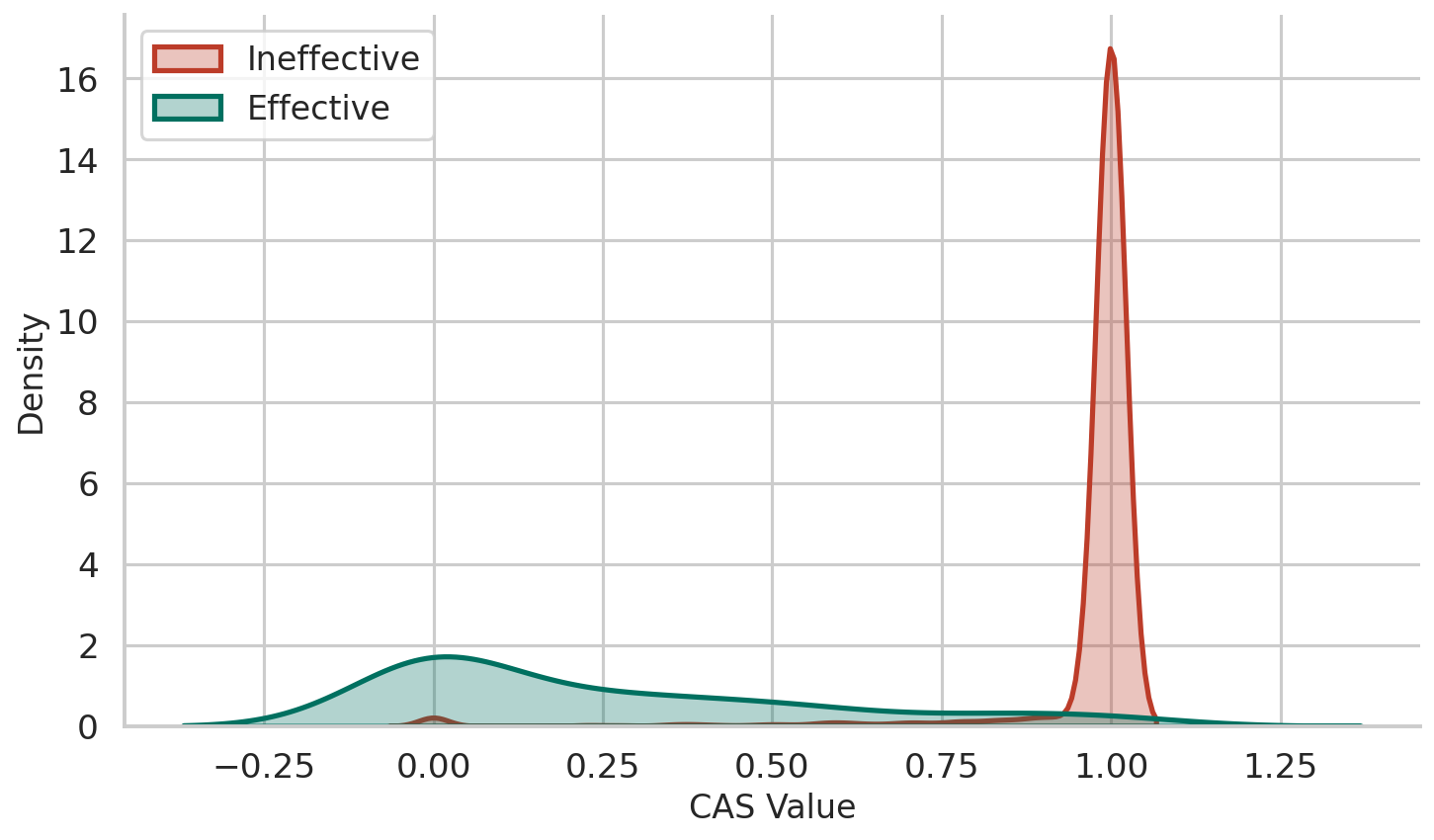}
        \caption{Caltech101}
    \end{subfigure}
    
    \caption{This figure shows CAS distribution for different adaptation cases.
The top row presents the distributions of negligible, harmful, and beneficial cases, while the bottom row groups them into ineffective and effective updates. }
    \label{fig:six_distributions}
\end{figure}
\begin{table*}[!h]
\centering
\caption{Standard deviations comparison of different selection strategies under various TTA methods with ViT-B/16 on ImageNet and its variants. }
\label{tab:vit_errorbar_imagenet}
\setlength{\tabcolsep}{8pt}
\resizebox{\linewidth}{!}{
\begin{tabular}{@{}llcccccccccccc@{}}
\toprule
\multirow{2}{*}{Method} & \multirow{2}{*}{Strategy} & \multicolumn{2}{c}{ImageNet} & \multicolumn{2}{c}{ImageNet-A} & \multicolumn{2}{c}{ImageNet-V} & \multicolumn{2}{c}{ImageNet-R} & \multicolumn{2}{c}{ImageNet-K} & \multicolumn{2}{c}{Avg.} \\
\cmidrule(lr){3-4} \cmidrule(lr){5-6} \cmidrule(lr){7-8} \cmidrule(lr){9-10} \cmidrule(lr){11-12} \cmidrule(l){13-14}
& & AUC{$\downarrow$} & AEP{$\downarrow$} & AUC{$\downarrow$} & AEP{$\downarrow$} & AUC{$\downarrow$} & AEP{$\downarrow$} & AUC{$\downarrow$} & AEP{$\downarrow$} & AUC{$\downarrow$} & AEP{$\downarrow$} & AUC{$\downarrow$} & AEP{$\downarrow$} \\
\midrule
\multirow{4}{*}{TPT~\cite{shu2022testtime}}
& Random & 0.0272 & \text{0.0000} & 0.0240 & 0.0006 & 0.7056 & 0.0066 & 0.0121 & 0.0006 & 0.1849 & 0.0015 & 0.1908 & 0.0019 \\
& Energy~\cite{liu2020energy} & \text{0.0016} & 0.0001 & 0.2401 & 0.0090 & 0.3025 & 0.0083 & 0.0441 & \text{0.0000} & 0.0625 & 0.0012 & 0.1302 & 0.0037 \\
& MCM~\cite{ming2022delving} & 0.0064 & 0.0001 & 0.3136 & 0.0053 & 0.4489 & \text{0.0000} & 0.0506 & 0.0018 & 0.1980 &\text{ 0.0001 }& 0.2035 & 0.0015\\
& \cellcolor{casbg}CAS & \cellcolor{casbg}0.0132 & \cellcolor{casbg}\text{0.0000} & \cellcolor{casbg}\text{0.0169} & \cellcolor{casbg}\text{0.0002} & \cellcolor{casbg}\text{0.1156} & \cellcolor{casbg}0.0012 & \cellcolor{casbg}\text{0.0012} & \cellcolor{casbg}0.0013 & \cellcolor{casbg}\text{0.0156} & \cellcolor{casbg}0.0004 & \cellcolor{casbg}\text{0.0325} & \cellcolor{casbg}\text{0.0006} \\
\midrule

\multirow{4}{*}{R-TPT~\cite{sheng2025illusion}}
& Random & 0.0342 & 0.0012 & 0.0210 & 0.0159 & 0.1056 & \text{0.0002} & 0.0144 & 0.0003 & 0.0002 & 0.0008 & 0.0351 & 0.0037 \\
& Energy~\cite{liu2020energy} & 0.0380 & \text{0.0000} & 0.2401 & \text{0.0003} & 0.0441 & 0.0021 & 0.0049 & \text{0.0000} & 0.0004 & \text{0.0005} & 0.0655 & \text{0.0006} \\
& MCM~\cite{ming2022delving} & 0.0272 & 0.0005 & 0.1024 & 0.0261 & 0.0090 & 0.0062 & 0.0156 & \text{0.0000} & \text{0.0001} & 0.0036 & 0.0309 & 0.0073 \\
& \cellcolor{casbg}CAS & \cellcolor{casbg}\text{0.0002} & \cellcolor{casbg}0.0002 & \cellcolor{casbg}\text{0.0196} & \cellcolor{casbg}0.0240 & \cellcolor{casbg}\text{0.0020} & \cellcolor{casbg}0.0042 & \cellcolor{casbg}\text{0.0036} & \cellcolor{casbg}\text{0.0000} & \cellcolor{casbg}0.0225 & \cellcolor{casbg}0.0034 & \cellcolor{casbg}\text{0.0096} & \cellcolor{casbg}0.0064 \\
\midrule

\multirow{4}{*}{STS~\cite{dafnis2025testtime}}
& Random & 0.0100 & 0.0019 & 0.0049 & \text{0.0000} & 0.0870 & 0.0001 & 0.0256 & 0.0009 & 0.0225 & 0.0006 & 0.0300 & 0.0007 \\
& Energy~\cite{liu2020energy} & 0.0006 & 0.0022 & 0.0110 & \text{0.0000} & \text{0.0006} & \text{0.0000} & 0.0182 & \text{0.0001} & \text{0.0064} & 0.0001 & 0.0074 & \text{0.0005} \\
& MCM~\cite{ming2022delving} & \text{0.0000} & \text{0.0018 }& \text{0.0004} & 0.0013 & 0.0064 & 0.0006 & 0.0090 & 0.0010 & 0.0156 & \text{0.0000 }& \text{0.0063 }& 0.0009 \\
& \cellcolor{casbg}CAS & \cellcolor{casbg}0.0012 & \cellcolor{casbg}\text{0.0018} & \cellcolor{casbg}0.0020 & \cellcolor{casbg}0.0028 & \cellcolor{casbg}0.0240 & \cellcolor{casbg}0.0001 & \cellcolor{casbg}\text{0.0000} & \cellcolor{casbg}0.0004 & \cellcolor{casbg}0.0169 & \cellcolor{casbg}0.0002 & \cellcolor{casbg}0.0088 & \cellcolor{casbg}0.0011 \\
\midrule

\multirow{4}{*}{ZERO~\cite{farina2024frustratingly}}
& Random & 0.0100 & \text{0.0050} & \text{0.0030} & \text{0.0001} & 0.0090 & \text{0.0023} & 0.0380 & 0.0024 & 0.0342 & 0.0003 & 0.0188 & \text{0.0020} \\
& Energy~\cite{liu2020energy} & 0.0072 & 0.0051 & 0.0081 & 0.0040 & 0.0729 & 0.0032 & 0.0121 & \text{0.0019} & 0.0225 & \text{0.0001} & 0.0246 & 0.0029 \\
& MCM~\cite{ming2022delving} & \text{0.0000} & 0.0056 & 0.0110 & \text{0.0001} & 0.0506 & 0.0025 & \text{0.0001} & 0.0033 & \text{0.0016} & 0.0002 & 0.0127 & 0.0023 \\
& \cellcolor{casbg}CAS & \cellcolor{casbg}\text{0.0000} & \cellcolor{casbg}0.0082 & \cellcolor{casbg}0.0342 & \cellcolor{casbg}0.0019 & \cellcolor{casbg}\text{0.0006} & \cellcolor{casbg}0.0030 & \cellcolor{casbg}0.0012 & \cellcolor{casbg}0.0052 & \cellcolor{casbg}\text{0.0016} & \cellcolor{casbg}0.0002 & \cellcolor{casbg}\text{0.0075} & \cellcolor{casbg}0.0037 \\
\bottomrule
\end{tabular}
}

\vspace{5em}

\centering
\caption{Standard deviations comparison of different selection strategies under various TTA methods with ViT-B/16 on fine-grained datasets. }
\label{tab:vit_errorbar_fg}
\renewcommand{\arraystretch}{0.8}
\setlength{\tabcolsep}{6pt}
\resizebox{1\linewidth}{!}{
\begin{tabular}{@{}llccccccccccccc@{}}
\toprule
Method & Strategy & Metric & Flow. & DTD & Pets & UCF & Cal. & Air. & Euro. & Cars & Food & SUN & Avg. \\
\midrule
\multirow{8}{*}{TPT~\cite{shu2022testtime}}
& \multirow{2}{*}{Random} & AUC &0.0342 & 0.1482 & 0.0182 & 0.4624 & 0.0009 & 1.0000 & 0.0380 & 0.1369 & 0.0042 & 0.0100 & 0.1853  \\
&  & AEP &0.0005 & 0.0008 & 0.0009 & 0.0300 & 0.0041 & 0.0007 & 0.0025 & 0.0018 & 0.0000 & 0.0020 & 0.0043 \\
\cmidrule(lr){2-14}
& \multirow{2}{*}{Energy~\cite{liu2020energy}} & AUC & 0.1640 & 0.1056 & 0.0552 & 0.0306 & 4.6440 & 0.0484 & 0.0420 & 1.8632 & 0.1122 & 0.0225 & 0.7088 \\
&  & AEP & 0.0001 & 0.0005 & 0.0001 & 0.0699 & 0.0002 & 0.0280 & 0.0014 & 0.0117 & 0.0000 & 0.0001 & 0.0112\\
\cmidrule(lr){2-14}
& \multirow{2}{*}{MCM~\cite{ming2022delving}} & AUC & 0.0484 & 0.7656 & 1.1990 & 0.0306 & 3.2942 & 1.2656 & 0.0306 & 0.0240 & 0.1444 & 0.0702 & 0.6873   \\
&  & AEP & 0.0017 & 0.0108 & 0.0000 & 0.0772 & 0.0021 & 0.0046 & 0.0004 & 0.0001 & 0.0001 & 0.0002 & 0.0097\\
\cmidrule(lr){2-14}
& \cellcolor{casbg} & \cellcolor{casbg}AUC & \cellcolor{casbg}0.6889 & \cellcolor{casbg}0.4290 & \cellcolor{casbg}0.3969 & \cellcolor{casbg}0.7482 & \cellcolor{casbg} 0.0306 & \cellcolor{casbg}0.6724 & \cellcolor{casbg}0.0009 & \cellcolor{casbg}0.0900  & \cellcolor{casbg}0.0004& \cellcolor{casbg} 0.0016 & \cellcolor{casbg}0.3059  \\
& \cellcolor{casbg}\multirow{-2}{*}{CAS} & \cellcolor{casbg}AEP & \cellcolor{casbg}0.0011 & \cellcolor{casbg}0.0022 & \cellcolor{casbg}0.0019& \cellcolor{casbg}0.0594 & \cellcolor{casbg} 0.0116  & \cellcolor{casbg}0.0207& \cellcolor{casbg}0.0029& \cellcolor{casbg}0.0047 & \cellcolor{casbg}0.0001  & \cellcolor{casbg}0.0004 & \cellcolor{casbg}0.0105 \\
\midrule

\multirow{8}{*}{R-TPT~\cite{sheng2025illusion}}
& \multirow{2}{*}{Random} & AUC &0.7921 & 1.1342 & 0.0144 & 0.0784 & 0.0240 & 2.5760 & 0.0324 & 0.3844 & 0.0441 & 0.0600 & 0.5140 \\
&  & AEP &  0.0020 & 0.0179 & 0.0002 & 0.0537 & 0.0011 & 0.0001 & 0.0293 & 0.0005 & 0.0004 & 0.0012 & 0.0106  \\
\cmidrule(lr){2-14}
& \multirow{2}{*}{Energy~\cite{liu2020energy}} & AUC & 0.0400 & 0.0420 & 0.7744 & 0.1444 & 0.6480 & 0.0009 & 0.0072 & 0.0169 & 0.0784 & 0.0042 & 0.1756\\
&  & AEP & 0.0084 & 0.0522 & 0.0000 & 0.0034 & 0.0020 & 0.0098 & 0.0336 & 0.0145 & 0.0018 & 0.0015 & 0.0127  \\
\cmidrule(lr){2-14}
& \multirow{2}{*}{MCM~\cite{ming2022delving}} & AUC & 0.5852 & 1.5500 & 0.8372 & 0.0009 & 2.8900 & 0.6480 & 0.0361 & 0.2209 & 0.2756 & 0.0121 & 0.7056\\
&  & AEP & 0.0044 & 0.1001 & 0.0000 & 0.0144 & 0.0029 & 0.0392 & 0.0263 & 0.0161 & 0.0032 & 0.0009 & 0.0208   \\
\cmidrule(lr){2-14}
& \cellcolor{casbg} & \cellcolor{casbg}AUC & \cellcolor{casbg}0.6889& \cellcolor{casbg}0.7832 & \cellcolor{casbg}0.0210 & \cellcolor{casbg}0.1521 & \cellcolor{casbg}0.0006 & \cellcolor{casbg}5.7121  & \cellcolor{casbg}0.0012  & \cellcolor{casbg}0.0110 & \cellcolor{casbg}0.0006 & \cellcolor{casbg}0.0016  & \cellcolor{casbg}0.7372 \\
& \cellcolor{casbg}\multirow{-2}{*}{CAS} & \cellcolor{casbg}AEP & \cellcolor{casbg}0.0098 & \cellcolor{casbg}0.0651 & \cellcolor{casbg}0.0002 & \cellcolor{casbg}0.0325 & \cellcolor{casbg}0.0001 & \cellcolor{casbg}0.0930 & \cellcolor{casbg}0.0218 & \cellcolor{casbg}0.0072& \cellcolor{casbg}0.0023  & \cellcolor{casbg}0.0018  & \cellcolor{casbg}0.0234 \\
\midrule

\multirow{8}{*}{STS~\cite{dafnis2025testtime}}
& \multirow{2}{*}{Random} & AUC & 0.5041 & 1.4762 & 0.7744 & 0.3721 & 0.0870 & 2.2350 & 0.2209 & 0.6006 & 0.0784 & 0.0992 & 0.6448  \\
&  & AEP & 0.0075 & 0.0000 & 0.0044 & 0.0079 & 0.0008 & 0.0687 & 0.0011 & 0.0003 & 0.0005 & 0.0013 & 0.0093 \\
\cmidrule(lr){2-14}
& \multirow{2}{*}{Energy~\cite{liu2020energy}} & AUC & 0.0016 & 0.0030 & 0.0169 & 1.7161 & 0.0576 & 0.2862 & 0.0144 & 0.0182 & 0.0182 & 0.0144 & 0.2147 \\
&  & AEP & 0.0228 & 0.0056 & 0.0108 & 0.0881 & 0.0178 & 0.1492 & 0.0001 & 0.0024 & 0.0001 & 0.0022 & 0.0299 \\
\cmidrule(lr){2-14}
& \multirow{2}{*}{MCM~\cite{ming2022delving}} & AUC &0.0961 & 1.3806 & 0.0002 & 0.4970 & 0.9216 & 1.2432 & 0.0420 & 0.1521 & 0.0400 & 0.0042 & 0.4377\\
&  & AEP & 0.0239 & 0.0024 & 0.0086 & 0.0584 & 0.0157 & 0.1277 & 0.0020 & 0.0165 & 0.0003 & 0.0035 & 0.0259  \\
\cmidrule(lr){2-14}
& \cellcolor{casbg} & \cellcolor{casbg}AUC & \cellcolor{casbg}\text{0.0342} & \cellcolor{casbg}0.0090  & \cellcolor{casbg}0.0020  & \cellcolor{casbg}0.0030  & \cellcolor{casbg}0.0004  & \cellcolor{casbg}0.1560  & \cellcolor{casbg}0.0036& \cellcolor{casbg}0.0020  & \cellcolor{casbg}0.0001 & \cellcolor{casbg}0.0049  & \cellcolor{casbg}0.0215  \\
& \cellcolor{casbg}\multirow{-2}{*}{CAS} & \cellcolor{casbg}AEP & \cellcolor{casbg}0.0176& \cellcolor{casbg}0.0057 & \cellcolor{casbg}0.0120 & \cellcolor{casbg}0.0481 & \cellcolor{casbg}0.0263 & \cellcolor{casbg}0.1517 & \cellcolor{casbg}0.0009 & \cellcolor{casbg}0.0056  & \cellcolor{casbg}0.0009& \cellcolor{casbg}0.0022& \cellcolor{casbg}0.0271 \\
\midrule

\multirow{8}{*}{ZERO~\cite{farina2024frustratingly}}
& \multirow{2}{*}{Random} & AUC & 0.2116 & 0.8836 & 1.5252 & 1.3225 & 0.0225 & 0.0196 & 0.0002 & 0.0064 & 0.2916 & 0.0484 & 0.4332 \\
&  & AEP & 0.0128 & 0.0023 & 0.0006 & 0.0287 & 0.0007 & 0.0002 & 0.0063 & 0.0001 & 0.0015 & 0.0008 & 0.0054 \\
\cmidrule(lr){2-14}
& \multirow{2}{*}{Energy~\cite{liu2020energy}} & AUC & 2.7225 & 0.8649 & 0.0400 & 0.0132 & 0.0020 & 0.0156 & 0.0484 & 0.0400 & 0.0132 & 0.0056 & 0.3765 \\
&  & AEP &0.0138 & 0.0000 & 0.0005 & 0.0421 & 0.0045 & 0.0001 & 0.0060 & 0.0003 & 0.0035 & 0.0002 & 0.0071 \\
\cmidrule(lr){2-14}
& \multirow{2}{*}{MCM~\cite{ming2022delving}} & AUC &0.8372 & 0.4290 & 0.5929 & 0.1681 & 0.0900 & 0.0441 & 0.0009 & 0.5329 & 0.0036 & 0.0289 & 0.2728 \\
&  & AEP & 0.0172 & 0.0083 & 0.0002 & 0.0328 & 0.0089 & 0.0004 & 0.0005 & 0.0061 & 0.0037 & 0.0000 & 0.0078\\
\cmidrule(lr){2-14}
& \cellcolor{casbg} & \cellcolor{casbg}AUC & \cellcolor{casbg}0.0121 & \cellcolor{casbg}0.0056& \cellcolor{casbg}0.0121 & \cellcolor{casbg}0.0324 & \cellcolor{casbg}0.7482& \cellcolor{casbg}4.3681 & \cellcolor{casbg}0.0380 & \cellcolor{casbg}0.0552 & \cellcolor{casbg}0.0000 & \cellcolor{casbg}0.0100 & \cellcolor{casbg}0.5282 \\
& \cellcolor{casbg}\multirow{-2}{*}{CAS} & \cellcolor{casbg}AEP & \cellcolor{casbg}0.0179 & \cellcolor{casbg}0.0064& \cellcolor{casbg}0.0000  & \cellcolor{casbg}0.0283  & \cellcolor{casbg}0.0034  & \cellcolor{casbg}0.0035 & \cellcolor{casbg}0.0018  & \cellcolor{casbg}0.0018& \cellcolor{casbg}0.0034 & \cellcolor{casbg}0.0007 & \cellcolor{casbg}0.0067  \\
\bottomrule
\end{tabular}
}
\end{table*}

\section{CAS Distribution}
To examine whether CAS can identify adaptation effectiveness, we visualize the distribution of CAS values for different cases across multiple datasets. 
As shown in Fig.~\ref{fig:six_distributions}, the CAS values exhibit clear separation among the three categories. 
Beneficial cases (i.e., effective adaptations) are concentrated around small CAS values, whereas negligible and harmful cases (i.e., ineffective adaptations) tend to produce significantly larger CAS values. 
In particular, harmful and negligible updates show similar distributions with peaks at high CAS regions, indicating that ineffective adaptations generally correspond to large CAS scores. 
This consistent pattern across datasets suggests that CAS serves as an effective indicator to distinguish effective and ineffective adaptation cases.

\section{Standard Deviations}
We detail the standard deviations of selection strategies on ImageNet and its variants in Table~\ref{tab:vit_errorbar_imagenet}. Furthermore, the results on fine-grained and downstream benchmarks are provided in Table~\ref{tab:vit_errorbar_fg}, which consistently show the stable performance of CAS across different domains.

\end{document}